\documentclass[pdflatex,sn-mathphys-num]{sn-jnl}% Math and Physical Sciences Numbered Reference Style
\usepackage{graphicx}%
\usepackage{multirow}%
\usepackage{amsmath,amssymb,amsfonts}%
\usepackage{amsthm}%
\usepackage{mathrsfs}%
\usepackage[title]{appendix}%
\usepackage{xcolor}%
\usepackage{textcomp}%
\usepackage{manyfoot}%
\usepackage{booktabs}%
\usepackage{algorithm}%
\usepackage{algorithmicx}%
\usepackage{algpseudocode}%
\usepackage{listings}%

\usepackage{tabularx,booktabs}
\usepackage{algorithm}
\usepackage{algpseudocode}
\usepackage{adjustbox}
\usepackage{comment}
\usepackage{pifont}     % for \cmark
\newcommand{\cmark}{\ding{51}} % ✓

\theoremstyle{thmstyleone}%
\theoremstyle{thmstyletwo}%

\theoremstyle{thmstylethree}%

\begin{document}

\title[Article Title]{A Comprehensive Review of Reinforcement Learning for Autonomous Driving in the CARLA Simulator}


\author[1]{\fnm{Elahe} \sur{Delavari}}\email{elahed@umich.edu}
\equalcont{These authors contributed equally to this work.}

\author[1]{\fnm{Feeza} \sur{Khan Khanzada}}\email{feezakk@umich.edu}
\equalcont{These authors contributed equally to this work.}

\author*[1]{\fnm{Jaerock} \sur{Kwon}}\email{jrkwon@umich.edu}

\affil*[1]{\orgdiv{Electrical and Computer Engineering}, \orgname{University Of Michigan-Dearborn}, \orgaddress{\street{4901 Evergreen Road}, \city{Dearborn}, \postcode{48128}, \state{MI}, \country{USA}}}


%\abstract{The abstract serves both as a general introduction to the topic and as a brief, non-technical summary of the main results and their implications. Authors are advised to check the author instructions for the journal they are submitting to for word limits and if structural elements like subheadings, citations, or equations are permitted.}

\abstract{
Autonomous‑driving research has recently embraced deep Reinforcement Learning (RL) as a promising framework for data‑driven decision making, yet a clear picture of how these algorithms are currently employed, benchmarked and evaluated is still missing. This survey fills that gap by systematically analysing around 100 peer‑reviewed papers that train, test or validate RL policies inside the open‑source CARLA simulator. We first categorize the literature by algorithmic family—model‑free, model‑based, hierarchical, and hybrid—and quantify their prevalence, highlighting that more than 80\% of existing studies still rely on model‑free methods such as DQN, PPO and SAC. Next, we explain the diverse state, action and reward formulations adopted across works, illustrating how choices of sensor modality (RGB, LiDAR, BEV, semantic maps, and carla kinematics states), control abstraction (discrete vs. continuous) and reward shaping are used across various literature. We also consolidate the evaluation landscape by listing the most common metrics (success rate, collision rate, lane deviation, driving score) and the towns, scenarios and traffic configurations used in CARLA benchmarks. Persistent challenges including sparse rewards, sim‑to‑real transfer, safety guarantees and limited behaviour diversity are distilled into a set of open research questions, and promising directions such as model‑based RL, meta‑learning and richer multi‑agent simulations are outlined. By providing a unified taxonomy, quantitative statistics and a critical discussion of limitations, this review aims to serve both as a reference for newcomers and as a roadmap for advancing RL‑based autonomous driving toward real‑world deployment. 
}




\maketitle
\section{Introduction}

Autonomous driving represents one of the most transformative technological frontiers of the 21st century, with the potential to revolutionize transportation by enhancing safety, efficiency, and accessibility. By reducing human error—widely recognized as the leading cause of traffic accidents \cite{nhtsa2015crashes}
— Autonomous Vehicles (AVs) promise safer roads, reduced congestion, and improved mobility. However, the development of robust and reliable AV systems remains a formidable challenge, primarily due to the complexity of real-world driving environments. These environments present dynamic traffic patterns, intricate multi-agent interactions, and rare but critical long-tail scenarios that traditional rule-based approaches struggle to handle effectively.

Reinforcement Learning (RL) has emerged as a powerful paradigm for training autonomous driving policies, offering an adaptive framework where agents learn by interacting with their environment and refining their decision-making strategies over time \cite{Sallab_2017}. Unlike conventional supervised learning approaches that require extensive labeled datasets, RL enables AVs to optimize actions dynamically based on rewards, making it particularly suited for sequential decision-making tasks such as lane-keeping, obstacle avoidance, intersection handling, and route planning. Despite its potential, deploying RL-based autonomous systems in the real world poses several challenges, including safety risks, computational demands, and the need for active environmental interaction for training.

To address these challenges, simulation platforms have become an indispensable tool for RL research in autonomous driving. Among these, CARLA (Car Learning to Act) \cite{CARLA} has established itself as a premier open-source simulator designed specifically for AV research. Built on Unreal Engine, CARLA provides high-fidelity urban and highway driving environments, supporting diverse road layouts, dynamic traffic conditions, various weather scenarios, and an extensive suite of sensors (e.g., RGB cameras, LiDAR, depth sensors, and semantic segmentation). This flexibility enables safe, controlled, and scalable RL training, allowing researchers to benchmark and refine driving policies before real-world deployment.

Over the years, CARLA has become a cornerstone for RL-driven autonomous driving research, fostering advancements across multiple dimensions. Researchers have leveraged a wide array of RL techniques  \cite{wu_lane_2022,goel_adaptive_2021,youssef_deep_2019,carton_using_2021,mohammed_unified_2024,zhang_toward_2022,yang_decision-making_2022,wang_vision-based_2023,wei_continual_2023,zhu_rita_2023,Think2Drive,Model-Based_Reinforcement_Learning_with_Isolated_Imaginations,Vision-Based_Autonomous_Car_Racing_Using_Deep_Imitative_Reinforcement_Learning,Learning_to_drive_from_a_world_on_rails,Autonomous_Driving_via_Knowledge-Enhanced,Deductive_Reinforcement_Learning,Integrating_Deep_Reinforcement_Learning_with_Model-based_Path_Planners_for_Automated_Driving}, including model-free and model-based approaches for handling variety of driving tasks. Furthermore, multimodal RL frameworks have been explored to incorporate rich sensor data, improving decision-making under uncertainty. While these methods have significantly advanced the field, challenges such as sparse reward signals, generalization across diverse environments, and the sim-to-real transfer gap remain open problems that demand further investigation.

This paper provides a comprehensive review of RL research within the CARLA simulator, analyzing the evolution of RL techniques and their applications in autonomous driving. The key contributions of this review are as follows:

\begin{itemize}
    \item \textbf{Survey of RL Formulations} – We explore various RL approaches employed in CARLA, including model-free methods such as the Deep Q-Network (DQN), Proximal Policy Optimization (PPO), and Soft Actor-Critic (SAC), Deep Deterministic Policy Gradient (DDPG); model-based techniques; hierarchical RL; and hybrid strategies that integrate multiple learning paradigms.
    \item \textbf{Comparison of State and Action Spaces} – We examine how different sensor modalities (camera, LiDAR, multi-modal fusion) and control abstractions (low-level throttle/steering vs. high-level discrete maneuvers) impact policy learning and performance.
    \item \textbf{Reward Engineering and Terminal condition in RL for Autonomous Driving} – We highlight the critical role of reward function design, analyzing how different formulations balance safety, comfort, and efficiency in various driving tasks.
    \item \textbf{Evaluation metrics and Scenarios for RL Policies} – We review the evaluation methodologies used in CARLA, including key performance metrics such as collision rates, lane deviation, speed regulation, and success rates in navigation tasks along with the different scenarios used for training and testing.
   \item \textbf{Challenges and Future Directions} – We discuss persistent challenges in RL-driven autonomous driving often mentioned in the literature, including generalization to unseen environments, robustness in high-density traffic scenarios, and interpretability of learned strategies. Additionally, we outline promising avenues for future research to bridge these gaps.

\end{itemize}

By synthesizing recent advancements and identifying key challenges, this review aims to serve as a valuable resource for researchers and practitioners in the field. As simulation platforms like CARLA continue to evolve, they will play an increasingly critical role in the pursuit of safe and reliable autonomous driving systems. Through this work, we hope to provide clarity on the state of RL in CARLA,  fostering further innovation and accelerating progress toward real-world deployment. This review focuses on single-agent RL approaches in the CARLA simulator. While Multi-Agent RL (MARL) is a relevant and growing area of research—particularly for urban traffic scenarios and coordination among multiple vehicles—it is beyond the scope of this review. we consider around 100 papers from IEEE and ACM.

\section{CARLA simulator}
CARLA is an open-source simulator explicitly designed to support research in autonomous driving and ADAS. Built on the Unreal Engine 4 \footnote{\url{https://www.unrealengine.com/}}, CARLA provides a photorealistic 3D environment with configurable urban and highway scenes that include diverse road layouts, dynamic weather, and day–night cycles. Crucially, it exposes a flexible Python/C++ API for scene manipulation, traffic generation, and sensor configuration, enabling researchers to script complex scenarios and integrate with ROS \cite{quigley2009ros}, Autoware \cite{autowarefoundaion_autoware}, or Apollo \cite{autowarefoundaion_autoware} stacks.

A key strength of CARLA lies in its extensible sensor suite. Out of the box, users can deploy RGB/depth cameras, LiDAR, radar, GPS, IMU, and semantic segmentation sensors—each with fully customizable intrinsic and extrinsic parameters. Custom sensors may also be defined via JSON, and ground-truth data (e.g., 2D/3D bounding boxes, semantic labels) can be obtained directly from the environment. This level of fidelity allows for rigorous evaluation of perception and sensor-fusion algorithms under controlled yet realistic conditions.

Underpinning CARLA’s realism is its client–server architecture. The server handles all physics, rendering, and world updates—leveraging GPU acceleration for accurate vehicle dynamics (including wheel friction, suspension, and center-of-mass modeling) and environment effects—while lightweight clients manage agent logic and scenario control. A built-in recorder facilitates seamless capture and replay of both sensor streams and control commands, supporting reproducible experiments and accelerated headless simulations.

In the comprehensive survey by Kaur et al. \cite{kaur2021surveysimulatorstestingselfdriving}, CARLA and LGSVL \cite{rong2020lgsvlsimulatorhighfidelity} emerge as the state-of-the-art open-source simulators for end-to-end autonomous-vehicle testing, outperforming MATLAB/Simulink, CarSim, PreScan, and Gazebo across key requirements such as high-fidelity 3D environments, sensor variety, and scenario management. Unlike Gazebo—which demands manual creation of models and scene geometry via XML—CARLA supplies urban maps and traffic infrastructure (traffic lights, stop signs, lane markings) automatically from OpenDRIVE files, thus drastically reducing setup overhead. Moreover, CARLA’s seamless integration with major autonomy stacks via native ROS bridges contrasts with the external tooling required by LGSVL and others.

CARLA’s open-source ecosystem further distinguishes it from proprietary platforms (e.g., Waymo’s CarCraft, Uber’s DataViz). Researchers can inspect, extend, or optimize core modules—ranging from physics engines to AI agents—and contribute back improvements. This collaborative model accelerates feature development (e.g., the recent addition of an RSS-based safety assurance module) and ensures that CARLA remains at the forefront of sim-to-real research challenges.

\section{Reinforcement Learning Methods}

RL methods used in CARLA vary widely in their learning paradigms, exploration strategies, and suitability for different types of control tasks. Broadly, they can be categorized into model-free and model-based approaches, with several hybrid and alternative techniques (e.g., distributional, Bayesian, and hierarchical RL) gaining traction in recent years. Model-free methods like DQN, PPO, and SAC have become standard choices for end-to-end driving due to their empirical robustness, while model-based approaches aim for higher sample efficiency by learning environment dynamics. This section reviews the RL algorithms applied in CARLA-based driving research, analyzing their design, use cases, and key contributions.

\begin{figure}[htbp]
    \centering
    \includegraphics[width=0.48\textwidth]{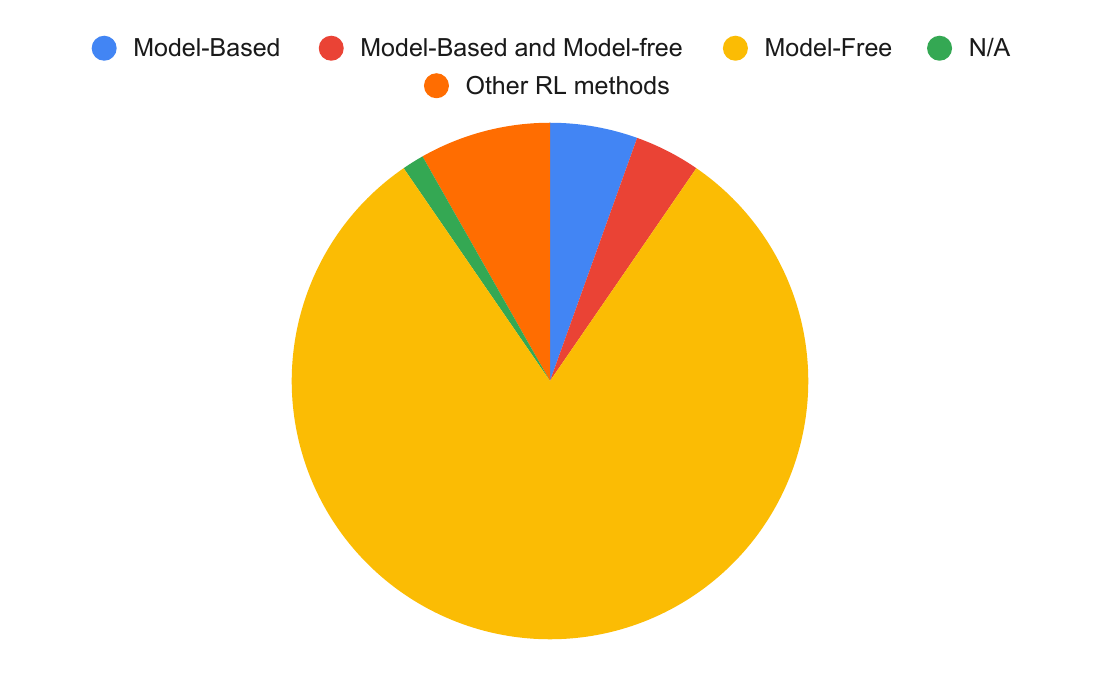} % Replace with actual file path
    \caption{Distribution of RL model types used in CARLA-based studies. Model-free approaches dominate the literature (over 80\%), while model-based and hybrid methods are significantly less explored.}
    \label{fig:rl_models_distribution}
\end{figure}

 Despite the potential for improved sample efficiency, MBRL remains underexplored in CARLA, as shown in Fig.~\ref{fig:rl_models_distribution}. The majority of studies continue to rely on model-free algorithms due to their relative ease of implementation and empirical robustness.

\begin{figure}[htbp]
    \centering
    \includegraphics[width=0.48\textwidth]{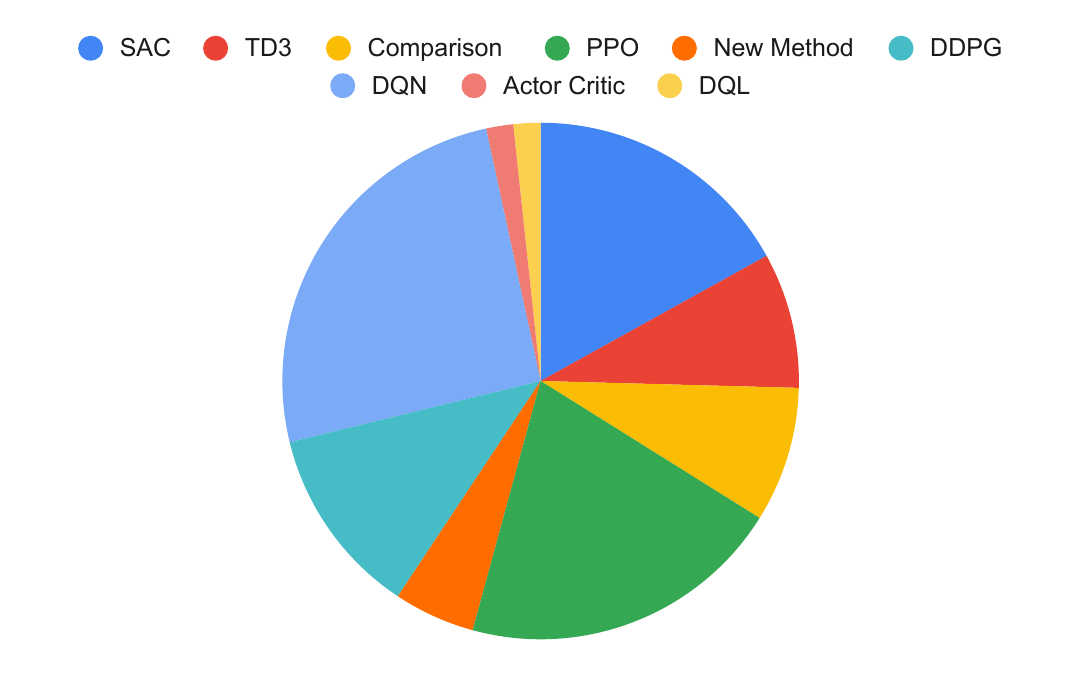} % Replace with actual file path
    \caption{Distribution of different model free RL. }
    \label{fig:rl_models_pie}
\end{figure}

As shown in Fig.~\ref{fig:rl_models_distribution}, model-free approaches dominate the CARLA literature, accounting for over 80 \% of all surveyed works, while model-based and hybrid methods remain relatively rare. Fig.~\ref{fig:rl_models_pie} drills into that model-free slice—highlighting the shares of DQN, PPO, SAC, DDPG, etc.  To complement these visual summaries, Table~\ref{tab:method_usage} provides a comprehensive listing of representative CARLA publications, organized by algorithmic family and method variant. This table serves as a roadmap for the subsections that follow, where we review each category in detail.

\subsection{Model-Free Reinforcement Learning}
Model-free RL (MFRL) methods learn optimal policies directly from experience without explicitly modeling the environment's transition dynamics. These approaches are particularly well-suited to high-dimensional and complex environments like CARLA, where learning accurate dynamics models can be challenging. Model-free methods are typically divided into value-based approaches (e.g., DQN), which estimate action-value functions, and policy-based or actor-critic methods (e.g., PPO, DDPG, SAC), which directly optimize the policy with or without value function guidance. Although model-free methods often require large amounts of training data, they are widely used for end-to-end driving tasks in the CARLA simulator, as shown in Fig. \ref{fig:rl_models_distribution}. 
%\subsubsection{Actor Critic - Feeza}

\subsubsection{Actor Critic}

Actor-critic methods combine value-based and policy-based RL by maintaining two main components: an \emph{actor} and a \emph{critic}. The actor is a parameterized policy, often denoted as \(\pi_\theta(a \mid s)\), which directly maps states to actions or action probabilities. The critic is a value function approximator, such as \(V_\omega(s)\) or \(Q_\omega(s,a)\), used to evaluate and guide updates for the actor.

In the standard RL setting, an agent interacts with the environment through states $s_t \in \mathcal{S}$, actions $a_t \in \mathcal{A}$, and rewards $r_t \in \mathbb{R}$. At each timestep $t$, the agent observes a state $s_t$, selects an action $a_t$ according to its policy, and receives a reward $r_t$ along with the next state $s_{t+1}$. 

In a typical actor-critic setting, the critic parameters \(\omega\) are updated by minimizing a Temporal-Difference (TD) error. For example, if the critic is a value function \(V_\omega\), the update target for a given transition \((s_t, a_t, r_t, s_{t+1})\) might be

\begin{equation}
 y_t = r_t + \gamma V_\omega(s_{t+1})
 \label{eq:target}
\end{equation}

and the critic's objective becomes

\begin{equation}
\mathcal{L}(\omega) = \left( V_\omega(s_t) - y_t \right)^2,
\label{eq:critic_loss}
\end{equation}

where \(\gamma\) is the discount factor.

The actor parameters \(\theta\) are updated by following the policy gradient, which often uses an advantage estimate to reduce variance. An example update rule is

\begin{equation}
\nabla_{\theta} J(\theta) \approx \frac{1}{N} \sum_{t=1}^{N} \nabla_{\theta} \log \pi_{\theta}(a_t \mid s_t) A_t,
\label{eq:policy_gradient}
\end{equation}

where \(A_t\) is an advantage function. One simple choice for the advantage is the TD error
\begin{equation}
A_t = r_t + \gamma V_\omega(s_{t+1}) - V_\omega(s_t).
\label{eq:advantage}
\end{equation}
Thus, the actor adjusts its policy in directions suggested by the critic’s estimated advantage of taking action \(a_t\) in state \(s_t\). By updating the critic to better approximate the value function and updating the actor to maximize returns according to the critic’s feedback, actor-critic methods can learn efficiently in environments with large or continuous action spaces. The learned critic provides lower-variance gradient estimates than pure policy gradient methods, and the actor allows the policy to be optimized continuously rather than relying solely on value-based approaches.  In the self-driving domain, Chronis \emph{et al.}~\cite{chronis_driving_2021} and Udatha \emph{et al.}~\cite{udatha_reinforcement_2023} have employed actor–critic variants using separate networks for policy and value estimation.

\subsubsection{Deep Deterministic Policy Gradient}

The DDPG algorithm~\cite{DDPG_paper} is a model-free, off-policy actor-critic method designed for environments with continuous action spaces. It extends the Deterministic Policy Gradient (DPG) framework~\cite{DPG_paper} by integrating deep neural networks and stabilization techniques introduced in the DQN~\cite{mnih2015human} algorithm. Due to its ability to directly output real-valued actions, DDPG has been widely used in autonomous driving tasks in simulators like CARLA.

DDPG employs two neural networks:
\begin{itemize}
    \item An \textbf{actor} network $\mu(s|\theta^\mu)$ that deterministically maps a state $s$ to a continuous action $a$.
    \item A \textbf{critic} network $Q(s,a|\theta^Q)$ that estimates the expected return (Q-value) of taking action $a$ in state $s$.
\end{itemize}

The critic is trained to minimize the loss between predicted Q-values and target Q-values, defined using the Bellman equation:
\begin{equation}
y_i = r_i + \gamma Q'\left(s_{i+1}, \mu'\left(s_{i+1}|\theta^{\mu'}\right) \middle| \theta^{Q'}\right),
\end{equation}
where:
\begin{itemize}
    \item $r_i$ is the reward at timestep $i$,
    \item $s_{i+1}$ is the next state,
    \item $Q'$ and $\mu'$ are target networks for the critic and actor respectively,
    \item $\gamma \in [0,1]$ is the discount factor,
    \item $\theta^{Q'}$, $\theta^{\mu'}$ are the parameters of the target networks.
\end{itemize}
These target networks are updated using a soft update rule:
\begin{equation}
\theta' \leftarrow \tau \theta + (1 - \tau) \theta',
\end{equation}
where $\tau \ll 1$ is a small constant (e.g., $\tau = 0.001$).
The actor is updated by maximizing the critic's estimate of the expected return. Using the chain rule, the policy gradient is computed as:
\begin{equation}
\nabla_{\theta^\mu} J \approx \mathbb{E}_{s \sim \mathcal{D}} \left[ \nabla_a Q(s, a|\theta^Q)\big|_{a = \mu(s)} \nabla_{\theta^\mu} \mu(s|\theta^\mu) \right],
\end{equation}
where $\mathcal{D}$ is the replay buffer distribution.
To improve stability and exploration, DDPG includes the following components:
\begin{itemize}
    \item \textbf{Replay buffer}: Stores past transitions $(s_t, a_t, r_t, s_{t+1})$ for off-policy training.
    \item \textbf{Target networks}: Slowly updated copies of the actor and critic to stabilize training.
    \item \textbf{Ornstein–Uhlenbeck process}: Temporally correlated noise added to actions for exploration in physical environments with inertia.
    \item \textbf{Batch normalization}: Applied to input and hidden layers to mitigate covariate shift and scale sensitivity.
\end{itemize}

Although DDPG was initially successful across various continuous control benchmarks, it is known to suffer from Q-function overestimation and instability during training. 
Several studies have explored and extended DDPG to address a variety of challenges in autonomous control. 
Wu et al. \cite{wu_lane_2022} applied DDPG to solve high-dimensional control problems, leveraging its suitability for continuous domains. Goel et al. \cite{goel_adaptive_2021} and Youssef et al. \cite{youssef_deep_2019} adopted standard DDPG setups for adaptive and general AV control tasks. Peng et al. \cite{peng_imitative_2020} proposed a hybrid Imitation
learning method fusing visual information with the additional steering angle calculated by Pure-Pursuit (PP) called IPP-RL framework that combines imitation learning with DDPG, using steering signals from a pure pursuit controller to guide the learning process. The study by P{\'e}rez-Gil et al.~\cite{Deep_reinforcement_learning_based_control_for_Autonomous_Vehicles_in_CARLA} presents a comparative analysis of DDPG and DQN, evaluating their control performance in autonomous driving tasks. To enhance learning efficiency, Fu et al.~\cite{Decision_Making_for_Autonomous_Driving_Via_Multimodal_Transforme} combined DDPG with an Information Bottleneck mechanism, aiming to reduce redundancy in state representation and improve sample efficiency. Additionally, Chen et al.~\cite{Towards_Autonomous_Driving} improved DDPG by incorporating self-attention models for visual perception, prioritized experience replay for more effective sampling, and Ornstein-Uhlenbeck noise to encourage better exploration. Zhang et al. \cite{zhangSelflearningLaneKeeping2021} proposed a self-learning lane-keeping algorithm using DDPG. Tsai et al.~\cite{tsaiAutonomousVehicleFollowingTechnique2023} applied DDPG and RDPG (Recurrent Deterministic Policy Gradient) showing that RDPG outperformed DDPG in generalization to unseen scenarios. Doe et al. \cite{doe_dsorl_2023} adopt a canonical actor–critic formulation in which the actor outputs continuous weighting parameters while the critic approximates the state–action value function. They stabilise learning through both an experience-replay buffer and slowly updated target networks. Building on the same foundation, Li et al. \cite{li_dynamic_2022} also use separate actor and critic networks together with replay and target mechanisms, emphasising the algorithm’s off-policy nature to optimise a deterministic policy in real-valued action spaces. Ahmed et al. \cite{ahmed_policy-based_2022} likewise employ DDPG, tailoring the actor to generate continuous steering, throttle and brake commands while the critic guides policy updates via Q-value estimates.

\subsubsection{Deep Q-Learning}

Deep Q-Learning (DQL) extends classical Q-learning by using a deep neural network as a function approximator for the action-value function. Instead of storing a Q-table for all state-action pairs, a parameterized neural network \(Q_\theta(s, a)\) is trained to estimate the expected return. Given a transition \((s, a, r, s')\), the \emph{target} for this network can be written as:

\begin{equation}
y = r + \gamma \max_{a'} Q_{\theta^-}(s', a'),
\label{eq:dqn_target}
\end{equation}

where \(\gamma\) is the discount factor, and \(Q_{\theta^-}\) is a \emph{target network} whose parameters \(\theta^-\) are periodically updated from the main network \(\theta\). The loss function for each mini-batch is

\begin{equation}
\mathcal{L}(\theta) = \left( Q_{\theta}(s, a) - y \right)^2.
\label{eq:q_loss}
\end{equation}

A key mechanism in DQL is the experience replay buffer, which stores past transitions. At each training step, a random mini-batch is sampled from this buffer to break correlation between consecutive samples and stabilize training.

By decoupling the action-value estimation from target computation through a slowly updated target network, and by training on random samples from the replay buffer, DQL mitigates instabilities that arise from training a neural network with correlated data and non-stationary targets. This approach has been highly successful in discrete action domains, particularly in Atari game environments, where it surpasses human-level performance for many titles. In AV domain, Cheng et al. \cite{cheng_longitudinal_2021} utilize such a DQL architecture to achieve effective and stable learning in their longitudinal control setting.

\subsubsection{Proximal Policy Optimization}

PPO~\cite{schulman2017proximal} is a model-free, on-policy policy gradient method that achieves stable and efficient policy updates through a first-order approximation to Trust Region Policy Optimization (TRPO). PPO has gained popularity for its simplicity, generality, and strong empirical performance across both discrete and continuous control tasks, making it a common choice in CARLA-based RL research. PPO maintains a stochastic policy \( \pi_\theta(a|s) \), which is optimized to maximize expected returns by improving the likelihood of advantageous actions. The vanilla policy gradient objective is given by:
\begin{equation}
L^{\text{PG}}(\theta) = \mathbb{E}_t \left[ \log \pi_\theta(a_t | s_t) \hat{A}_t \right],
\end{equation}
where \( \hat{A}_t \) is an estimator of the advantage function at timestep \( t \). While this objective works in principle, performing multiple updates using the same batch of data often leads to instability due to large policy changes.

To address this, PPO introduces a clipped surrogate objective that restricts policy updates to a conservative range. Defining the probability ratio
\begin{equation}
r_t(\theta) = \frac{\pi_\theta(a_t | s_t)}{\pi_{\theta_{\text{old}}}(a_t | s_t)},
\end{equation}
the clipped objective becomes:

\begin{align}
L^{\text{CLIP}}(\theta) = \mathbb{E}_t \big[ \min \big(
& r_t(\theta) \hat{A}_t, \notag \\
& \text{clip}(r_t(\theta), 1 - \epsilon, 1 + \epsilon) \hat{A}_t
\big) \big],
\label{eq:ppo_clip}
\end{align}
where \( \epsilon \) is a small constant (typically 0.1 or 0.2) controlling the trust region size. This objective penalizes updates that move the new policy too far from the old one, ensuring stable learning while still enabling improvement when \( \hat{A}_t \) is positive.
The complete loss function includes terms for value function fitting and entropy regularization:
\begin{equation}
L^{\text{PPO}}(\theta) = \mathbb{E}_t \left[ L^{\text{CLIP}}_t(\theta) - c_1 L^{\text{VF}}_t(\theta) + c_2 S[\pi_\theta](s_t) \right],
\end{equation}
where:
\begin{itemize}
    \item \( L^{\text{VF}}_t = (V_\theta(s_t) - V_t^{\text{target}})^2 \) is the value function loss,
    \item \( S[\pi_\theta](s_t) \) is the entropy of the policy to encourage exploration,
    \item \( c_1 \), \( c_2 \) are weighting coefficients.
\end{itemize}
Each PPO iteration involves collecting rollouts from the environment using the current policy \( \pi_{\theta_{\text{old}}} \), computing advantages (e.g., via Generalized Advantage Estimation), and performing several epochs of minibatch updates to optimize the objective.
PPO avoids the complexity of second-order optimization (as in TRPO) and is easy to integrate with deep architectures.

PPO’s training stability and simplicity have made it the dominant MFRL algorithm in autonomous driving research, leading to its use in a wide range of studies that build on its core strengths. Building on these advantages, Carton et al. \cite{carton_using_2021} implemented PPO directly for autonomous control tasks, while Mohammed et al. \cite{mohammed_unified_2024} used PPO within the RLlib framework to leverage distributed and scalable training. To improve learning stability and exploration, Deng et al. \cite{deng_context_2023} integrated improved exploration strategies, while a follow-up study \cite{deng_context-aware_2024} employed PPO with clipped objectives and Generalized Advantage Estimation (GAE) with the context-aware state as input. Zhao et al. \cite{zhao_end--end_2024} proposed a fully end-to-end autonomous driving pipeline based on PPO. Wu et al. \cite{wu_proximal_2023} also employed PPO for control policy training.

The “Roach” expert agent developed by Zhang et al.~\cite{End-to-End_Urban_Driving_by_Imitating_a_Reinforcement_Learning_Coach} was trained using PPO, demonstrating its effectiveness for expert-level policy imitation. Advanced perception mechanisms have been integrated with PPO in several studies. For instance, Agarwal et al.~\cite{Learning_Urban_Driving_Policies_using_Deep_Reinforcement_Learning} incorporated pretrained BEV-SemSeg autoencoders, while Trumpp et al.~\cite{Efficient_Learning_of_Urban_Driving} employed a Recurrent DriveNet with LSTM layers for temporal state propagation. Xing et al.~\cite{Domain_Adaptation_In_Reinforcement_Learning} addressed domain generalization by combining PPO with a Cycle-Consistent Variational Autoencoder (VAE), enabling zero-shot policy transfer by disentangling domain-specific and general features. Anzalone et al.~\cite{Reinforced_Curriculum_Learning_For_Autonomous_Driving} explored curriculum learning, applying PPO across a five-stage training progression—from simplified conditions to dense urban scenarios. Finally, Silva et al.~\cite{Addressing_Lane_Keeping_and_Intersections} trained three distinct PPO-based sub-policies (lane following, turning left, turning right) under a high-level planner, using data augmentation techniques to enhance robustness and generalization.

Albilani et al. \cite{albilani_guided_2023} embed a PPO variant called GPPO inside a hierarchical Option-Critic framework: expert demonstrations generated by Answer-Set Programming guide low-level option policies, while a high-level controller selects among those options, yielding a two-tier architecture that accelerates learning and improves interpretability.
Martínez-Gómez et al. \cite{martinez_gomez_temporal_2023} adopt a more conventional single-layer PPO agent whose shared actor–critic network outputs continuous speed references for the ego vehicle, capitalising on PPO’s clipped-surrogate objective to ensure stable updates in dense-traffic scenarios. Jin et al. \cite{jin_vwpefficient_2024} enhance the baseline with generalised advantage estimation and prioritised experience replay, producing “PPOE,” which speeds convergence and reduces collision frequency in their highway benchmark. Finally, Shi et al. \cite{shi_efficient_2023} modify PPO for a hybrid high-/low-level decomposition: a convolutional actor–critic network processes deformed occupancy grids and traffic-rule vectors, a hybrid reward couples behavioural and motion objectives, and a small imitation-learning warm-start improves early performance.

\subsubsection{Trust Region Policy Optimization}
 TRPO is a policy gradient method that ensures monotonic policy improvement by constraining the step size during updates, making it well-suited for tasks requiring stable learning. Although less commonly used than PPO due to its computational complexity, TRPO has been successfully applied in autonomous driving scenarios that demand precise control. Gutierrez-Moreno et al. \cite{gutierrez-morenoDecisionMakingAutonomous2024} conducted a comparative study evaluating TRPO alongside PPO, DQN, and A2C. In their experiments targeting merge scenarios, TRPO achieved the highest success rate (92.9\%) and was ultimately chosen for deployment. This result highlights TRPO’s strength in producing reliable policies for complex decision-making tasks under safety-critical constraints.

\subsubsection{Deep Q-Network}

The Deep Q-Network (DQN) algorithm~\cite{mnih2015human} was the first deep RL method to achieve human-level performance on a wide range of tasks using only high-dimensional raw visual input. It combines Q-learning with deep convolutional neural networks to approximate the action-value function \( Q(s, a) \), enabling end-to-end learning of control policies from pixels.
DQN estimates the optimal action-value function:

\begin{align}
Q^*(s,a) = \max_\pi \mathbb{E} \big[\, 
& r_t + \gamma r_{t+1} + \gamma^2 r_{t+2} + \cdots \notag \\
& \mid\, s_t = s, a_t = a, \pi 
\,\big],
\label{eq:optimal_q}
\end{align}
which represents the maximum expected return after taking action \( a \) in state \( s \), and thereafter following policy \( \pi \). The action-value function is approximated by a neural network \( Q(s, a; \theta) \) with parameters \( \theta \), trained to minimize the temporal difference (TD) error using the loss:
\begin{equation}
L_i(\theta_i) = \mathbb{E}_{(s,a,r,s')} \left[ \left( y_i - Q(s, a; \theta_i) \right)^2 \right],
\end{equation}
where the target is defined as:
\begin{equation}
y_i = r + \gamma \max_{a'} Q(s', a'; \theta_i^-),
\end{equation}
and \( \theta_i^- \) are the parameters of a separate \textit{target network} that is held fixed for several iterations and then periodically updated to match \( \theta_i \).

To stabilize learning, DQN introduces two key techniques:
\begin{itemize}
    \item \textbf{Experience replay}: Transitions \( (s_t, a_t, r_t, s_{t+1}) \) are stored in a buffer and sampled uniformly to decorrelate training data and improve data efficiency.
    \item \textbf{Target network}: A separate target Q-network \( Q(s, a; \theta^-) \) is used to compute the TD target \( y_i \), reducing the risk of divergence during updates.
\end{itemize}

DQN uses an \(\epsilon\)-greedy policy for exploration and is typically implemented with a convolutional Q-network that maps image-based states to Q-values over discrete actions.

DQN and its variants—despite being designed for discrete action spaces—have been widely applied in autonomous driving (notably in CARLA) by discretizing continuous controls such as steering and throttle. Notably, Zhang et al. \cite{zhang_toward_2022} implemented a basic DQN with a convolutional neural network (CNN) for image processing. Yang et al. \cite{yang_decision-making_2022} significantly enhanced DQN performance using guided training (G-DDQN), state representation networks (GR-DDQN), dueling architectures (GRSD-DDQN), and safety rule constraints (GRS-DDQN). Bai et al. \cite{bai_fen-dqn_2023} proposed FEN-DQN, which integrates a Feature Extraction Network composed of ResNet-50 and LSTM to generate affordance representations such as distance and angle. Muhammed et al. \cite{muhammed_developing_2021} introduced a Double 3-Step C51 DQN combining Categorical DQN with n-step returns and double Q-learning to reduce value overestimation. Elallid et al. \cite{elallid_dqn-based_2022} used a standard DQN to control an AV and avoid collisions. 

Two studies by Chekroun et al.~\cite{GRI:General_Reinforced_Imitation} and Toromanoff et al.~\cite{End-to-End_Model-Free_Reinforcement_Learning_for_Urban_Driving} employed Rainbow-IQN Ape-X, a variant of DQN that incorporates expert demonstration integration, distributional RL, and prioritized experience replay in distributed settings. Safety-critical decision-making involving braking and driving modules was explored by Marouane et al.~\cite{Safe_Navigation_Based_on_Deep_Q-Network_Algorithm_Using} using DQN. Other studies applied DQN to specific driving scenarios: Muhtadin et al.~\cite{Deep_Reinforcement_Learning_Control_Strategy_at_Roundabout_for_i-CAR} focused on roundabouts, Elallid et al.~\cite{A_Reinforcement_Learning_Based_Approach_for_Controlling_Autonomous_Vehicles_in_Complex_Scenarios} addressed complex traffic conditions, and May et al.~\cite{Using_the_CARLA_Simulator_to_Train_A_Deep_Q_Self-Driving} used CARLA-based simulations for self-driving control. Temporal dynamics were modeled using DQN with CNN-LSTM integration by Ahmed et al.~\cite{A_Deep_Q-Network_Reinforcement_Learning-Based}, while Clemmons et al.~\cite{Reinforcement_Learning-Based_Guidance_of_Autonomous_Vehicles} implemented a DQN-based controller with a non-visual state space and reward shaping to guide an ego vehicle through dynamic, multi-vehicle environments in CARLA. Li et al.~\cite{Learning_Automated_Driving_in_Complex_Intersection_Scenarios_Based} introduced three Q-learning variants—T-DQN, T-DDQN, and T-D3QN—for handling intersections. Finally, Liu et al.~\cite{A_Methodology_Based_on_Deep_Reinforcement_Learning} employed Double Deep Q-Learning (DDQN) to scale RL to large state spaces in complex urban settings. Weng et al. \cite{frijiDQNBasedAutonomousCarFollowing2020} proposed a DQN-based car-following framework that uses RGB-D image inputs to guide control in autonomous driving. Hishmeh et al. \cite{hishmehDeerHeadlightsShort2020} proposed a short-term planning framework for autonomous driving using the Dueling Deep Q-Network (Dueling DQN) algorithm. Dagdanov et al.~\cite{221016567DeFIXDetecting} proposed DeFIX, a hybrid IL-RL framework where a DQN agent is trained on failure scenarios of an IL policy. The method uses a policy classifier to switch between the IL agent and specialized RL agents trained on mini-scenarios extracted from infractions, improving safety and performance in challenging CARLA urban tasks.

Deshpande et al. \cite{deshpande_navigation_2021} propose a thresholded lexicographic Deep Q-learning architecture in which two independent Double DQN agents are trained separately for the safety and speed objectives. At execution time, their outputs are merged lexicographically: the safety agent’s action is accepted outright unless its estimated Q-value falls below a predefined threshold, in which case the speed agent’s recommendation is used. Maintaining distinct replay buffers and networks for the two objectives ensures that safety remains the higher-priority criterion without sacrificing the sample efficiency of off-policy Q-learning. An explainability-oriented study \cite{noauthor_explainable_nodate} augments a baseline DQN with Bayesian deep learning via Monte-Carlo Dropout. Multiple stochastic forward passes yield both the mean and standard deviation of each action’s Q-value; the agent then selects the action whose expected return is maximal and whose epistemic uncertainty lies below a user-defined threshold. This simple uncertainty filter provides a principled mechanism for trading off performance against risk, rendering the learned policy more transparent and trustworthy in safety-critical settings.

\subsubsection{Soft Actor Critic}

Soft Actor-Critic (SAC) \cite{haarnoja_soft_2018} is a state-of-the-art, off-policy actor-critic algorithm that aims to maximize not only the expected cumulative reward but also the \emph{entropy} of the policy. By including an entropy term in the objective, SAC encourages more stochastic policies, which helps achieve better exploration in continuous control tasks and improves robustness during training.
Unlike standard RL objectives that solely maximize expected reward, SAC incorporates an entropy term to encourage exploration. The overall objective for SAC can be written as:

\begin{equation}
    J(\pi) = \mathbb{E}_{\substack{s_t \sim \rho_\pi \\ a_t \sim \pi(\cdot | s_t)}} 
    \left[ 
        \sum_{t=0}^{\infty} \gamma^t \bigl( r(s_t, a_t) + \alpha \mathcal{H}\bigl(\pi(\cdot|s_t)\bigr) \bigr)
    \right],
\end{equation}

where \( \rho_\pi \) denotes the state distribution under the policy \( \pi \), 
\( r(s_t, a_t) \) is the reward at time \( t \), 
\( \gamma \) is the discount factor, 
and \( \alpha \) is a temperature parameter that determines the trade-off between reward maximization and policy entropy. The term \( \mathcal{H}(\pi(\cdot|s_t)) \) represents the entropy of the policy at state \( s_t \).

SAC maintains two separate action-value (Q) networks, \( Q_{\theta_1} \) and \( Q_{\theta_2} \), to mitigate overestimation bias. Both networks are trained to minimize a soft Bellman residual:

\begin{align}
J_Q(\theta_i) 
= \, & \mathbb{E}_{(s_t, a_t, r_t, s_{t+1}) \sim \mathcal{D}} \Biggl[
    \frac{1}{2} \Bigl(
        Q_{\theta_i}(s_t, a_t) \notag \\
    & \quad - y^{\text{soft}}(r_t, s_{t+1}, a_{t+1}) 
    \Bigr)^2 
\Biggr], 
\quad i = 1, 2
\label{eq:sac_critic_loss}
\end{align}

where \( \mathcal{D} \) is a replay buffer of past transitions and

\begin{align}
y^{\text{soft}}(r_t, s_{t+1}, a_{t+1}) = r_t + \gamma \Big(
& \min_{j \in \{1,2\}} Q_{\bar{\theta}_j}(s_{t+1}, a_{t+1}) \notag \\
& - \alpha \log \pi_\phi(a_{t+1} \mid s_{t+1}) \Big).
\label{eq:soft_target}
\end{align}

Here, \( \bar{\theta}_j \) indicates the parameters of a target network for the corresponding \( Q \)-function, and \( a_{t+1} \) is sampled from the current policy \( \pi_\phi(\cdot \mid s_{t+1}) \).
The policy \( \pi_\phi(a_t \mid s_t) \) is typically parameterized as a Gaussian distribution whose mean and covariance are outputs of a neural network. The actor (policy) is updated by minimizing:

\begin{align}
J_\pi(\phi) = \mathbb{E}_{s_t \sim \mathcal{D}} \biggl[
    \mathbb{E}_{a_t \sim \pi_\phi(\cdot \mid s_t)} \Bigl(
    & \alpha \log \pi_\phi(a_t \mid s_t) \notag \\
    & - \min_{i \in \{1,2\}} Q_{\theta_i}(s_t, a_t)
    \Bigr)
\biggr].
\label{eq:sac_actor}
\end{align}

Intuitively, this objective encourages the policy to select actions that maximize the action-value function, while also penalizing low-entropy policies through the \( \alpha \log \pi_\phi \) term.
The temperature parameter \(\alpha\) balances the reward maximization term against the policy entropy term. Instead of fixing \(\alpha\), SAC often employs an \emph{automatic tuning} mechanism, which optimizes \(\alpha\) by solving:

\begin{equation}
    J(\alpha) = \mathbb{E}_{a_t \sim \pi_\phi(\cdot \mid s_t)} 
    \left[ -\alpha \left( \log \pi_\phi(a_t \mid s_t) + \bar{\mathcal{H}} \right)\right],
\label{eq:sac_entropy_loss}
\end{equation}

where \(\bar{\mathcal{H}}\) is a target entropy, typically chosen based on the action space dimension.

A wide range of studies have employed or extended SAC to enhance decision-making and control for autonomous driving. For instance, hierarchical extensions of SAC have been investigated, with DDPG variants included for baseline comparisons \cite{wang_vision-based_2023}. Wei et al. \cite{wei_continual_2023} introduced a continual RL framework based on SAC—referred to as EM-SAC—that aims to improve both adaptability and continual learning capabilities in autonomous driving decision and control systems.

Additionally, imitation learning methods, such as Behavior Cloning (BC) \cite{pomerleau1991efficient}, GAIL \cite{ho2016generative}, InfoGAIL \cite{li2017infogail}, and MAGAIL \cite{song2018multi}, have been integrated with traffic simulators to train social vehicles. In this context, Zhu et al. \cite{zhu_rita_2023} introduced the RITA traffic environment and demonstrated how an ego vehicle can be trained directly using SAC or GAIL, subsequently benchmarking these RL policies within the RITA-generated traffic environment. Other researchers have leveraged SAC specifically as a model-free, off-policy RL algorithm to address online decision-making tasks \cite{savari_online_2021}. To boost data efficiency, Wu et al. \cite{wu_reinforcement_2023} augmented SAC with Expert Demonstrations Augmentation (EDA) and Mixed Priority Sampling (MPS). Aghdasian et al. \cite{aghdasian_autonomous_2023} further examined sensor-, image-, and fusion-based observation modalities under SAC-driven policies.

Beyond these approaches, hybrid or augmented architectures for SAC have also been proposed. For example, Huang et al. \cite{huang_simoun_2023} developed Simoun, built on top of SAC and enriched with a dual-path network architecture (capturing motion and appearance) as well as a consistency-guided curiosity module. Focused Experience Replay (FER), introduced as a method to improve stability and learning speed, has been combined with SAC (principally SACv2) and shown to be adaptable to DDPG and TD3 \cite{Enhanced_Off-Policy_Reinforcement_Learning_With_Focused_Experience_Replay}. Other lines of work further illustrate the versatility of SAC, whether integrated into custom reward functions \cite{An_Integrated_Reward_Function} or applied to specialized driving scenarios such as autonomous emergency vehicle operation \cite{Urban_Autonomous_Driving_of_Emergency_Vehicles_with_Reinforcement_Learning}. Han and Yilmaz \cite{hanLearningDriveUsing2022} proposed a hybrid control strategy, SIRL, that combines a sparse expert policy with an SAC-based reinforcement learner using Bayesian Controller Fusion to enable safe exploration and accelerated training in urban driving scenarios. Igoe et al. \cite{igoeMultiAlphaSoftActorCritic2023} introduced Multi-Alpha Soft Actor-Critic (MAS), which treats the entropy coefficient as a learnable distribution to mitigate the stochastic bias inherent in SAC. Evaluated in the CARLA simulator, MAS demonstrated more stable and less overly cautious driving behaviors compared to standard SAC, while preserving sample efficiency and policy robustness. 
Gupta et al. \cite{gupta_hylear_2023} introduces NavSAC, a tailored Soft Actor–Critic agent within HyLEAR, encodes a $400\times400$ intention image plus reward, speed, and prior action, then branches into $V_{\psi}$, $Q_{\theta}$, and policy $\pi_{\phi}$ over \{Accelerate, Maintain, Decelerate\}.  
Off‑policy training with entropy‑regularised SAC losses uses a replay buffer periodically infused with planner demonstrations, fusing imitation and exploration for faster, stable learning.

\subsubsection{Twin Delayed Deep Deterministic Policy Gradient}
 
Twin Delayed Deep Deterministic Policy Gradient (TD3) \cite{fujimoto_addressing_2018} is designed for continuous control and addresses the overestimation bias of Q-values often observed in actor-critic methods like DDPG. TD3 tackles the Q‑value overestimation that plagues DDPG by (1) training two independent critics and backing up the minimum of their estimates, (2) delaying actor updates so the policy sees more accurate values, and (3) adding clipped Gaussian noise to target actions to smooth the critics’ learning targets.
These three tweaks combine to yield much stabler training and higher returns on standard continuous‑control benchmarks

To mitigate overestimation, TD3 takes the \emph{minimum} of the two target critics when computing the TD target. First, add small clipped noise to the target action:

\begin{equation}
\tilde{a} = \pi_{\phi'}(s') + \epsilon, \quad
\epsilon \sim \text{clip}\left(\mathcal{N}(0,\sigma), -c, c\right).
\label{eq:target_policy_noise}
\end{equation}

The target value for a sampled transition \((s, a, r, s')\) is then
\begin{equation}
y = r + \gamma \min\left( Q_{\theta_1'}(s', \tilde{a}), Q_{\theta_2'}(s', \tilde{a}) \right).
\label{eq:td3_target}
\end{equation}
Each critic network \(Q_{\theta_i}\) is updated by minimizing the mean-squared error between its predicted Q-value and the target \(y\):
\begin{equation}
L(\theta_i) = \frac{1}{N} \sum_{(s,a,r,s') \sim \mathcal{D}}
\left( Q_{\theta_i}(s, a) - y \right)^2,
\label{eq:td3_critic_loss}
\end{equation}

where \(N\) is the minibatch size. Both critics are updated at every time step.

The actor network \( \pi_\phi \) is updated less frequently (e.g., once for every two updates of the critics). The gradient for the actor is computed by:

\begin{equation}
\nabla_{\phi} J \approx \frac{1}{N} \sum_{s \sim \mathcal{D}}
\nabla_{a} Q_{\theta_1}(s, a)\bigg|_{a = \pi_{\phi}(s)} \;
\nabla_{\phi} \pi_{\phi}(s). 
\label{eq:td3_actor_gradient}
\end{equation}

After each set of updates, the target networks are softly updated as:

\begin{equation}
\theta'_i \leftarrow \tau\,\theta_i + (1 - \tau)\,\theta'_i, \quad
\phi' \leftarrow \tau\,\phi + (1 - \tau)\,\phi',
\label{eq:td3_soft_update}
\end{equation}

where \( \tau \ll 1 \) is the target update rate (e.g., \( \tau = 0.005 \)).

Several recent studies have enhanced and applied the TD3 algorithm to address safety, exploration, and performance in autonomous driving tasks. Jia et al. \cite{jia_safe_2023} proposed an improved TD3 variant that enforces safety constraints via a Lagrangian formulation, ensuring that the control policy respects prescribed safety limits. In another work, Xu et al. \cite{xu_end--end_2023} introduced Pri-TD3, which builds upon the TD3 framework with Prioritized Experience Replay (PER) to mitigate Q-value overestimation and reduce policy variance more effectively than simpler methods such as DDPG.

Liao et al. \cite{liao_lateral_2023} employed TD3 for real-time, safe, and comfortable lateral control, focusing on collision avoidance strategies under realistic driving conditions. Furthermore, other researchers have explored noise injection into the actor outputs during training to promote more robust exploration and improve policy convergence \cite{Deep_Reinforcement_Learning_for_Autonomous_Vehicle_Intersection_Navigation}. In the domain of intersection navigation, TD3 has been combined with a Local Goal Velocity (LGV) mechanism and further augmented with PER to refine the learning process and expedite the development of effective driving policies \cite{Deep-Reinforcement-Learning-Based_Driving_Policy_at_Intersections}. Li \textit{et~al.}\,\cite{li_decision-making_2024} extend TD3 with a local‐attention safety architecture in which the actor network is augmented by an {ego‐attention module.  
This module dynamically identifies socially critical surrounding vehicles, enabling the policy to focus its continuous control outputs on the most influential interactions. To further promote collision avoidance, the authors design a custom safety-reward that penalises proximity violations while still rewarding efficient progress. The resulting algorithm retains TD3’s twin critic networks and delayed policy updates, yet the attention mechanism and safety‐aware shaping together yield a policy that is both risk‐averse and time-efficient in dense traffic scenarios.

\subsubsection{New Methods}

There are certain new RL-based methods introduced in the literature. For instance, khalil et al. \cite{khalil_exploiting_2023} leverages SAC's entropy-regularized objective to encourage exploration. SAC trains a policy (actor) and value function (critic) jointly in an off-policy manner by sampling from a replay buffer and updating parameters using Q-value targets .

In a different vein, Qi et al. \cite{Cognitive_Reinforcement_Learning} introduces a new framework merges MFRL with human cognitive processes to enhance decision-making. Specifically, the framework replaces conventional neural network decision-making modules with a cognitive model that follows the “Perception–Memory–Reasoning–Action” cycle, thereby infusing human-like reasoning loops into the learning process. To ensure transparency and interpretability, predefined production rules govern the decision-making steps of this cognitive model .

Couto et al. \cite{coutoHierarchicalGenerativeAdversarial2024} proposed hGAIL, a hierarchical imitation learning framework that integrates Generative Adversarial Imitation Learning (GAIL) with Proximal Policy Optimization (PPO) for policy training. This model-free, on-policy approach enables more structured policy learning by combining adversarial imitation with hierarchical decision-making.

Jaafra et al. \cite{jaafraContextAwareAutonomousDriving2019} proposed a context-aware autonomous driving framework that leverages meta-RL for rapid policy adaptation in dynamic urban environments. Their method combines a gradient-based meta-learning approach—using a Model-Agnostic Meta-Learning (MAML) strategy—to train a meta-policy capable of quickly adapting to new driving scenarios with minimal additional data.

Baheri et~al \cite{baheri_vision-based_2020} depart from value- and policy-gradient paradigms by adopting an evolutionary RL approach based on the Covariance Matrix Adaptation Evolution Strategy. CMA–ES maintains a population of candidate controllers whose parameter vectors are sampled from a multivariate normal distribution; after roll-outs, the distribution’s mean and covariance are iteratively updated to favour higher-performing individuals. By foregoing gradient information altogether, this black-box optimisation scheme sidesteps issues of non-differentiability and sparse rewards, enabling direct search in the controller’s parameter space while gradually adapting exploration to the covariance structure of successful policies.

Moreover, distributed off-policy RL has been employed in several recent approaches. These methods incorporate elements such as clipped double Q-learning from TD3, as well as the maximum entropy concept from SAC, to refine learning stability and exploration \cite{Versatile_and_Efficien}.

\subsubsection{Comparitive Studies}
Some studies consider different methods in RL and conducted comparative studies. For instance, a variety of test scenarios have been constructed in recent studies to evaluate learned models rigorously and to provide standardized benchmark results under well-defined evaluation metrics \cite{wang_benchmarking_2021}. One such learning framework partitions the overall driving task into distinct subtasks—lateral control (lane keeping and lane changing), longitudinal control, and decision-making—to address the multi-faceted nature of autonomous driving. Continuous-action subtasks (lane keeping, lane changing, and longitudinal control) are trained using an Actor-Critic approach (specifically, DDPG), while discrete decision-making tasks employ a DQN. To balance exploration and exploitation, multi-head actor networks introduce adaptive noise in the action selection mechanism \cite{A_Decision_Control_Method}.

Comparative evaluations of D3QN, A2C, and PPO and their variants have also been reported in intersection scenarios, where decision-making complexities often increase \cite{xu_decision-making_2022}. In another line of research, a child’s ride-on toy car was redesigned to serve as a self-driving platform; this setup was employed to compare the performance of three RL models (DQN, DDQN, and D3DQN) under real-world constraints \cite{manikandan_ad_2023}. Further expanding on efficiency concerns, Frauenknecht et al. \cite{frauenknecht_data-efficient_2023} investigated data-efficient RL approaches for vehicle control, highlighting the importance of reducing sample complexity and computational overhead without sacrificing policy performance. They compare SAC, PETS-MPPI, MBPO, REDQ for their work.

\subsection{Model-Based Reinforcement Learning}

Model-based RL (MBRL) uses a learned or predefined model of the environment to guide policy optimization. Instead of relying exclusively on repeated interaction with the real environment, the agent maintains an approximate transition and reward function, often denoted by

\begin{equation}
\hat{f}_\theta(s,a) \approx s', \quad
\hat{r}_\theta(s,a) \approx r,
\label{eq:world_model_approx}
\end{equation}

where \( \theta \) are the model parameters, \( s \in \mathcal{S} \) is the current state, \( a \in \mathcal{A} \) is the action, \( r \) is the immediate reward, and \( s' \) is the next state. These approximations can be learned by minimizing prediction errors across a dataset of real transitions collected from environment interactions:

\begin{align}
\mathcal{L}(\theta) 
= \sum_{(s,a,s',r) \in \mathcal{D}} \Big\{
& \big\| \hat{f}_\theta(s,a) - s' \big\|^2 \notag \\
& + \big\| \hat{r}_\theta(s,a) - r \big\|^2
\Big\}.
\label{eq:world_model_loss}
\end{align}

Once the model is learned, it can be used to \emph{simulate} experience without requiring costly real-world rollouts. The agent can perform imaginary rollouts by repeatedly applying the learned transition and reward functions:

\begin{equation}
s_{t+1}^{(sim)} = \hat{f}_\theta(s_{t}^{(sim)}, a_{t}^{(sim)}), \quad
r_{t+1}^{(sim)} = \hat{r}_\theta(s_{t}^{(sim)}, a_{t}^{(sim)}),
\label{eq:world_model_rollout}
\end{equation}

These simulated transitions are then used to evaluate or improve the policy. For instance, if the agent uses a parameterized policy \(\pi_\phi(a \mid s)\), it can be improved by maximizing expected returns in simulation. In a policy-gradient framework, this might involve updating \(\phi\) via

\begin{equation}
\nabla_{\phi} J 
\approx 
\frac{1}{N} \sum 
\nabla_{\phi} \log \pi_\phi(a_t \mid s_t)\, G_t,
\label{eq:reinforce_gradient}
\end{equation}

where \(G_t\) is an estimate of the return from simulated or real data.
A popular strategy is to alternate between collecting real data and using the learned model for planning or imagination-based updates. This idea, originally introduced in the \emph{Dyna} framework \cite{sutton_dyna_1991}, typically follows these steps: gather transitions \((s,a,s',r)\) from the real environment, update the model \(\hat{f}_\theta, \hat{r}_\theta\) using this data, and then generate additional ``imagined'' transitions by rolling out the updated model to refine the policy or value function.

Although MBRL can dramatically improve sample efficiency, it also introduces potential model bias: if the learned model is inaccurate in certain state-action regions, planning on this incorrect model can lead to suboptimal or destabilizing actions. Addressing this challenge often involves estimating uncertainty (e.g., with ensemble models) or restricting simulation rollouts to shorter horizons where the model remains more reliable. In practice, successful methods like PETS (Probabilistic Ensembles with Trajectory Sampling) \cite{chua_deep_2018}, MBPO (Model-Based Policy Optimization) \cite{janner_when_2021}, or Dreamer \cite{hafner_learning_2019} combine these ideas to balance the benefits of high sample efficiency with robust policy learning.

MBRL techniques have recently garnered increasing attention for autonomous driving applications, as they aim to leverage internal world models to improve sample efficiency and reduce the need for risky real-world exploration. For instance, DreamerV3 \cite{hafner_mastering_2024} has been utilized to learn a latent representation of the driving environment, with the policy trained within this learned latent space for enhanced efficiency \cite{Think2Drive}. Expanding on the Dreamer paradigm, an approach called Iso-Dream++ decouples controllable and noncontrollable dynamics in the latent space, incorporating min-max variance constraints to avert training collapse and a sparse dependency mechanism to capture indirect influences between these decoupled states. This architecture generates latent imaginations from isolated controllable and noncontrollable factors, which guide the policy learning process \cite{Model-Based_Reinforcement_Learning_with_Isolated_Imaginations}.

Deep Imitative Reinforcement Learning (DIRL) represents another variant that combines imitation learning (IL) with a MBRL framework, using a predictive world model—Reveries-net—to simulate intricate interactions and refine driving policies in an offline fashion \cite{Vision-Based_Autonomous_Car_Racing_Using_Deep_Imitative_Reinforcement_Learning}. In a further illustration of the benefits of MBRL, a “world-on-rails” methodology leverages pre-recorded trajectories and treats the environment as non-responsive to ego-vehicle actions. This simplifies the modeling process, enabling the use of a forward model to predict vehicle dynamics while action-value functions are computed via tabular Bellman updates. The resulting action-value functions then supervise a visuomotor policy through distillation, effectively sidestepping the need for direct exploration or the risk of executing unsafe policies \cite{Learning_to_drive_from_a_world_on_rails}. By integrating world modeling with dense offline supervision, such approaches demonstrate that MBRL can be both sample efficient and robust for autonomous driving tasks.

%\subsection{Other Methods - Elahe}

\subsection{Hybrid RL (Combination of MBRL and MFRL)}

Recent advancements have explored synergistic integrations of MBRL and MFRL to enhance both safety and performance in autonomous driving systems. One approach leverages a model-free constraint—captured via distributional RL in the form of a cost critic network to quantify risk—alongside a model-based High-Order Control Barrier Function (HOCBF) to enforce vehicle-dynamics safety constraints. The proposed framework employs Lagrangian optimization to update the policy under these constraints and integrates imitation learning during early training phases for improved stability \cite{Autonomous_Driving_via_Knowledge-Enhanced}.

Another example, termed Deductive Reinforcement Learning (DeRL), merges DDPG with a model-based “Deduction Reasoner” (DR). The DR utilizes a learned environment model to project future states and rewards, providing a “self-assessment value” that guides policy improvements. Additionally, a Semantic Encoder Module (SEM) extracts low-dimensional, robust representations, thereby reinforcing both interpretability and efficacy \cite{Deductive_Reinforcement_Learning}.

In parallel, research has also combined a DQN agent with the classical, model-based A* path planner. Here, the A* algorithm supplies waypoints, and the DRL agent learns to follow these waypoints while actively avoiding collisions. This hybrid solution underscores the importance of integrating established planning algorithms with learning-based modules to bolster reliability and safety \cite{Integrating_Deep_Reinforcement_Learning_with_Model-based_Path_Planners_for_Automated_Driving}.

\subsection{Alternative RL Methods}

While methods like DQN, PPO, and SAC are frequently used for autonomous driving in CARLA, a number of less-explored but promising RL approaches have also been applied. These methods offer alternative perspectives on decision-making, uncertainty handling, and policy structure. Below, we provide a brief overview of these methods, citing their foundational work as well as their application in the CARLA simulator.

\subsubsection{Bayesian Reinforcement Learning}
Bayesian RL (BRL) maintains a posterior distribution over environment dynamics or value functions, allowing for uncertainty-aware exploration. A comprehensive survey of BRL techniques was presented by Ghavamzadeh et al.~\cite{ghavamzadeh2015bayesian}. In the CARLA context, BRL was used by Gharaee et al. ~\cite{A_Bayesian_Approach_to_Reinforcement_Learning_of_Vision-Based_Vehicular_Control}, where Gaussian Mixture Models (GMMs) were employed to cluster perceptual states and model action probabilities. This approach combined GMM-based modeling with temporal difference learning, enabling adaptive policy updates based on observation similarity.

\subsubsection{Hierarchical Reinforcement Learning}
Hierarchical Reinforcement Learning (HRL) decomposes complex tasks into hierarchies of subpolicies. The options framework introduced by Sutton et al.~\cite{sutton1999between} laid the groundwork for HRL by enabling temporal abstractions in RL. In the work by Li et al.~\cite{A_Hierarchical_Autonomous_Driving_Framework}, the authors introduced a hierarchical framework that integrates Imitation Learning (IL) and RL. IL is used to manage low-level control tasks, such as trajectory following, while RL is responsible for high-level decision-making, including navigation and collision avoidance. The framework first learns from expert demonstrations and then refines its policy through RL, enabling more robust behavior in complex and dynamic driving environments. Separately, Wang et al. ~\cite{Hierarchical_Reinforcement_Learning_with_Successor_Representation_for_Intelligent_Vehicle_Collision_Avoidance_of_Dynamic_Pedestrian} adopted a two-level HRL structure designed for collision avoidance involving dynamic pedestrians. The upper layer selects from a set of Decision Primitives (DPs) based on a Predictive Risk Map (PRM), while the lower layer executes the chosen primitive through detailed control over steering, acceleration, and braking. This design allowed the agent to make safe and context-sensitive decisions in highly dynamic environments. He et al. \cite{heAdaptiveDecisionMaking2022} proposed a HRL framework for autonomous driving, where the upper decision-making layer selects motion primitives using a Double DQN, and the lower execution layer applies DDPG to realize continuous control. This architecture allows for skill reuse across scenarios and improves generalization by decomposing high-level decisions from low-level controls. 
Ben Naveed et al. \cite{naveedTrajectoryPlanningAutonomous2020} proposed a HRL framework for trajectory planning in autonomous driving, combining a high-level policy for maneuver selection (lane follow/wait vs. lane change) with low-level planners for trajectory generation. A PID controller was used to track waypoints instead of directly applying low-level controls, ensuring smooth and safe execution. Gangopadhyay et al. \cite{gangopadhyayHierarchicalProgramTriggeredReinforcement2022} introduced a Hierarchical Program-Triggered RL (HPRL) framework that combines symbolic programming with modular RL agents for autonomous driving. Each agent is trained for a specific maneuver (e.g., lane change, turn, straight drive) using DQN or DDPG, while a structured program governs task execution and enforces safety via embedded assertions. This design improves interpretability and verifiability by isolating learning to simple sub-tasks and validating the high-level controller with formal methods.

\subsubsection{Maximum Entropy Reinforcement Learning}
Maximum Entropy Reinforcement Learning (MaxEnt RL) encourages policies that maximize both reward and entropy, promoting exploration and robustness. The principle was originally introduced in the context of inverse RL by Ziebart~\cite{ziebart2010modeling}, and later extended to RL through energy-based policies by Haarnoja et al.~\cite{haarnoja2017reinforcement}. Khalil et al.~\cite{Integration_of_Motion_Prediction_with_End-to-end_Latent_RL_for_Self-Driving_Vehicles} proposed a Sequential Latent MaxEnt RL framework for use in CARLA. This approach combined latent-space RL with motion prediction via a probabilistic graphical model, enhancing decision-making in dynamic driving scenarios.

\subsubsection{Distributional Reinforcement Learning.}
Distributional RL learns the full distribution of returns \( Z(s,a) \) rather than only their expectation \( Q(s,a) \). This approach was first introduced by Bellemare et al.~\cite{bellemare2017distributional} and further refined through the Fully Parameterized Quantile Function (FQF) by Yang et al.~\cite{yang2019fully}. Chen et al. \cite{chen_motion_2024} applied FQF in CARLA to learn separate policies for path and speed planning, allowing the model to better handle uncertainty due to occlusions and unpredictable pedestrian behavior.

\subsubsection{Skill-Based Hierarchical Offline Reinforcement Learning}
Skill-based RL allows high-level policies to select from learned low-level behaviors. In offline settings, this approach has been explored by Sharma et al.~\cite{sharma2020emergent}, who proposed latent skill discovery in multi-task settings. Li et al. \cite{Boosting_Offline_Reinforcement_Learning} adapted this concept to CARLA by training a high-level planner using a two-branch Variational Autoencoder (VAE) that captured both discrete and continuous driving skills, while the low-level controller executed steering and throttle commands accordingly.

%%%%%%%%%%%%%%%%%%%%%%%%%%%
%%%%%%%%%%%%%%%%%%%%%%%%%%%
\begin{table}[t]
\centering
\caption{Key CARLA publications by RL method.}
\label{tab:method_usage}
%\resizebox{\columnwidth}{!}{%
  \begin{tabular}{l p{5cm}}
\toprule
\textbf{Category / Method} & \textbf{Key CARLA Papers / Variants} \\ 
\midrule

\textbf{MFRL} & \\
\quad Actor–Critic & \cite{chronis_driving_2021,udatha_reinforcement_2023} \\
\quad DQN & \cite{zhang_toward_2022,yang_decision-making_2022,bai_fen-dqn_2023,
muhammed_developing_2021,elallid_dqn-based_2022,GRI:General_Reinforced_Imitation,
End-to-End_Model-Free_Reinforcement_Learning_for_Urban_Driving,
Safe_Navigation_Based_on_Deep_Q-Network_Algorithm_Using,
Deep_Reinforcement_Learning_Control_Strategy_at_Roundabout_for_i-CAR,
A_Reinforcement_Learning_Based_Approach_for_Controlling_Autonomous_Vehicles_in_Complex_Scenarios,
Using_the_CARLA_Simulator_to_Train_A_Deep_Q_Self-Driving,
A_Deep_Q-Network_Reinforcement_Learning-Based,
Reinforcement_Learning-Based_Guidance_of_Autonomous_Vehicles,
Learning_Automated_Driving_in_Complex_Intersection_Scenarios_Based,
A_Methodology_Based_on_Deep_Reinforcement_Learning,frijiDQNBasedAutonomousCarFollowing2020,
hishmehDeerHeadlightsShort2020,221016567DeFIXDetecting,deshpande_navigation_2021,
noauthor_explainable_nodate} \\
\quad DQL & \cite{cheng_longitudinal_2021,hussonnois_end--end_2022} \\
\quad PPO & \cite{carton_using_2021,mohammed_unified_2024,deng_context_2023,
deng_context-aware_2024,zhao_end--end_2024,wu_proximal_2023,
End-to-End_Urban_Driving_by_Imitating_a_Reinforcement_Learning_Coach,
Learning_Urban_Driving_Policies_using_Deep_Reinforcement_Learning,
Efficient_Learning_of_Urban_Driving,Domain_Adaptation_In_Reinforcement_Learning,
Reinforced_Curriculum_Learning_For_Autonomous_Driving,
Addressing_Lane_Keeping_and_Intersections,albilani_guided_2023,
martinez_gomez_temporal_2023,jin_vwpefficient_2024,shi_efficient_2023} \\
\quad TRPO & \cite{gutierrez-morenoDecisionMakingAutonomous2024} \\
\quad DDPG & \cite{wu_lane_2022,goel_adaptive_2021,youssef_deep_2019,
peng_imitative_2020,Deep_reinforcement_learning_based_control_for_Autonomous_Vehicles_in_CARLA,
Decision_Making_for_Autonomous_Driving_Via_Multimodal_Transforme,Towards_Autonomous_Driving,
zhangSelflearningLaneKeeping2021,tsaiAutonomousVehicleFollowingTechnique2023,
doe_dsorl_2023,li_dynamic_2022,ahmed_policy-based_2022} \\
\quad SAC & \cite{wang_vision-based_2023,wei_continual_2023,zhu_rita_2023,
savari_online_2021,wu_reinforcement_2023,aghdasian_autonomous_2023,
huang_simoun_2023,Enhanced_Off-Policy_Reinforcement_Learning_With_Focused_Experience_Replay,
An_Integrated_Reward_Function,Urban_Autonomous_Driving_of_Emergency_Vehicles_with_Reinforcement_Learning,
hanLearningDriveUsing2022,igoeMultiAlphaSoftActorCritic2023,gupta_hylear_2023} \\
\quad TD3 & \cite{jia_safe_2023,xu_end--end_2023,liao_lateral_2023,
Deep_Reinforcement_Learning_for_Autonomous_Vehicle_Intersection_Navigation,
Deep-Reinforcement-Learning-Based_Driving_Policy_at_Intersections,
li_decision-making_2024} \\

\midrule
\textbf{MBRL} & \cite{Think2Drive,Model-Based_Reinforcement_Learning_with_Isolated_Imaginations,
Vision-Based_Autonomous_Car_Racing_Using_Deep_Imitative_Reinforcement_Learning,
Learning_to_drive_from_a_world_on_rails} \\

\midrule
\textbf{Hybrid RL} & \\
\quad Model-Based + Model-Free & \cite{Autonomous_Driving_via_Knowledge-Enhanced,
Deductive_Reinforcement_Learning,Integrating_Deep_Reinforcement_Learning_with_Model-based_Path_Planners_for_Automated_Driving} \\

\midrule
\textbf{Alternative / Specialized RL Methods} & \\
\quad Bayesian RL & \cite{A_Bayesian_Approach_to_Reinforcement_Learning_of_Vision-Based_Vehicular_Control} \\
\quad Hierarchical RL & \cite{A_Hierarchical_Autonomous_Driving_Framework,
Hierarchical_Reinforcement_Learning_with_Successor_Representation_for_Intelligent_Vehicle_Collision_Avoidance_of_Dynamic_Pedestrian,
heAdaptiveDecisionMaking2022,naveedTrajectoryPlanningAutonomous2020,
gangopadhyayHierarchicalProgramTriggeredReinforcement2022} \\
\quad Distributional RL & \cite{chen_motion_2024} \\
\quad Maximum Entropy RL & \cite{Integration_of_Motion_Prediction_with_End-to-end_Latent_RL_for_Self-Driving_Vehicles} \\
\quad Skill-Based Hierarchical Offline RL & \cite{Boosting_Offline_Reinforcement_Learning} \\
\quad New / Evolutionary Methods & \cite{khalil_exploiting_2023,Cognitive_Reinforcement_Learning,
Versatile_and_Efficien,coutoHierarchicalGenerativeAdversarial2024,
jaafraContextAwareAutonomousDriving2019,baheri_vision-based_2020} \\

\midrule
\textbf{Comparative Studies} & \cite{wang_benchmarking_2021,A_Decision_Control_Method,
xu_decision-making_2022,manikandan_ad_2023,frauenknecht_data-efficient_2023} \\

\bottomrule
\end{tabular}%

\end{table}

\section{State Space Representations}
In RL for autonomous driving, the design of the observation space plays a critical role in shaping the agent's perception and decision-making capabilities. CARLA provides a rich set of sensor modalities—such as RGB cameras, LiDAR, semantic segmentation, and direct vehicle state information—enabling diverse representations of the environment. Researchers often choose or combine these modalities based on task complexity, computational constraints, and the desired level of abstraction. This section surveys commonly used observation spaces in CARLA-based RL studies, grouped by their sensor types and fusion strategies, and discusses how each representation affects policy learning and generalization.

\subsection{Front Facing RGB Images}

Several studies adopt a single-stream visual input as the core observation. For instance, Elallid et al.~\cite{elallid_dqn-based_2022} use a front-facing grayscale image, initially sized at $640 \times 480$ but cropped and resized to $192 \times 256$ to reduce irrelevant background. This preprocessed image is then flattened to feed into a Deep Q-Network (DQN). Zhang et al.~\cite{zhang_toward_2022} also rely heavily on pixel observations, with lane-deviation and collision indicators influencing only the reward function rather than the agent’s direct observation space. Liu et al. ~\cite{A_Methodology_Based_on_Deep_Reinforcement_Learning},  uses end-to-end vision-based learning that processes standard front-facing RGB images with lane invasion and collision dectector sensor. Meanwhile, Ahmad et al. ~\cite{A_Deep_Q-Network_Reinforcement_Learning-Based} fuses an RGB image ($400 \times 400 \times 3$) with vehicle speed and the angle from the road center, thereby adding minimal numeric inputs to anchor visual perception. Although purely vision-based designs can suffice for lane-following tasks, many researchers find that incorporating at least some numeric data—such as speed or steering angle—improves training stability and convergence rates. Xing et al.  \cite{Domain_Adaptation_In_Reinforcement_Learning} uses a raw RGB front facing image for Latent Unified State Representation (LUSR) that separates domain-general features (e.g., vehicle dynamics) from domain-specific variations like weather, thus making the agent’s policy more transferrable. In a model-based setting, Pan et al. \cite{Model-Based_Reinforcement_Learning_with_Isolated_Imaginations} decouples the state into controllable (ego-vehicle) and noncontrollable (other vehicles) factors from raw RGB front facing camera frame, enhancing planning accuracy.

Other works emphasize the incorporation of kinematic or command data to provide richer contextual cues. Carton et al.~\cite{carton_using_2021} employ a front-facing RGB camera at around $192 \times 64$ resolution and augment it with vehicle speed and steering angle. Mohammed et al.~\cite{mohammed_unified_2024} propose fusing a sequence of front-facing RGB frames, extracted via a pre-trained CNN, with current speed and route speed limits, either in early convolutional layers or at the final fully connected layers. By integrating control signals into the state, Wu et al.~\cite{wu_proximal_2023} provide the agent with steering, throttle, and brake values that are concatenated with a 64-dimensional feature embedding from a front-facing RGB feed ($160 \times 80 \times 3$). Similarly, Zhao et al.~\cite{zhao_end--end_2024} compress RGB frame using a Variational Autoencoder (VAE) and combine these latent features with vehicle speed, throttle, orientation, distance from lane center, waypoint and previous steering angle, yielding a compact yet information-rich input. Peng et al.~\cite{peng_imitative_2020} go a step further by including a Pure Pursuit (PP) steering angle in addition to speed and high-level route commands, demonstrating how geometric planning cues can guide visual learning to achieve smoother control policies.

Dimensionality reduction strategies frequently appear to manage the high computational demands of processing raw images. Savari et al.~\cite{savari_online_2021} employ an auto-encoder to generate a latent representation from the front camera view before feeding it into the Soft Actor-Critic (SAC) algorithm. Deductive RL~\cite{Deductive_Reinforcement_Learning} takes an alternative approach by giving the actor a reduced observation (monocular front-facing camera + forward speed), while the critic receives a privileged state from CARLA (e.g., distances to pedestrians or vehicles). This allows the actor to learn from a realistic sensor setup, but the critic’s extra information stabilizes training. Meanwhile, Zhao et al.~\cite{zhao_real-time_2021} leverage a Bird’s Eye View (BEV) semantic map only during training as an auxiliary supervision signal, relying solely on the front-facing ResNet50-processed RGB images plus high-level navigational commands at inference, thus ensuring real-time viability.

Temporal stacking of frames is also commonly leveraged for better motion awareness. Huang et al. ~\cite{huang_simoun_2023} Stacks a sequence of recent frames, raw image frames processed by a dual-path CNN: one path learns appearance features per frame, the other learns motion features from frame-to-frame changes. Elallid et al. ~\cite{Deep_Reinforcement_Learning_for_Autonomous_Vehicle_Intersection_Navigation} provides the agent with four consecutive grayscale images ($84 \times 84$) to capture short-term environment changes critical for intersection decisions. Elallid et al. ~\cite{A_Reinforcement_Learning_Based_Approach_for_Controlling_Autonomous_Vehicles_in_Complex_Scenarios} similarly stacks four RGB frames (downsampled to $144 \times 144$ grayscale) and appends a 2D goal vector representing the target location. In racing contexts, Cai et al. ~\cite{Vision-Based_Autonomous_Car_Racing_Using_Deep_Imitative_Reinforcement_Learning} uses four-frame stacks of $96 \times 96 \times 3$ inputs plus vehicle speed, exploiting temporal cues to execute rapid maneuvers. Li et al. ~\cite{Learning_Automated_Driving_in_Complex_Intersection_Scenarios_Based} processes two consecutive monocular images (captured at times $t$ and $t-2$) to estimate ego-vehicle motion in intersections. In a somewhat different direction, Chen et al. ~\cite{Towards_Autonomous_Driving} encodes the front camera ($64 \times 64 \times 3$) with a self-attention network and fuses additional sensor data (speed, acceleration, position), shifting focus toward efficient multi-modal integration. Finally, Li et al. ~\cite{A_Hierarchical_Autonomous_Driving_Framework} demonstrates how feature extraction with ResNet, combined with vehicle speed and high-level navigation commands (e.g., Turn Right, Turn Left, Go Straight), allows for hierarchical policy learning that separates strategic decision-making from direct motor control. Meanwhile, Youssef et al.~\cite{youssef_deep_2019} also utilize front-facing images for feature extraction but place special emphasis on high-level route instructions (turn left, turn right, or go straight), integrated through fully connected layers or the final actor network.

%%%%%%%%%%%%%%%%%%%%%%%%%%%%%%%%%%%%%%%%%%%%%%%%%%%%%%%%%%%%%%%%

\subsection{Bird Eye View Images}

A growing number of studies have explored bird’s-eye view (BEV) representations to provide a structured, top-down depiction of the driving environment in CARLA. Such views typically display the ego vehicle’s surroundings—including lanes, vehicles, and traffic signals—in a spatially coherent format, facilitating more interpretable learning of road geometry and multi-agent interactions. For instance, Wang et al.~\cite{wang_benchmarking_2021} generate BEV images at resolutions of $64 \times 64$, placing the ego vehicle at the bottom center of the frame, with color-coding for ego-vehicle (blue), lanes (red) and surrounding vehicles (green). This strategy provides the RL agent with an easily parsable scene layout, improving its capacity for lane-changing and collision avoidance. Similarly, Deng et al.~\cite{deng_context_2023} enhance decision-making by integrating BEV inputs with a 6-dimensional measurement vector (speed, acceleration, steering angle, gear state) and a historical trajectory encoded via Gated Recurrent Units (GRU). This approach leverages short-term motion context to address partial observability issues, thereby promoting more consistent control in dynamic traffic scenarios.

In many BEV-based methods, the representation is augmented with additional high-level or kinematic data to capture the ego vehicle’s pose and local objectives. Deng et al.~\cite{deng_context-aware_2024} build upon their earlier work by blending BEV images with real-time vehicle measurements and historical trajectories. They employ recurrent architectures (GRU, LSTM, MLP, or Transformers) to encode past observations, actions, and rewards, consolidating them into a latent vector that is then fed to the RL policy. Historical data improves situational awareness in partially observable environments, leading to smoother handling of dynamic road users. LiDAR sensors can also reinforce the BEV perspective by supplying real-time object detection and depth cues. For example, Manikandan et al.~\cite{manikandan_ad_2023} combine LiDAR-based obstacle distances with camera perception and kinematic signals (velocity, steering angle, lane offset) to better detect and localize objects in all directions. Fused RGB-LiDAR point clouds or RGB-D images likewise help the agent maintain a detailed map of surrounding vehicles, enabling more informed decisions under complex traffic conditions.

Several other works implement specialized BEV semantic segmentation or multi-channel images to highlight distinct road features. In Li et al.~\cite{Think2Drive}, the state space includes semantic masks, bounding boxes, and traffic signal states, while Zhang et al.~\cite{End-to-End_Urban_Driving_by_Imitating_a_Reinforcement_Learning_Coach} encodes drivable areas, lane boundaries, and pedestrians in separate segmentation channels. Extended approach, such as Agarwal et al.~\cite{Learning_Urban_Driving_Policies_using_Deep_Reinforcement_Learning} combine BEV segmentation (vehicles, pedestrians, traffic lights) with waypoints from a global planner, along with kinematic data (speed, distance to goal, steering angle) to produce a richly detailed representation. Trumpp et al. ~\cite{Efficient_Learning_of_Urban_Driving} uses four stacked bev images into LSTM-based modules to capture temporal dependencies, further refining predictions and behavior modeling.

To handle highly dynamic scenarios, some methods integrate motion prediction or incorporate specialized LiDAR-based maps into the BEV. Khalil et al.~\cite{Integration_of_Motion_Prediction_with_End-to-end_Latent_RL_for_Self-Driving_Vehicles} uses two BEV images: one for LiDAR data (ground and above-ground point clouds with route waypoints) and another for predicted trajectories of surrounding vehicles extracted from MotionNet \cite{wu_motionnet_2020}. Likewise, Li et al.~\cite{Boosting_Offline_Reinforcement_Learning} retains a BEV semantic segmentation image plus a measurement vector, while Tong et al. ~\cite{Urban_Autonomous_Driving_of_Emergency_Vehicles_with_Reinforcement_Learning} \ adopts a $256 \times 256$ BEV image covering a 35\,m radius around the ego vehicle. This latter work encodes an ego-vehicle state (speed, angular gap to route, lateral distance, and junction indicator), ensuring that emergency-vehicle maneuvers are sensitive to both local geometry and mission-critical efficiency. Altogether, these BEV-centric frameworks demonstrate the benefits of structured, top-down map representations, especially when complemented by numeric vehicle states, temporal models, or motion prediction modules. They enable more interpretable and robust policy learning in multi-agent and urban driving environments, where spatial awareness and accurate trajectory planning are essential.

%%%%%%%%%%%%%%%%%%%%%%%%%%%%%%%%%%%%%%%%%%%%%%%%%%%%%%%%%%%%%%%%

\subsection{Semantic Segmentic Images}

Semantic segmentation has been widely adopted as a critical component of state representations in RL-based autonomous driving frameworks, often to encode meaningful spatial and categorical cues for decision-making. For instance, Wang et al.~\cite{Versatile_and_Efficien} integrates drivable area segmentation alongside vehicle speed, providing the RL agent with a clear delineation of navigable surfaces. In \cite{GRI:General_Reinforced_Imitation} Chekroun et al. used three RGB camera streams (center, left, right) are processed by pretrained encoders to extract segmentation features, traffic light status, and intersection presence over short temporal windows (stacking the frames), enabling the agent to capture both semantic and temporal information. Likewise, the pipeline in~\cite{End-to-End_Model-Free_Reinforcement_Learning_for_Urban_Driving} Toromanoff et al. employs a ResNet-18 pretrained on semantic segmentation data to encode visual frames and includes auxiliary signals such as traffic light states, intersection recognition, and lane positioning. These aggregated features are then stacked over time, furnishing the RL module with a richer temporal perspective.

In a more sensor-fused approach, the work in \cite{Safe_Navigation_Based_on_Deep_Q-Network_Algorithm_Using} Marouane et al. combines minimum distance to obstacles (obtained from depth and segmentation images), collision indicators, orientation angles, lateral distance metrics, and vehicle velocity, encapsulating both semantic and geometric factors critical to safe navigation. Going further, Gharaee et al. \cite{A_Bayesian_Approach_to_Reinforcement_Learning_of_Vision-Based_Vehicular_Control} adapted a beyesian perspective, where segmentation maps are segmented into six spatial regions (three vertical, two horizontal), and histograms of class distributions across five semantic categories— road, road-line, off-road, static objects, dynamic objects —are normalized and concatenated to form a compact yet informative state vector.

Other works concentrate on tailoring segmentation outputs to specific driving tasks. In \cite{Deep_Reinforcement_Learning_Control_Strategy_at_Roundabout_for_i-CAR} Muhtadin et al. used grayscale segmentation images (or advanced binary masks) filter out non-drivable terrain to simplify roundabout navigation policies. Finally, Silva et al. \cite{Addressing_Lane_Keeping_and_Intersections} processes front-facing semantic segmentation frames at 84 × 168 pixels and stacks four consecutive frames to account for temporal context. These semantic inputs are merged with vehicle measurements—such as throttle, steering angle, speed, and heading—to form a unified representation that is passed through a CNN-based encoder (inspired by IMPALA), enabling robust lane-keeping and intersection handling. Collectively, these works underscore the versatility of semantic segmentation in RL-based driving, as it provides structured, high-level features that enhance situational awareness and facilitate safer and more efficient decision-making.

%%%%%%%%%%%%%%%%%%%%%%%%%%%%%%%%%%%%%%%%%%%%%%%%%%%%%%%%%%%%%%%%

\subsection{Hybrid State Space}

Several recent studies adopt \textit{hybrid} state representations that fuse high-level perception and low-level vehicle data to tackle the complexity of autonomous driving tasks. For instance, Wang et al. \cite{wang_vision-based_2023} propose a hierarchical approach that splits decision-making between a high-level agent, which selects maneuvers (e.g., lane following, turns), and a low-level agent controlling steering, speed, and throttle. Their observation space combines semantic segmentation (ENet) for drivable areas, object detection (YOLOv4) for vehicles and pedestrians, and vehicle kinematics, leading to more informed decisions in urban settings. Khalil et al.~\cite{khalil_exploiting_2023} also employ a fused representation, relying on LiCaNext outputs to project LiDAR-based object detection and motion fields onto a BEV. By including temporal history and previous control inputs, they maintain critical environmental context while reducing the dimensionality of raw sensor data. Bai et al.~\cite{bai_fen-dqn_2023} focus on a Feature Extraction Network (FEN) that processes camera images into semantic affordances—such as center distance and hazard indicators—while also including the agent’s speed, traffic light state, and leading vehicle parameters, helping a DQN handle the complexity of dynamic traffic. Muhammed et al.~\cite{muhammed_developing_2021} merge depth and semantic segmentation images to provide pixel-wise distance-to-objects and class labels, processed by a CNN for enhanced spatial understanding and obstacle avoidance. Aghdasian et al. \cite{aghdasian_autonomous_2023} similarly combine high-level visual features from a residual CNN with low-level kinematic data (e.g., velocity, orientation, lane-center distance), creating a fused latent space that balances scene perception and precise motion control. In a different vein, the work P´erez-Gil et al. \cite{Deep_reinforcement_learning_based_control_for_Autonomous_Vehicles_in_CARLA} integrates either raw RGB images or waypoints with a driving feature vector, which includes vehicle speed and lane center information, highlighting the importance of combining visual cues with essential geometric features.

Additional hybrid strategies leverage latent representations or domain adaptation to ensure more robust policy learning. For example, Chen et al.~\cite{Learning_to_drive_from_a_world_on_rails} fuses multi-camera (three) inputs to produce  high-level navigation commands, Anzalone et al. \cite{Reinforced_Curriculum_Learning_For_Autonomous_Driving} stitches multi-camera RGB images over time and combines them with road, vehicle, and navigational features for more efficient training in dynamic environments. Transformer-based multimodal fusion is seen in \cite{Decision_Making_for_Autonomous_Driving_Via_Multimodal_Transforme}, where Fu et al. combined camera images, LiDAR bird’s-eye pseudo-images, and ego-vehicle dynamics to enable attention-based decision-making. Similarly, May et al.~\cite{Using_the_CARLA_Simulator_to_Train_A_Deep_Q_Self-Driving} merges data from three RGB cameras, radar, GPS, IMU, and collision sensors into a single representation. The integration of model-based path planners with deep RL is explored by Yurtsever et al.~\cite{Integrating_Deep_Reinforcement_Learning_with_Model-based_Path_Planners_for_Automated_Driving}, combining CNN-encoded camera data with vehicle states and waypoint distances. Finally, Jo et al.~\cite{An_Integrated_Reward_Function} processes camera and LiDAR bird’s-eye views into a low-dimensional latent space alongside vehicle kinematics, underscoring how compressed multimodal features can guide effective policy learning. Collectively, these works illustrate that hybrid state representations—spanning raw imagery, semantic cues, object detection outputs, vehicle kinematics, and driver or route-level attributes—consistently improve situational awareness and decision-making performance in RL-based autonomous driving systems.

\subsection{CARLA Kinematics only State Space}

Below is a consolidated description of a typical \emph{CARLA kinematic-based observation} for autonomous driving tasks in RL, integrating approaches from multiple recent works % ~\cite{jia_safe_2023,xu_decision-making_2022,wei_continual_2023,wu_lane_2022,chen_motion_2024,goel_adaptive_2021,frauenknecht_data-efficient_2023,yang_decision-making_2022,liao_lateral_2023,Deep_Reinforcement_Learning_based_control_algorithms:_Training_and_validation_using_the_ROS_Framework,Cognitive_Reinforcement_Learning,Autonomous_Driving_via_Knowledge-Enhanced,A_Decision_Control_Method,Enhanced_Off-Policy_Reinforcement_Learning_With_Focused_Experience_Replay,Reinforcement_Learning-Based_Guidance_of_Autonomous_Vehicles,Hierarchical_Reinforcement_Learning_with_Successor_Representation_for_Intelligent_Vehicle_Collision_Avoidance_of_Dynamic_Pedestrian,Deep-Reinforcement-Learning-Based_Driving_Policy_at_Intersections,wu_reinforcement_2023,chronis_driving_2021}.
Although individual studies vary in the precise composition and dimensionality
of the observation vector, the following elements commonly appear.

\begin{enumerate}
%\subsubsection{Control Dimension}
\item Ego-Vehicle Kinematic State:

% \subsubsection{Ego-Vehicle Kinematic State}
\begin{itemize}
\item Position and Orientation:
A core component is the ego vehicle’s global or local-frame position (often denoted \(x,y\))
and heading angle (\(\text{yaw}\)). Many works encode yaw as \(\cos(\text{yaw})\) and
\(\sin(\text{yaw})\) to avoid discontinuities~\cite{Deep-Reinforcement-Learning-Based_Driving_Policy_at_Intersections,yang_decision-making_2022}. Some papers also track roll, pitch, or higher-order motion states (e.g., yaw rate, pitch rate) to capture full 3D pose dynamics~\cite{frauenknecht_data-efficient_2023}. Others include an explicit \emph{heading error} relative to a desired path or lane center instead of absolute yaw~\cite{goel_adaptive_2021,Cognitive_Reinforcement_Learning}.

\item Velocity and Acceleration:
The longitudinal speed \(v\) (m/s) is almost always present. Many studies also include lateral velocity \(v_y\), longitudinal acceleration \(a_x\), or even higher derivatives like \(\dot{x},\,\dot{y},\,\ddot{x},\,\ddot{y}\) in local coordinates to capture the ego vehicle’s instantaneous motion~\cite{frauenknecht_data-efficient_2023,chen_motion_2024,jia_safe_2023,wei_continual_2023}. Yaw rate \(\dot{\gamma}\) can similarly be added~\cite{wu_lane_2022,wu_reinforcement_2023}.

\item Lateral/Longitudinal Deviations:
Many lane-keeping or lane-changing tasks explicitly track the ego vehicle’s lateral offset from a reference lane boundary~\cite{liao_lateral_2023,Autonomous_Driving_via_Knowledge-Enhanced}, cross-track error~\cite{goel_adaptive_2021}, or distance to a goal position~\cite{wei_continual_2023,wu_reinforcement_2023}.

\end{itemize}

\item{Surrounding Traffic and Obstacles}

\begin{itemize}
\item Relative Positions and Velocities:
To handle multi-vehicle traffic, numerous works include distances or relative coordinates of up to two or more neighboring vehicles~\cite{jia_safe_2023,wu_lane_2022,yang_decision-making_2022}.
Relative velocity components \((\Delta v_x,\Delta v_y)\) are frequently used~\cite{jia_safe_2023}, and some studies extend this to angle or heading differences~\cite{liao_lateral_2023}.

\item Distance to Obstacles or Road Boundaries:
In maneuvers such as lane changing or parking, the agent may need direct measures of distance to lane edges and obstacles on each side~\cite{liao_lateral_2023,wu_reinforcement_2023}.
In unsignalized intersection or pedestrian-avoidance tasks, explicit collision-risk terms or booleans (e.g., ``obstacle ahead'') are added~\cite{Cognitive_Reinforcement_Learning,Hierarchical_Reinforcement_Learning_with_Successor_Representation_for_Intelligent_Vehicle_Collision_Avoidance_of_Dynamic_Pedestrian}.

\item Use of Graph or Waypoint Structures:
Some recent research vectorizes map information (lane segments, HD map nodes) and adjacency relationships (predecessor/successor/left/right) to better capture road topology~\cite{Deep-Reinforcement-Learning-Based_Driving_Policy_at_Intersections,Autonomous_Driving_via_Knowledge-Enhanced}.
Others represent upcoming waypoints directly as coordinates in the ego frame~\cite{Deep_Reinforcement_Learning_based_control_algorithms:_Training_and_validation_using_the_ROS_Framework}.
\end{itemize}

\item{Additional Task-Specific Features}

\begin{itemize}

\item Lane Indicators and Discrete Events:
Discrete lane IDs, turn signal statuses, or lane-change flags are occasionally encoded to help the network differentiate maneuver intentions~\cite{yang_decision-making_2022,jia_safe_2023}.

\item Driver-Centric Metrics:
In driver-assistance contexts, some works include ``driver stress'' or ``excitement'' measures as direct numeric inputs, reflecting a more human-centric policy~\cite{chronis_driving_2021}.

\item Road Speed Limits and Traffic Signals:
For tasks involving traffic rules, the current speed limit or traffic signal states (green/red) may appear in the observation vector~\cite{chronis_driving_2021,Enhanced_Off-Policy_Reinforcement_Learning_With_Focused_Experience_Replay}.

\item Minimal vs.\ Extended Observation Spaces:
Simpler methods might only use \((x,y,\text{yaw},v)\)~\cite{Cognitive_Reinforcement_Learning,Reinforcement_Learning-Based_Guidance_of_Autonomous_Vehicles}, while more advanced approaches incorporate up to 20--30 features or even on-board sensor readings (lidar/radar arrays) processed into bounding distances~\cite{wu_reinforcement_2023,liao_lateral_2023}.
\end{itemize}

\item{Typical Dimensionality and Implementation Details}

\begin{itemize}
\item Observation Dimension:
Ranges from as few as 4--6 scalars (for minimal steering or speed control tasks)~\cite{goel_adaptive_2021,Cognitive_Reinforcement_Learning} up to 20+ dimensional vectors when including multiple vehicles, and lane offsets. Some wrappers for CARLA yield 6--10 continuous values by default~\cite{Enhanced_Off-Policy_Reinforcement_Learning_With_Focused_Experience_Replay}.

\item Local vs.\ Global Coordinates:
Many prefer an \emph{ego-centric} frame, ensuring the ego vehicle is always at the origin with heading zero~\cite{chen_motion_2024,Deep-Reinforcement-Learning-Based_Driving_Policy_at_Intersections}. This can simplify learning because the agent’s network does not need to adapt to arbitrary global positions.

\item Stacking Observations:
Combining consecutive frames (e.g., last four states) is widely adopted to capture short-term dynamics and reduce partial observability~\cite{xu_decision-making_2022,A_Decision_Control_Method,chen_motion_2024}.

\end{itemize}
\end{enumerate}
In sum, CARLA kinematic-based observations typically revolve around the ego vehicle’s position, orientation, and velocities, often augmented by lane-relative or goal-relative errors, and surrounding-vehicle states. These standardized kinematic features provide a strong foundation for policy learning in lane keeping, lane changing, intersection negotiation, and various maneuvering scenarios in urban or highway settings~\cite{jia_safe_2023,xu_decision-making_2022,wei_continual_2023,wu_lane_2022,chen_motion_2024,goel_adaptive_2021,yang_decision-making_2022,liao_lateral_2023,Deep_Reinforcement_Learning_based_control_algorithms:_Training_and_validation_using_the_ROS_Framework,Cognitive_Reinforcement_Learning,Autonomous_Driving_via_Knowledge-Enhanced,A_Decision_Control_Method,Enhanced_Off-Policy_Reinforcement_Learning_With_Focused_Experience_Replay,Reinforcement_Learning-Based_Guidance_of_Autonomous_Vehicles,Hierarchical_Reinforcement_Learning_with_Successor_Representation_for_Intelligent_Vehicle_Collision_Avoidance_of_Dynamic_Pedestrian,Deep-Reinforcement-Learning-Based_Driving_Policy_at_Intersections,wu_reinforcement_2023,chronis_driving_2021}.
Depending on the task, this base observation can be expanded with additional environment, driver, or sensor data, creating a flexible framework for RL in autonomous driving.

\section{Action Space}

The choice of action space is a fundamental design decision in RL for autonomous driving, as it directly shapes the agent’s ability to explore, learn, and execute precise maneuvers. In CARLA-based driving tasks, action spaces have been defined in discrete, and continuous forms to balance trade-offs between sample efficiency, control fidelity, and interpretability. We considered a new category for hierarchical studies as their action space included two different levels. In this section, we first review discrete formulations—ranging from coarse, high-level primitives to fine-grained binning strategies—before examining continuous designs that offer smooth control via normalized or custom scaled outputs. We then discuss hierarchical studies and how they considered their action space.

%%%%%%%%%%%%%%%%%
\subsection{Discrete Action Space}
%%%%%%%%%%%%%%%%%
Action spaces in RL for autonomous driving using CARLA can be broadly categorized into discrete, continuous, or hybrid. In autonomous driving applications using RL, the design of the discrete action space plays a critical role in balancing learning complexity and control precision. Across the literature, the discrete action space can be categorized along several dimensions. These categories, along with representative examples, are summarized in Table~\ref{tab:action_space_categorization}.

\begin{enumerate}
%\subsubsection{Control Dimension}
\item Control Dimension:
\begin{itemize}
    \item {Lateral Control}: Decisions are focused on lateral maneuvers (e.g., lane changes or steering adjustments).
    \item {Longitudinal Control}: The focus is on speed-related decisions (e.g., throttle and braking).
    \item {Combined Control}: Both lateral and longitudinal controls are considered together.
\end{itemize}

\item{Granularity}:
\begin{itemize}
    \item Coarse: Only a few discrete actions (often fewer than 10) are defined. These actions are mostly high-level actions like accelerate, decelarate, turn left, turn right, that are mapped to small combination of the steering and acceleration.
    \item Moderate: An intermediate number of actions (typically in the range of 10--30).
    \item Fine: A detailed discretization that divides the control range into many bins (e.g., 27, 108, 190 actions).
\end{itemize}

\item{Mapping Strategy{*}}:
\begin{itemize}
    \item Direct Mapping: Discrete actions are mapped directly to specific low-level control commands (e.g., steering and throttle values), which are applied to the environment without additional processing.
    \item Indirect Mapping: Discrete actions are translated into control signals by an external controller or planning module (e.g., a PID).
\end{itemize}
\end{enumerate}

\begingroup
\renewcommand\thefootnote{*}%
\footnotetext{In this study we categorized the mapping strategy of the papers that has not explicitly mentioned the low level controller as direct.}
\endgroup
%\subsubsection{Discussion}

This categorization illustrates the diversity of design choices in discrete action space formulations for autonomous driving. A majority of work uses flat action space with direct mapping with different granularity. Some like \cite{Safe_Navigation_Based_on_Deep_Q-Network_Algorithm_Using,A_Reinforcement_Learning_Based_Approach_for_Controlling_Autonomous_Vehicles_in_Complex_Scenarios,hussonnois_end--end_2022,shi_efficient_2023} use coarse action space to simplify the decision-making, while others \cite{End-to-End_Model-Free_Reinforcement_Learning_for_Urban_Driving,elallid_dqn-based_2022,GRI:General_Reinforced_Imitation} use fine-grained discretizations (up to 190 actions) to achieve better control precision. A small portion of studies incorporate indirect mapping, where the action chosen by the RL agent is given to a PID controller for execution \cite{221016567DeFIXDetecting,yang_decision-making_2022}.  Finally, hierarchical approaches—such as \cite{A_Hierarchical_Autonomous_Driving_Framework,naveedTrajectoryPlanningAutonomous2020,gangopadhyayHierarchicalProgramTriggeredReinforcement2022}  employ a multi-layered decision process, where a high-level policy selects abstract maneuvers (e.g., ”turn left”) that are refined into low-level controls. These strategies help address the complexity of continuous control while offering modularity and interpretability.

\begin{table}[htbp]
\centering
%\tiny
\caption{Taxonomy of Discrete Action Space Designs in RL-Based Autonomous Driving}
\label{tab:action_space_categorization}

%\begin{adjustbox}{max width=\textwidth}
%\begin{adjustbox}{max width=\columnwidth}

\begin{tabular}{p{2.2cm} p{2cm} p{1.2cm} p{2cm}}
\toprule
\textbf{Control Dim.} & \textbf{Granularity} & \textbf{Mapping Strategy} & \textbf{References} \\
\midrule

Lateral only & Coarse-Moderate & Direct &
\cite{wang_benchmarking_2021,A_Methodology_Based_on_Deep_Reinforcement_Learning} \\
\midrule

Longitudinal only & Coarse-Fine & Direct &
 \cite{muhammed_developing_2021,cheng_longitudinal_2021,bai_fen-dqn_2023,Learning_Automated_Driving_in_Complex_Intersection_Scenarios_Based,gutierrez-morenoDecisionMakingAutonomous2024,deshpande_navigation_2021}
\\
\midrule

Combined  & Coarse & Direct &
 \cite{Safe_Navigation_Based_on_Deep_Q-Network_Algorithm_Using,A_Reinforcement_Learning_Based_Approach_for_Controlling_Autonomous_Vehicles_in_Complex_Scenarios,Using_the_CARLA_Simulator_to_Train_A_Deep_Q_Self-Driving,Reinforcement_Learning-Based_Guidance_of_Autonomous_Vehicles,Deep_Reinforcement_Learning_Control_Strategy_at_Roundabout_for_i-CAR,A_Bayesian_Approach_to_Reinforcement_Learning_of_Vision-Based_Vehicular_Control,Hierarchical_Reinforcement_Learning_with_Successor_Representation_for_Intelligent_Vehicle_Collision_Avoidance_of_Dynamic_Pedestrian,frijiDQNBasedAutonomousCarFollowing2020,hishmehDeerHeadlightsShort2020,gupta_hylear_2023,hussonnois_end--end_2022,shi_efficient_2023,noauthor_explainable_nodate,manikandan_ad_2023}

\\
\midrule

Combined  & Fine-Moderate & Direct &  \cite{GRI:General_Reinforced_Imitation,elallid_dqn-based_2022,Think2Drive,End-to-End_Model-Free_Reinforcement_Learning_for_Urban_Driving,Deep_reinforcement_learning_based_control_for_Autonomous_Vehicles_in_CARLA,Learning_to_drive_from_a_world_on_rails,A_Deep_Q-Network_Reinforcement_Learning-Based,Integrating_Deep_Reinforcement_Learning_with_Model-based_Path_Planners_for_Automated_Driving,zhang_toward_2022}
\\
\midrule

%Combined & Flat & Moderate & Direct &
%\cite{Deep_reinforcement_learning_based_control_for_Autonomous_Vehicles_in_CARLA}, \cite{Learning_to_drive_from_a_world_on_rails}. \\
%\midrule

Combined  & Coarse-Moderate & Indirect (PID) & \cite{chronis_driving_2021,yang_decision-making_2022,221016567DeFIXDetecting}
\\
%\midrule

%Combined & Flat  & Coarse & Indirect (PID) & \cite{chronis_driving_2021}. \\
%\midrule
%Combined & Flat (steering angles categorized into broad groups) & Coarse-Moderate & Direct &
% \cite{manikandan_ad_2023}. \\
%\midrule

%Combined & Flat (discretized values but exact numbers not mentioned) & Moderate & Direct &
%\cite{Integrating_Deep_Reinforcement_Learning_with_Model-based_Path_Planners_for_Automated_Driving},\cite{zhang_toward_2022} \\
%\midrule

%Combined (steering + continuous others) & Flat (steering discretized, others continuous) & Fine & Direct &  \cite{A_Deep_Q-Network_Reinforcement_Learning-Based} \\
%\midrule

%%%%%Combined & Hierarchical  & Coarse & Direct & \cite{A_Hierarchical_Autonomous_Driving_Framework,naveedTrajectoryPlanningAutonomous2020,gangopadhyayHierarchicalProgramTriggeredReinforcement2022,albilani_guided_2023}

%%%%%\\
%%%%\midrule

%%%%Combined & Hierarchical & Moderate & Indirect (PID) &  \cite{chen_motion_2024} \\

\bottomrule
\end{tabular}
%\end{adjustbox}
\end{table}

%%%%%%%%%%%%%%%%%%%%%%%%%%%%%
\subsection{Continuous Action Space}
%%%%%%%%%%%%%%%%%%%%%%%%%%%%%

In RL for autonomous driving, continuous action spaces are widely used to provide fine-grained control over vehicle motion. These spaces enable smooth steering and acceleration adjustments, which are essential for realistic and safe driving behavior, but they also introduce challenges in terms of training stability and interpretability. Table~\ref{tab:control_action_mapping} summarizes representative studies and categorizes continuous action space designs across several key dimensions.
\begin{enumerate}
%\subsubsection{Control Dimensionality}
\item Control Dimensionality:
\begin{itemize}
    \item Lateral Only: The agent controls only the steering angle, with speed managed externally or held constant.
    \item Longitudinal Only: The policy controls speed via throttle, or braking.
    \item Combined Control: The most common approach, where the agent jointly predicts steering and speed commands. This may include:
    \begin{itemize}
        \item Combined (negative braking): The Steering range is  $[-1,1]$ and Throttle and brake values in $[-1,1]$, with negative values ($[-1,0]$) representing braking.
        \item Combined (no brake): The Steering range is  $[-1,1]$ and  Throttle is bounded in $[0,1]$, with braking implicitly handled by reducing throttle.
        \item Combined (separate brake): The Steering range is  $[-1,1]$ and  Throttle is bounded in $[0,1]$, Brake is represented as a separate continuous output in range $[0,1]$.
    \end{itemize}
\end{itemize}

\item Action Range and Normalization:
\begin{itemize}
    \item Normalized Range: Many studies use normalized outputs (e.g., $[-1,1]$) and apply activation functions like \texttt{tanh} to bound the values.
    \item Dedicated or Tuned Ranges: A few studies customize control ranges, for example scaling steering from $[-0.5, 0.5]$ to $[-40^\circ, 40^\circ]$ \cite{Learning_Urban_Driving_Policies_using_Deep_Reinforcement_Learning}.
\end{itemize}

\item Mapping Strategy:
\begin{itemize}
    \item Direct Mapping: The majority of methods directly apply the policy’s output to the simulator’s low-level control interface.
    \item Indirect Mapping: A subset of works uses the agent’s output to specify control targets (e.g., speed or acceleration), which are then achieved using external control modules such as PID controllers.
\end{itemize}
\end{enumerate}

As shown in Table~\ref{tab:control_action_mapping}, continuous action space design is diverse, yet several patterns emerge. Most approaches adopt a combined control structure where throttle and steering are handled simultaneously—either as normalized values in $[-1, 1]$ \cite{deng_context-aware_2024,wu_reinforcement_2023}, bounded distributions like Beta \cite{End-to-End_Urban_Driving_by_Imitating_a_Reinforcement_Learning_Coach}, or custom physical units \cite{Cognitive_Reinforcement_Learning}. Several studies simplify the control interface by merging throttle and brake into a single signal, using negative values to represent braking \cite{End-to-End_Urban_Driving_by_Imitating_a_Reinforcement_Learning_Coach,Reinforced_Curriculum_Learning_For_Autonomous_Driving}, while others maintain separate channels for acceleration and braking to increase control expressiveness and realism \cite{Deep_Reinforcement_Learning_for_Autonomous_Vehicle_Intersection_Navigation, tsaiAutonomousVehicleFollowingTechnique2023}. In longitudinal-only setups, throttle or acceleration is learned by the agent, while steering is handled via traditional controllers like PID or Stanley, enabling modular architectures \cite{Efficient_Learning_of_Urban_Driving, Deep-Reinforcement-Learning-Based_Driving_Policy_at_Intersections}. Interestingly, several works employ hybrid pipelines where the RL policy outputs high-level targets (e.g., desired speed or acceleration), which are then executed by a PID controller \cite{Deep-Reinforcement-Learning-Based_Driving_Policy_at_Intersections, igoeMultiAlphaSoftActorCritic2023}. A few recent studies introduce predictive control over future steps \cite{zhao_real-time_2021} or encode control commands using task-specific scalings or semantic units \cite{xu_end--end_2023}, showing a trend toward physically interpretable action spaces. Finally, lateral-only designs \cite{liao_lateral_2023, goel_adaptive_2021} demonstrate that decoupling steering from speed control remains useful for focused lane-keeping tasks or simplified evaluations. Overall, the diversity in continuous action space design reflects ongoing efforts to balance precision, safety, and learnability in real-world driving scenarios.

\begin{table}[htbp]
  \centering
  %\small
  %\tiny
  \caption{Taxonomy of Continuous Action Space Designs in RL-Based Autonomous
Driving}
  \label{tab:control_action_mapping}
 % \begin{adjustbox}{max width=\columnwidth}
  \begin{tabular}{p{2.5cm} p{2cm} p{2cm} p{3cm}}
    \toprule
    \textbf{Control Type} & \textbf{Action Range}  & \textbf{Mapping} & \textbf{Reference} \\

    \midrule

    %%%%%%%%%%%%%%% longitudinal only %%%%%%%%%%
    % 2
    Longitudinal only & Dedicated range & Direct & \cite{xu_decision-making_2022,wei_continual_2023,li_decision-making_2024,Efficient_Learning_of_Urban_Driving}
    \\
    \midrule

    % 2
    Longitudinal only & Dedicated range & Indirect & \cite{Deep-Reinforcement-Learning-Based_Driving_Policy_at_Intersections,martinez_gomez_temporal_2023,udatha_reinforcement_2023}
    \\
    \midrule

    %%%%%%%%%%%%%% Lateral only %%%%%%%%%%%%%%%%
     % 10
    Lateral only & Dedicated range  & Direct &  \cite{liao_lateral_2023}\\
    \midrule

    % 10
    Lateral only & Dedicated range  & Indirect & \cite{goel_adaptive_2021} \\
    \midrule

        % 7
    %Combined (or Lateral only in 1-action) & The exact range not explicitly mentioned 1-action: Steering; 2-action: Steering, Acceleration & & \textcolor{red}{we should decide if we want to add this into other categories}Experiments include both steering-only and combined control settings \cite{mohammed_unified_2024}. \\
    %\midrule
    
    % 1
    Combined & Not Mentioned & Not Mentioned  &  \cite{jia_safe_2023,carton_using_2021,khalil_exploiting_2023,wu_lane_2022,zhu_rita_2023,zhao_end--end_2024,zhao_real-time_2021,peng_imitative_2020,Domain_Adaptation_In_Reinforcement_Learning,Decision_Making_for_Autonomous_Driving_Via_Multimodal_Transforme,Enhanced_Off-Policy_Reinforcement_Learning_With_Focused_Experience_Replay,An_Integrated_Reward_Function,Towards_Autonomous_Driving,heAdaptiveDecisionMaking2022,jaafraContextAwareAutonomousDriving2019,mohammed_unified_2024,jin_vwpefficient_2024,huang_simoun_2023}

    \\
    \midrule

    %%%%%%%%%%%% combined negative for braking %%%%%%%%%%%%%%
    % 9
    Combined (negative braking) & Steering: $[-1,1]$, Throttle and Brake: $[-1,1]$ & Direct &\cite{deng_context_2023,deng_context-aware_2024,wu_reinforcement_2023,coutoHierarchicalGenerativeAdversarial2024,zhangSelflearningLaneKeeping2021,li_dynamic_2022,baheri_vision-based_2020,frauenknecht_data-efficient_2023,youssef_deep_2019,Versatile_and_Efficien,End-to-End_Urban_Driving_by_Imitating_a_Reinforcement_Learning_Coach,Reinforced_Curriculum_Learning_For_Autonomous_Driving}
    
    \\
    \midrule

        % 9
    Combined (negative braking) & Steering: $[-1,1]$, Throttle and Brake: $[-1,1]$ & Indirect & \cite{igoeMultiAlphaSoftActorCritic2023,Urban_Autonomous_Driving_of_Emergency_Vehicles_with_Reinforcement_Learning}
    
    \\
    \midrule

    %%%%%%%%%%%%%% combined but no braking %%%%%%%%%%%%%%%%
     % 11
    Combined (no brake) & Steering: $[-1,1]$, Throttle: $[0,1]$ & Direct & \cite{savari_online_2021,wu_proximal_2023,aghdasian_autonomous_2023,Deep_reinforcement_learning_based_control_for_Autonomous_Vehicles_in_CARLA,Vision-Based_Autonomous_Car_Racing_Using_Deep_Imitative_Reinforcement_Learning,Addressing_Lane_Keeping_and_Intersections} \\
    \midrule

    %%%%%%%%%%%%% brake seperately %%%%%%%%%%%%%%%%%%%%%%%%
    % 34
    Combined (separate brake) & Steering: $[-1,1]$, Acceleration: $[0,1]$, Braking: $[0,1]$ & Direct & \cite{Deep_Reinforcement_Learning_for_Autonomous_Vehicle_Intersection_Navigation,Deductive_Reinforcement_Learning,tsaiAutonomousVehicleFollowingTechnique2023,udatha_reinforcement_2023,hanLearningDriveUsing2022}
    
    \\
    \midrule

    %%%%%%%%%%%%% combined specific range %%%%%%%%%%%%%%%%%
     % 12
    Combined (specific range) & Dedicated range & Direct & \cite{xu_end--end_2023,Model-Based_Reinforcement_Learning_with_Isolated_Imaginations,Cognitive_Reinforcement_Learning} \\
    \midrule

    % 12
    Combined (specific range) & Dedicated range & Indirect & \cite{Learning_Urban_Driving_Policies_using_Deep_Reinforcement_Learning}  \\
    %\midrule

    \bottomrule
  \end{tabular}
 % \end{adjustbox}
\end{table}

%%%%%%%%%%%%%%%%%%%%%%%%
\subsection{Hierarchical Studies}
%%%%%%%%%%%%%%%%%%%%%%%%

\begin{table*}[!t]
\centering
\small
\caption{Two‐tier hierarchical action space designs}
\setlength\tabcolsep{6pt} % tweak cell padding if you like
%\begin{tabularx}{\textwidth}{@{}p{2.5cm} X X@{}}
\begin{tabularx}{\textwidth}{@{}X X p{2.5cm}@{}}
\toprule
  \textbf{High‐Level Actions} & \textbf{Low‐Level Actions / Controller} & \textbf{Reference} \\
\midrule

   Follow lane; Go straight; Turn left; Turn right
  & Imitation‐learning policies mapping observations to continuous steering [-1,1], throttle [0,1] and brake [0,1] 
  & \cite{A_Hierarchical_Autonomous_Driving_Framework} \\[4pt]

   Lane follow / wait; Lane change
  & PID‐tracked waypoints selected as sub‐goals 
  & \cite{naveedTrajectoryPlanningAutonomous2020} \\[4pt]

   Maneuver‐specific agents: Straight‐drive; Turn‐left/right; Lane‐change
  & Dedicated discrete action set of steering and throttle for each manuever 
  & \cite{gangopadhyayHierarchicalProgramTriggeredReinforcement2022} \\[4pt]

   Options via Option‐Critic: Lane following; Intersection handling; Obstacle avoidance
  & Atomic maneuvers (accelerate; brake; turn left; turn right) via PPO; Answer Set Programming (ASP) rules override 
  & \cite{albilani_guided_2023} \\[4pt]

   Five lateral offsets (Frenet frame waypoints)  followed by PID controller.
  &Six discrete brake/throttle settings 
  & \cite{chen_motion_2024} \\[4pt]

   Road‐following; Left turn; Right turn
  & Continuous throttle/brake/steer sampled from bounded Beta distributions in range [0,1]. These raw commands are then passed through PID controllers. 
  & \cite{wang_vision-based_2023} \\[4pt]

   Maintain lane; Change left; Change right
  & DDPG actor‐critic outputs continuous steering [-1,1] or accel/brake [-1,1] 
  & \cite{A_Decision_Control_Method} \\
\bottomrule
\end{tabularx}

\label{tab:hierarchical}
\end{table*}

Across these seven works mentioned in Table \ref{tab:hierarchical}, a two-tier hierarchy is the common thread: a small, discrete set of high-level maneuvers drives exploration and safety, while low-level continuous controllers (imitation-learning policies, PID loops, or actor-critic networks) deliver smooth vehicle commands. The high-level action vocabularies vary—from simple lane-following vs. lane-change choices to richer option sets in Option-Critic architectures—but always aim to reduce policy complexity. At the low level, controllers are typically PID-based or learned via DDPG/PPO, striking a balance between precision and learnability. Together, these designs show that decoupling decision planning from continuous actuation not only improves sample efficiency but also enhances interpretability and provable safety in autonomous-driving RL systems.

\section{Reward}
Reward design is a fundamental component of RL in autonomous driving, as it directly shapes the agent's behavior. A wide variety of reward terms have been proposed in the literature, reflecting the diverse priorities of different driving tasks—such as safety, goal completion, comfort, and rule compliance. Fig. \ref{fig:reward-distribution} shows the distribution of reward terms across papers. Notably, collision penalties, speed control, and deviation from the lane center are among the most frequently used rewards, reflecting the emphasis on safety and lane-keeping behavior. Other commonly used rewards include goal-based rewards, heading alignment, and steering smoothness (e.g., jerk and steering cost), which guide the agent toward efficient and comfortable driving.

\begin{figure}[htbp]
    \centering
    \includegraphics[width=0.50\textwidth]{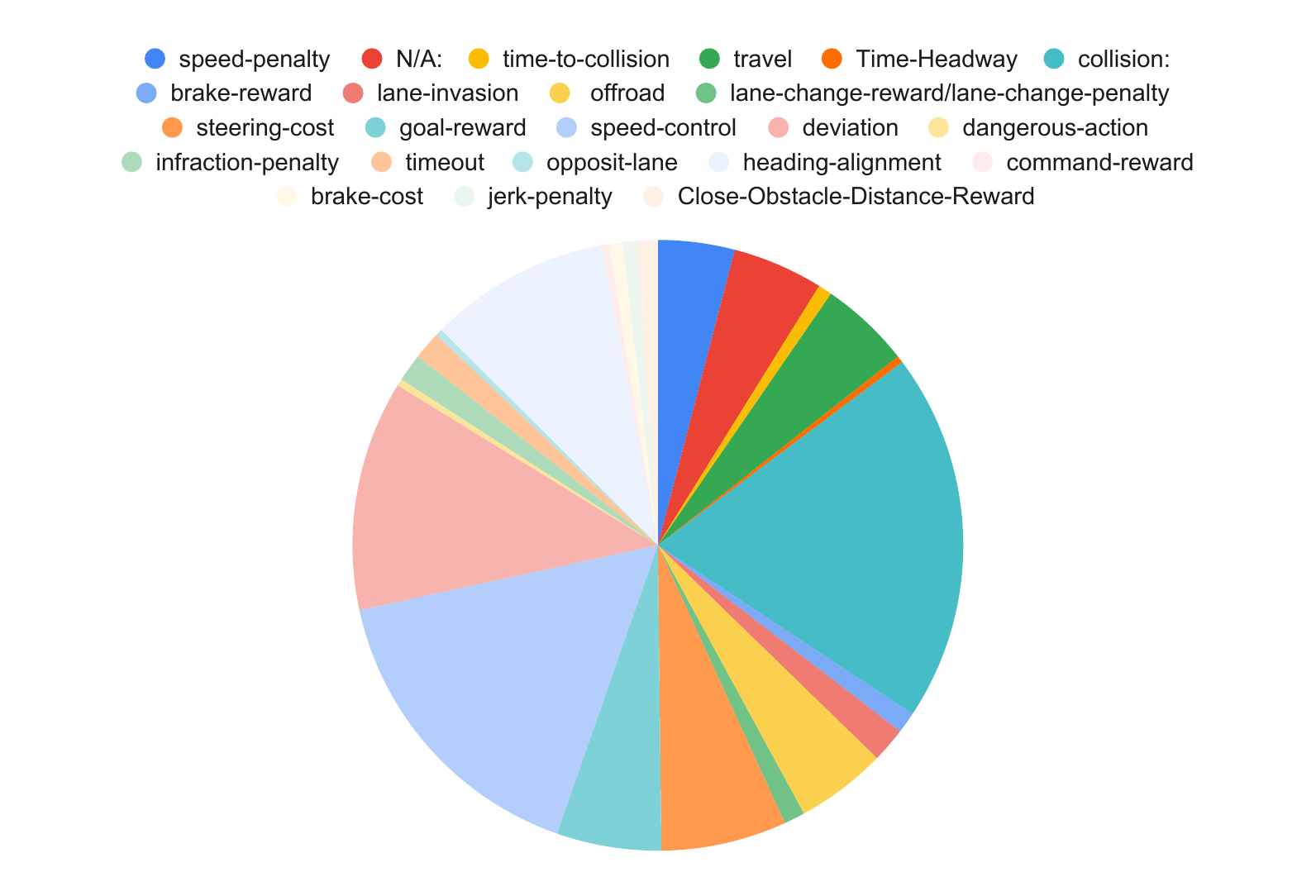} % Adjust path and size
    \caption{Distribution of reward types used in RL for autonomous driving tasks.}
    \label{fig:reward-distribution}
\end{figure}

Table \ref{tab:reward_categorization} categorizes the different reward types along with their purpose and density, and cites key works that have employed them. This categorization aims to help future researchers choose or design reward functions suited to their task.
\begin{table*}[htbp]
\centering
%\tiny
\small
\caption{Categorization of Reward Functions Used in RL for Autonomous Driving}
\label{tab:reward_categorization}

%\begin{adjustbox}{max width=\columnwidth}
\begin{adjustbox}{max width=\textwidth}
\begin{tabular}{p{3cm} p{2.2cm} p{2cm} p{6cm}} % Adjust column width as needed
\toprule
\textbf{Reward Term} & \textbf{Purpose} & \textbf{Dense/Sparse} &\textbf{References} \\
\midrule

\textbf{Collision} & Safety & Sparse&\cite{Versatile_and_Efficien,Deep_reinforcement_learning_based_control_for_Autonomous_Vehicles_in_CARLA,Vision-Based_Autonomous_Car_Racing_Using_Deep_Imitative_Reinforcement_Learning,End-to-End_Urban_Driving_by_Imitating_a_Reinforcement_Learning_Coach,Learning_Urban_Driving_Policies_using_Deep_Reinforcement_Learning,Efficient_Learning_of_Urban_Driving,Model-Based_Reinforcement_Learning_with_Isolated_Imaginations,Reinforced_Curriculum_Learning_For_Autonomous_Driving,Autonomous_Driving_via_Knowledge-Enhanced,A_Decision_Control_Method,Decision_Making_for_Autonomous_Driving_Via_Multimodal_Transforme,Deep_Reinforcement_Learning_for_Autonomous_Vehicle_Intersection_Navigation,Deductive_Reinforcement_Learning,Safe_Navigation_Based_on_Deep_Q-Network_Algorithm_Using,A_Bayesian_Approach_to_Reinforcement_Learning_of_Vision-Based_Vehicular_Control,A_Hierarchical_Autonomous_Driving_Framework,A_Methodology_Based_on_Deep_Reinforcement_Learning,Deep_Reinforcement_Learning_Control_Strategy_at_Roundabout_for_i-CAR,Integration_of_Motion_Prediction_with_End-to-end_Latent_RL_for_Self-Driving_Vehicles,A_Reinforcement_Learning_Based_Approach_for_Controlling_Autonomous_Vehicles_in_Complex_Scenarios,Using_the_CARLA_Simulator_to_Train_A_Deep_Q_Self-Driving,Integrating_Deep_Reinforcement_Learning_with_Model-based_Path_Planners_for_Automated_Driving,Reinforcement_Learning-Based_Guidance_of_Autonomous_Vehicles,Learning_Automated_Driving_in_Complex_Intersection_Scenarios_Based,Hierarchical_Reinforcement_Learning_with_Successor_Representation_for_Intelligent_Vehicle_Collision_Avoidance_of_Dynamic_Pedestrian,Towards_Autonomous_Driving,A_Deep_Q-Network_Reinforcement_Learning-Based,Urban_Autonomous_Driving_of_Emergency_Vehicles_with_Reinforcement_Learning,Deep-Reinforcement-Learning-Based_Driving_Policy_at_Intersections,wang_vision-based_2023,xu_decision-making_2022,wei_continual_2023,wang_benchmarking_2021,carton_using_2021,khalil_exploiting_2023, wu_lane_2022, mohammed_unified_2024, chen_motion_2024, zhang_toward_2022, goel_adaptive_2021, xu_end--end_2023, deng_context-aware_2024, wu_reinforcement_2023, zhao_end--end_2024, frauenknecht_data-efficient_2023, yang_decision-making_2022, bai_fen-dqn_2023, wu_proximal_2023, muhammed_developing_2021, youssef_deep_2019, liao_lateral_2023, peng_imitative_2020, elallid_dqn-based_2022,gupta_hylear_2023,hussonnois_end--end_2022,deshpande_navigation_2021,albilani_guided_2023,martinez_gomez_temporal_2023,jin_vwpefficient_2024,shi_efficient_2023,ahmed_policy-based_2022,udatha_reinforcement_2023,noauthor_explainable_nodate,gutierrez-morenoDecisionMakingAutonomous2024,heAdaptiveDecisionMaking2022,tsaiAutonomousVehicleFollowingTechnique2023,naveedTrajectoryPlanningAutonomous2020,gangopadhyayHierarchicalProgramTriggeredReinforcement2022,frijiDQNBasedAutonomousCarFollowing2020,jaafraContextAwareAutonomousDriving2019,hishmehDeerHeadlightsShort2020,221016567DeFIXDetecting}
\\
\midrule

\textbf{Speed Control} &  Efficiency & Dense&\cite{Versatile_and_Efficien,Think2Drive,GRI:General_Reinforced_Imitation,End-to-End_Model-Free_Reinforcement_Learning_for_Urban_Driving,Vision-Based_Autonomous_Car_Racing_Using_Deep_Imitative_Reinforcement_Learning,End-to-End_Urban_Driving_by_Imitating_a_Reinforcement_Learning_Coach,Learning_Urban_Driving_Policies_using_Deep_Reinforcement_Learning,Efficient_Learning_of_Urban_Driving,Learning_to_drive_from_a_world_on_rails,Reinforced_Curriculum_Learning_For_Autonomous_Driving,Autonomous_Driving_via_Knowledge-Enhanced,A_Decision_Control_Method,Decision_Making_for_Autonomous_Driving_Via_Multimodal_Transforme,Deep_Reinforcement_Learning_for_Autonomous_Vehicle_Intersection_Navigation,Deductive_Reinforcement_Learning,Safe_Navigation_Based_on_Deep_Q-Network_Algorithm_Using,A_Bayesian_Approach_to_Reinforcement_Learning_of_Vision-Based_Vehicular_Control,Integration_of_Motion_Prediction_with_End-to-end_Latent_RL_for_Self-Driving_Vehicles,Addressing_Lane_Keeping_and_Intersections,Using_the_CARLA_Simulator_to_Train_A_Deep_Q_Self-Driving,Integrating_Deep_Reinforcement_Learning_with_Model-based_Path_Planners_for_Automated_Driving,An_Integrated_Reward_Function,Hierarchical_Reinforcement_Learning_with_Successor_Representation_for_Intelligent_Vehicle_Collision_Avoidance_of_Dynamic_Pedestrian,Towards_Autonomous_Driving,A_Deep_Q-Network_Reinforcement_Learning-Based,Urban_Autonomous_Driving_of_Emergency_Vehicles_with_Reinforcement_Learning,wang_vision-based_2023, wei_continual_2023, wang_benchmarking_2021, carton_using_2021, khalil_exploiting_2023, wu_lane_2022, chen_motion_2024, deng_context_2023, savari_online_2021, xu_end--end_2023, zhao_end--end_2024, yang_decision-making_2022, bai_fen-dqn_2023, muhammed_developing_2021, youssef_deep_2019, aghdasian_autonomous_2023, huang_simoun_2023, peng_imitative_2020,gutierrez-morenoDecisionMakingAutonomous2024,igoeMultiAlphaSoftActorCritic2023,heAdaptiveDecisionMaking2022,zhangSelflearningLaneKeeping2021,jaafraContextAwareAutonomousDriving2019,221016567DeFIXDetecting,gupta_hylear_2023,deshpande_navigation_2021,albilani_guided_2023,martinez_gomez_temporal_2023,jin_vwpefficient_2024,li_decision-making_2024,baheri_vision-based_2020,shi_efficient_2023,ahmed_policy-based_2022}\\
\midrule

\textbf{Deviation} & Safety & Dense & \cite{Versatile_and_Efficien}, \cite{Think2Drive}, \cite{Deep_reinforcement_learning_based_control_for_Autonomous_Vehicles_in_CARLA,GRI:General_Reinforced_Imitation,End-to-End_Model-Free_Reinforcement_Learning_for_Urban_Driving,Vision-Based_Autonomous_Car_Racing_Using_Deep_Imitative_Reinforcement_Learning,End-to-End_Urban_Driving_by_Imitating_a_Reinforcement_Learning_Coach,Learning_Urban_Driving_Policies_using_Deep_Reinforcement_Learning,Efficient_Learning_of_Urban_Driving,Learning_to_drive_from_a_world_on_rails,Reinforced_Curriculum_Learning_For_Autonomous_Driving,Autonomous_Driving_via_Knowledge-Enhanced,A_Decision_Control_Method,Decision_Making_for_Autonomous_Driving_Via_Multimodal_Transforme,Deductive_Reinforcement_Learning,Deep_Reinforcement_Learning_Control_Strategy_at_Roundabout_for_i-CAR,Integration_of_Motion_Prediction_with_End-to-end_Latent_RL_for_Self-Driving_Vehicles,Addressing_Lane_Keeping_and_Intersections,Reinforcement_Learning-Based_Guidance_of_Autonomous_Vehicles,An_Integrated_Reward_Function,Towards_Autonomous_Driving,A_Deep_Q-Network_Reinforcement_Learning-Based,Urban_Autonomous_Driving_of_Emergency_Vehicles_with_Reinforcement_Learning,carton_using_2021, chen_motion_2024, goel_adaptive_2021, xu_end--end_2023, frauenknecht_data-efficient_2023, wu_proximal_2023, youssef_deep_2019, aghdasian_autonomous_2023, liao_lateral_2023, huang_simoun_2023,hanLearningDriveUsing2022,igoeMultiAlphaSoftActorCritic2023,zhangSelflearningLaneKeeping2021,tsaiAutonomousVehicleFollowingTechnique2023,gangopadhyayHierarchicalProgramTriggeredReinforcement2022}\\
\midrule

\textbf{Goal Reward} & 	Task Completion & Sparse &\cite{Deep_Reinforcement_Learning_for_Autonomous_Vehicle_Intersection_Navigation,A_Hierarchical_Autonomous_Driving_Framework,A_Reinforcement_Learning_Based_Approach_for_Controlling_Autonomous_Vehicles_in_Complex_Scenarios,Using_the_CARLA_Simulator_to_Train_A_Deep_Q_Self-Driving,Integrating_Deep_Reinforcement_Learning_with_Model-based_Path_Planners_for_Automated_Driving,Reinforcement_Learning-Based_Guidance_of_Autonomous_Vehicles,Hierarchical_Reinforcement_Learning_with_Successor_Representation_for_Intelligent_Vehicle_Collision_Avoidance_of_Dynamic_Pedestrian,Deep-Reinforcement-Learning-Based_Driving_Policy_at_Intersections,jia_safe_2023, chen_motion_2024, wu_reinforcement_2023, frauenknecht_data-efficient_2023, bai_fen-dqn_2023, muhammed_developing_2021, huang_simoun_2023,gutierrez-morenoDecisionMakingAutonomous2024,hanLearningDriveUsing2022,heAdaptiveDecisionMaking2022,naveedTrajectoryPlanningAutonomous2020,gangopadhyayHierarchicalProgramTriggeredReinforcement2022,gupta_hylear_2023,hussonnois_end--end_2022,li_decision-making_2024,baheri_vision-based_2020,shi_efficient_2023,udatha_reinforcement_2023}
\\
\midrule

\textbf{Heading Alignment} & Safety & Dense& \cite{Versatile_and_Efficien,GRI:General_Reinforced_Imitation,End-to-End_Model-Free_Reinforcement_Learning_for_Urban_Driving,Vision-Based_Autonomous_Car_Racing_Using_Deep_Imitative_Reinforcement_Learning,End-to-End_Urban_Driving_by_Imitating_a_Reinforcement_Learning_Coach,Learning_to_drive_from_a_world_on_rails,Reinforced_Curriculum_Learning_For_Autonomous_Driving,Autonomous_Driving_via_Knowledge-Enhanced,A_Decision_Control_Method,Decision_Making_for_Autonomous_Driving_Via_Multimodal_Transforme,Deductive_Reinforcement_Learning,Deep_Reinforcement_Learning_Control_Strategy_at_Roundabout_for_i-CAR,Addressing_Lane_Keeping_and_Intersections,Reinforcement_Learning-Based_Guidance_of_Autonomous_Vehicles,An_Integrated_Reward_Function,Towards_Autonomous_Driving,A_Deep_Q-Network_Reinforcement_Learning-Based,Urban_Autonomous_Driving_of_Emergency_Vehicles_with_Reinforcement_Learning,jia_safe_2023, carton_using_2021, khalil_exploiting_2023, mohammed_unified_2024, chen_motion_2024, zhang_toward_2022, manikandan_ad_2023, wu_proximal_2023,hanLearningDriveUsing2022,gangopadhyayHierarchicalProgramTriggeredReinforcement2022,frijiDQNBasedAutonomousCarFollowing2020,albilani_guided_2023,jin_vwpefficient_2024}\\
\midrule

\textbf{Steering Cost} & Comfort & Dense& \cite{Think2Drive,End-to-End_Urban_Driving_by_Imitating_a_Reinforcement_Learning_Coach,Efficient_Learning_of_Urban_Driving,Model-Based_Reinforcement_Learning_with_Isolated_Imaginations,Autonomous_Driving_via_Knowledge-Enhanced,Decision_Making_for_Autonomous_Driving_Via_Multimodal_Transforme,Deductive_Reinforcement_Learning,Integration_of_Motion_Prediction_with_End-to-end_Latent_RL_for_Self-Driving_Vehicles,wang_vision-based_2023, khalil_exploiting_2023, deng_context_2023, xu_end--end_2023, deng_context-aware_2024, frauenknecht_data-efficient_2023, yang_decision-making_2022, youssef_deep_2019, liao_lateral_2023, peng_imitative_2020,gupta_hylear_2023,jin_vwpefficient_2024,hishmehDeerHeadlightsShort2020}\\
\midrule

\textbf{Jerk Penalty} & Comfort  & Dense & \cite{xu_decision-making_2022,wei_continual_2023,hishmehDeerHeadlightsShort2020,shi_efficient_2023} \\
\midrule

\textbf{Brake Cost/Reward} & Comfort &Dense & \cite{wu_proximal_2023,xu_end--end_2023,bai_fen-dqn_2023,Learning_to_drive_from_a_world_on_rails,Safe_Navigation_Based_on_Deep_Q-Network_Algorithm_Using}\\
\midrule

\textbf{Lane Invasion} & Safety & Dense/ Sparse &\cite{Versatile_and_Efficien,Deep_reinforcement_learning_based_control_for_Autonomous_Vehicles_in_CARLA,Learning_Urban_Driving_Policies_using_Deep_Reinforcement_Learning,A_Methodology_Based_on_Deep_Reinforcement_Learning,A_Reinforcement_Learning_Based_Approach_for_Controlling_Autonomous_Vehicles_in_Complex_Scenarios} \\
\midrule

\textbf{Offroad} & Safety & Dense &\cite{Deep_Reinforcement_Learning_for_Autonomous_Vehicle_Intersection_Navigation,Hierarchical_Reinforcement_Learning_with_Successor_Representation_for_Intelligent_Vehicle_Collision_Avoidance_of_Dynamic_Pedestrian,Urban_Autonomous_Driving_of_Emergency_Vehicles_with_Reinforcement_Learning,wang_vision-based_2023, carton_using_2021, savari_online_2021, youssef_deep_2019,hishmehDeerHeadlightsShort2020,baheri_vision-based_2020} \\
\midrule

%\textbf{Time-to-Collision} & Penalizes states with high likelihood of imminent collisions based on predicted time-to-impact. \cite{jia_safe_2023,wei_continual_2023}\\

\textbf{Timeout} & Efficiency &Sparse& \cite{wu_reinforcement_2023,zhao_end--end_2024,Urban_Autonomous_Driving_of_Emergency_Vehicles_with_Reinforcement_Learning,Deep-Reinforcement-Learning-Based_Driving_Policy_at_Intersections} \\
\midrule

\textbf{Infraction} & Safety & Dense & \cite{deng_context-aware_2024},\cite{End-to-End_Urban_Driving_by_Imitating_a_Reinforcement_Learning_Coach,Learning_Urban_Driving_Policies_using_Deep_Reinforcement_Learning,Efficient_Learning_of_Urban_Driving,221016567DeFIXDetecting,shi_efficient_2023,tsaiAutonomousVehicleFollowingTechnique2023}\\
\midrule

\textbf{Travel} & Efficiency/Progress & Dense & \cite{Think2Drive}, \cite{Deep_Reinforcement_Learning_for_Autonomous_Vehicle_Intersection_Navigation,A_Reinforcement_Learning_Based_Approach_for_Controlling_Autonomous_Vehicles_in_Complex_Scenarios,Using_the_CARLA_Simulator_to_Train_A_Deep_Q_Self-Driving,Reinforcement_Learning-Based_Guidance_of_Autonomous_Vehicles,wang_vision-based_2023, xu_decision-making_2022, wei_continual_2023, khalil_exploiting_2023, xu_end--end_2023, deng_context-aware_2024, wu_reinforcement_2023,naveedTrajectoryPlanningAutonomous2020,jaafraContextAwareAutonomousDriving2019}\\
\midrule
%\textbf{Time-Headway} & \\

\textbf{Lane Change} & Safety/Efficiency & Sparse & \cite{A_Decision_Control_Method,wang_benchmarking_2021,wu_lane_2022}\\
\midrule
\textbf{Close-Obstacle Distance} & Safety & Dense & \cite{Towards_Autonomous_Driving,A_Deep_Q-Network_Reinforcement_Learning-Based,manikandan_ad_2023} \\
\midrule
%\textbf{Dangerous Action} & Penalizes unsafe behaviors like sharp turns at high speed or risky overtakes. \\

\textbf{Speed Penalty} & Safety & Dense &  \cite{wang_vision-based_2023, xu_decision-making_2022, mohammed_unified_2024, xu_end--end_2023, deng_context-aware_2024, wu_reinforcement_2023, zhao_end--end_2024, muhammed_developing_2021,Integration_of_Motion_Prediction_with_End-to-end_Latent_RL_for_Self-Driving_Vehicles,Learning_Automated_Driving_in_Complex_Intersection_Scenarios_Based,Urban_Autonomous_Driving_of_Emergency_Vehicles_with_Reinforcement_Learning,hishmehDeerHeadlightsShort2020,221016567DeFIXDetecting,hishmehDeerHeadlightsShort2020,221016567DeFIXDetecting} \\
\midrule
\textbf{Others} &  N/A & N/A &\cite{A_Hierarchical_Autonomous_Driving_Framework,A_Bayesian_Approach_to_Reinforcement_Learning_of_Vision-Based_Vehicular_Control,wei_continual_2023,savari_online_2021,jia_safe_2023,noauthor_explainable_nodate,li_decision-making_2024,martinez_gomez_temporal_2023,hussonnois_end--end_2022,gupta_hylear_2023,heAdaptiveDecisionMaking2022,tsaiAutonomousVehicleFollowingTechnique2023,naveedTrajectoryPlanningAutonomous2020,gangopadhyayHierarchicalProgramTriggeredReinforcement2022,jaafraContextAwareAutonomousDriving2019}

 \\
 
%\textbf{N/A} & Reward function not clearly specified or categorized in the paper. Placeholder for ambiguous cases. \cite{Deep_Reinforcement_Learning_based_control_algorithms:_Training_and_validation_using_the_ROS_Framework}, \cite{Domain_Adaptation_In_Reinforcement_Learning} , \cite{Enhanced_Off-Policy_Reinforcement_Learning_With_Focused_Experience_Replay} , \cite{Boosting_Offline_Reinforcement_Learning}, \cite{zhu_rita_2023, chronis_driving_2021, cheng_longitudinal_2021},\cite{coutoHierarchicalGenerativeAdversarial2024}\\

\bottomrule
\end{tabular}
\end{adjustbox}
\end{table*}

\begin{table}[ht]
  \centering
  \renewcommand{\arraystretch}{0.9} % tighten row height
  \setlength{\tabcolsep}{6pt}       % tighten column separation
  \caption{Definitions of Specialized Reward Terms}

    %\begin{adjustbox}{max width=\textwidth}
   % \begin{adjustbox}{max width=\columnwidth}

  \label{tab:reward_glossary}
  \begin{tabular}{p{2.5cm} p{8cm}}
    \toprule
    \textbf{Term} & \textbf{Definition} \\ 
    \midrule
    Collision & Penalizes any collision with other vehicles, pedestrians, or objects to ensure safe driving. \\
    \addlinespace
    
    Speed Control & Rewards the agent for maintaining a desired speed or penalizes deviation from a target speed range. Used to balance safety and efficiency. \\
    \addlinespace
    
    Deviation & Penalizes the lateral offset from the lane center or planned trajectory to promote lane-keeping behavior. \\
    \addlinespace
    
    Goal Reward & Provides a positive reward when the agent reaches its destination or finishing the task at hand. Encourages task completion. \\
    \addlinespace
    
    Heading Alignment & Rewards alignment between the vehicle's heading and the direction of the lane or target waypoint.\\
    \addlinespace

    Steering Cost & Penalizes abrupt or large steering commands to encourage smoother control and comfort.\\
    \addlinespace

    Jerk & Penalizes rapid changes in acceleration or steering, promoting smoother and more comfortable driving.\\
    \addlinespace

    Brake Cost/Reward & Penalizes unnecessary or hard braking (cost) or rewards proper and timely use of braking (reward), aiming to reduce discomfort and unsafe stops.\\
    \addlinespace

    Lane Invasion & Penalizes crossing over lane boundaries without intention, reflecting rule violation or unsafe maneuvering.\\
    \addlinespace

    Offroad & Penalizes driving off the drivable road area, promoting adherence to lane and road limits.\\
    \addlinespace

    Timeout & Penalizes failure to reach the goal within a time limit, encouraging efficient navigation.\\
    \addlinespace

    Infraction & Penalizes the agent for violating traffic rules like running red lights, ignoring stop signs, etc.\\
    \addlinespace

    Travel & Encourages consistent forward movement or progress through the environment based on the distance to the goal location.\\
    \addlinespace

    Lane Change & Rewards safe and necessary lane changes or penalizes unnecessary or unsafe ones.\\
    \addlinespace

    Close-Obstacle Distance & Rewards the agent for maintaining a safe distance from nearby obstacles while navigating close environments.\\
    \addlinespace

    Speed Penalty & Penalizes speeding or exceeding safe velocity thresholds. Often complements speed control rewards. It can instead penalize the car not moving at all.\\
    %\addlinespace
    %\dots & \dots \\ 
    \bottomrule
  \end{tabular}
%  \end{adjustbox}
\end{table}
The definition of the most popular reward terms are mentioned in Table \ref{tab:reward_glossary}.
Several studies introduce a diverse set of specialized reward components beyond the core categories to capture high-level commands, safety distances, comfort, efficiency, and rule compliance. \emph{Command rewards} incentivize correct execution of navigational instructions—such as turning at intersections—in hierarchical frameworks \cite{A_Hierarchical_Autonomous_Driving_Framework}. \emph{Opposite-lane penalties} discourage entering or remaining in lanes of oncoming traffic \cite{A_Bayesian_Approach_to_Reinforcement_Learning_of_Vision-Based_Vehicular_Control}. \emph{Time-headway rewards} promote maintaining a safe time gap behind lead vehicles to avoid tailgating \cite{wei_continual_2023}, while \emph{time-to-collision penalties} penalize states with high imminent-collision risk \cite{jia_safe_2023,wei_continual_2023}. \emph{Dangerous-action penalties} discourage deviations from expert human behavior \cite{savari_online_2021}. \emph{Collision-free-driving bonuses} provide a small positive reward for each collision-free timestep \cite{noauthor_explainable_nodate}. \emph{Risk-reward terms} encode the time difference between the ego vehicle’s departure and a social vehicle’s arrival at a conflict area \cite{li_decision-making_2024}, while \emph{time-differential rewards} align maneuvers temporally with available merge slots \cite{martinez_gomez_temporal_2023}. \emph{Waypoint-following rewards} grant a bonus upon reaching designated waypoints \cite{hussonnois_end--end_2022}. \emph{Failure penalties} impose costs for not reaching the goal, and \emph{near-miss penalties} penalize moderate negative reward when pedestrians enter a predefined safety buffer \cite{gupta_hylear_2023}. \emph{Fast-completion rewards} encourage rapid task execution \cite{heAdaptiveDecisionMaking2022}. \emph{Steering or throttle jitter penalties} discourage abrupt control changes to ensure smoother, more comfortable driving \cite{tsaiAutonomousVehicleFollowingTechnique2023}. \emph{Safe-dist-to-lead-vehicle terms} maintain a safe following distance from the car ahead \cite{tsaiAutonomousVehicleFollowingTechnique2023,naveedTrajectoryPlanningAutonomous2020}, while \emph{regular-timestep penalties and progress rewards} balance efficiency with purposeful movement \cite{naveedTrajectoryPlanningAutonomous2020}. \emph{Unsmoothness penalties} discourage maneuvers that conflict with prior velocity profiles, promoting smoother transitions \cite{naveedTrajectoryPlanningAutonomous2020}. \emph{Dist-to-lead-vehicle metrics} quantify the longitudinal gap to the lead vehicle \cite{gangopadhyayHierarchicalProgramTriggeredReinforcement2022}, and \emph{dist-to-nearest-obj rewards} promote safe clearance from nearby obstacles \cite{gangopadhyayHierarchicalProgramTriggeredReinforcement2022}. Finally, \emph{incorrect-lane} and \emph{incorrect-steer penalties} discourage misaligned lane occupancy and excessive steering during maneuvers \cite{gangopadhyayHierarchicalProgramTriggeredReinforcement2022}, and \emph{drive-side-walk penalties} ensure adherence to drivable space by penalizing overlap with sidewalks \cite{jaafraContextAwareAutonomousDriving2019}.

\section{Terminal Condition}
In RL, terminal conditions define when an episode ends. In CARLA-based driving tasks, terminal conditions are essential for shaping episode boundaries and determining when an agent has completed a task successfully or failed due to violations. Common terminal triggers include collisions, lane invasions, reaching a goal location, going off-road, timeouts, or violating traffic rules. Properly chosen terminal conditions ensure that the learning process focuses on meaningful behaviors while penalizing unsafe or inefficient actions. They also influence the agent’s return estimates and training stability, particularly in sparse-reward settings. Table \ref{tab:termination_conditions} summarized the most common terminal conditions used in the literature along with references. Fig. \ref{fig:terminal_distribution} illustrates the distribution of terminal conditions.

\begin{figure}[htbp]
    \centering
    \includegraphics[width=0.5\textwidth]{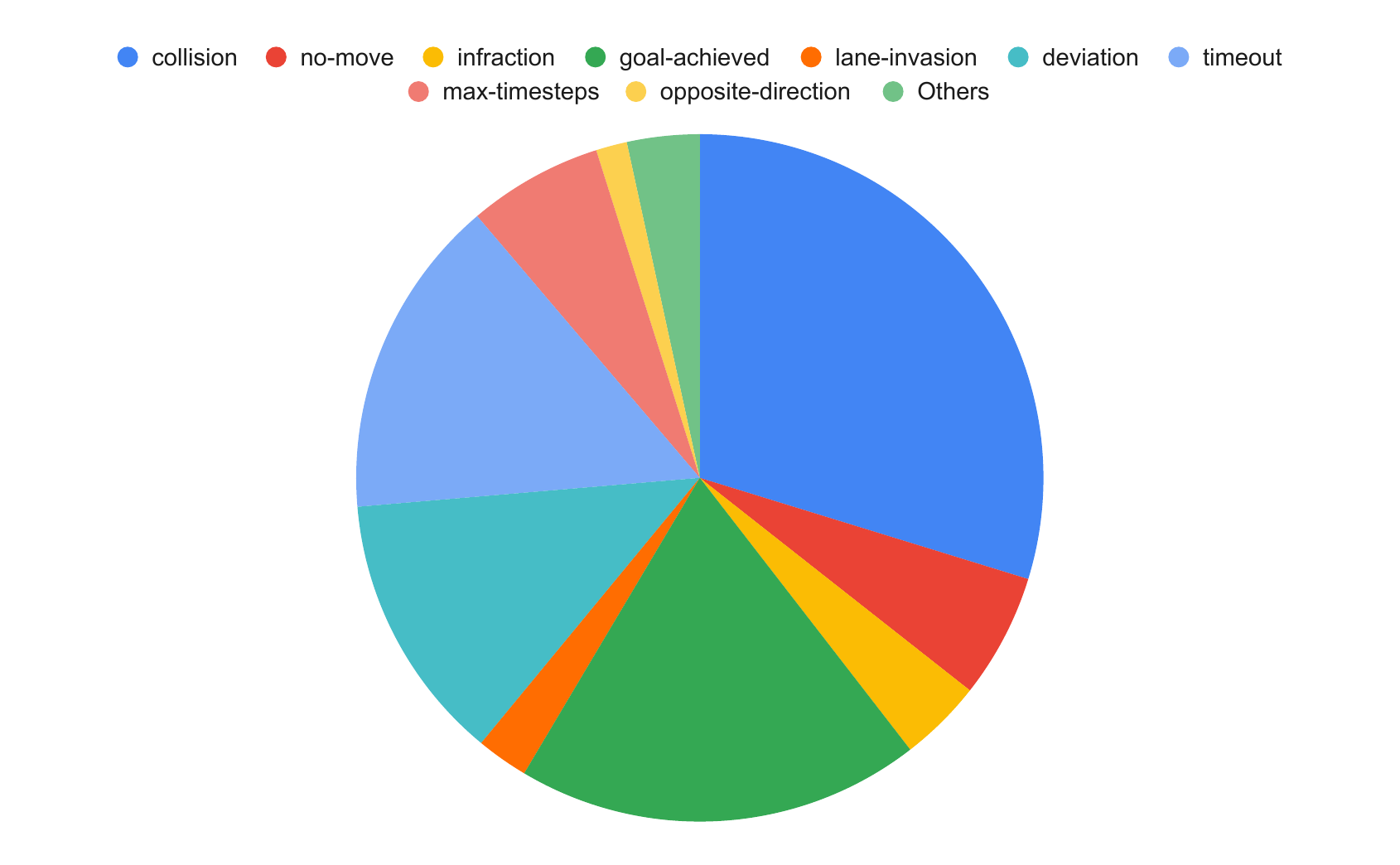} % Adjust path and size
    \caption{Distribution of terminal conditions used in RL for autonomous driving tasks.}
    \label{fig:terminal_distribution}
\end{figure}

\begin{table*}[htbp]
\centering
\small
\caption{Common Termination Conditions in Autonomous Driving Scenarios (with example citations).}
\label{tab:termination_conditions}
%\begin{adjustbox}{max width=\columnwidth}
\begin{adjustbox}{max width=\textwidth}
\begin{tabular}{p{2cm} p{1cm} p{6cm} p{5cm}} % Adjust column width as needed
\toprule
\textbf{Condition} & \textbf{Fatal} & \textbf{Description} & \textbf{References} \\
\midrule

Collision 
& Y & The ego vehicle’s bounding box intersects with another object 
(e.g., another vehicle, pedestrian, barrier). 
Triggers include rear-end or side-impact collisions. &   
\cite{Versatile_and_Efficien,Think2Drive,Deep_Reinforcement_Learning_based_control_algorithms:_Training_and_validation_using_the_ROS_Framework,
End-to-End_Model-Free_Reinforcement_Learning_for_Urban_Driving,
End-to-End_Urban_Driving_by_Imitating_a_Reinforcement_Learning_Coach,
Deep_Reinforcement_Learning_for_Autonomous_Vehicle_Intersection_Navigation,
Deep_Reinforcement_Learning_Control_Strategy_at_Roundabout_for_i-CAR,
Integration_of_Motion_Prediction_with_End-to-end_Latent_RL_for_Self-Driving_Vehicles,
A_Reinforcement_Learning_Based_Approach_for_Controlling_Autonomous_Vehicles_in_Complex_Scenarios,
Using_the_CARLA_Simulator_to_Train_A_Deep_Q_Self-Driving,
Integrating_Deep_Reinforcement_Learning_with_Model-based_Path_Planners_for_Automated_Driving,
Reinforcement_Learning-Based_Guidance_of_Autonomous_Vehicles,
A_Deep_Q-Network_Reinforcement_Learning-Based,
Urban_Autonomous_Driving_of_Emergency_Vehicles_with_Reinforcement_Learning,
wang_vision-based_2023, jia_safe_2023, xu_decision-making_2022, 
wei_continual_2023, wang_benchmarking_2021, carton_using_2021,
khalil_exploiting_2023, wu_lane_2022, mohammed_unified_2024,
chen_motion_2024, deng_context_2023, zhang_toward_2022, goel_adaptive_2021,
manikandan_ad_2023, xu_end--end_2023, cheng_longitudinal_2021,
deng_context-aware_2024, wu_reinforcement_2023, zhao_end--end_2024,
yang_decision-making_2022, bai_fen-dqn_2023, wu_proximal_2023,
muhammed_developing_2021, youssef_deep_2019, aghdasian_autonomous_2023,
liao_lateral_2023, huang_simoun_2023, peng_imitative_2020,
elallid_dqn-based_2022,gupta_hylear_2023,hussonnois_end--end_2022,deshpande_navigation_2021,martinez_gomez_temporal_2023,li_decision-making_2024,baheri_vision-based_2020,shi_efficient_2023,ahmed_policy-based_2022,udatha_reinforcement_2023,noauthor_explainable_nodate,tsaiAutonomousVehicleFollowingTechnique2023,frijiDQNBasedAutonomousCarFollowing2020,jaafraContextAwareAutonomousDriving2019,hishmehDeerHeadlightsShort2020,221016567DeFIXDetecting}\\
\addlinespace

No-move
& N & The ego vehicle remains stationary or moves below a minimal threshold (speed or distance) 
for a set duration or number of timesteps. 
Common examples include failing to move for over 30--60 seconds. &   \cite{Versatile_and_Efficien,End-to-End_Model-Free_Reinforcement_Learning_for_Urban_Driving,
End-to-End_Urban_Driving_by_Imitating_a_Reinforcement_Learning_Coach,
wang_vision-based_2023,carton_using_2021,deng_context_2023,
manikandan_ad_2023,wu_reinforcement_2023,zhao_end--end_2024,jin_vwpefficient_2024,albilani_guided_2023}\\
\addlinespace

Infraction 
& Y & The ego vehicle violates traffic rules (e.g., running red lights, ignoring stop signs, 
illegal turns). Sometimes includes driving on the wrong side of the road.  &   \cite{Think2Drive,End-to-End_Model-Free_Reinforcement_Learning_for_Urban_Driving,
End-to-End_Urban_Driving_by_Imitating_a_Reinforcement_Learning_Coach,
Learning_Urban_Driving_Policies_using_Deep_Reinforcement_Learning,
deng_context_2023,deng_context-aware_2024,shi_efficient_2023,coutoHierarchicalGenerativeAdversarial2024}\\
\addlinespace

Goal-achieved 
& N & The ego vehicle successfully reaches its target destination or completes the route 
(e.g., arriving at a waypoint or parking spot). &   \cite{Think2Drive,Learning_Urban_Driving_Policies_using_Deep_Reinforcement_Learning,
Deep_Reinforcement_Learning_for_Autonomous_Vehicle_Intersection_Navigation,
A_Reinforcement_Learning_Based_Approach_for_Controlling_Autonomous_Vehicles_in_Complex_Scenarios,
Using_the_CARLA_Simulator_to_Train_A_Deep_Q_Self-Driving,
Integrating_Deep_Reinforcement_Learning_with_Model-based_Path_Planners_for_Automated_Driving,
Reinforcement_Learning-Based_Guidance_of_Autonomous_Vehicles,
wang_vision-based_2023,jia_safe_2023,xu_decision-making_2022,
wu_lane_2022,mohammed_unified_2024,chen_motion_2024,
xu_end--end_2023,deng_context-aware_2024,wu_reinforcement_2023,
bai_fen-dqn_2023,muhammed_developing_2021,youssef_deep_2019,
aghdasian_autonomous_2023,liao_lateral_2023, peng_imitative_2020,
elallid_dqn-based_2022,gupta_hylear_2023,hussonnois_end--end_2022,deshpande_navigation_2021,albilani_guided_2023,martinez_gomez_temporal_2023,jin_vwpefficient_2024,li_decision-making_2024,li_dynamic_2022,baheri_vision-based_2020,shi_efficient_2023,ahmed_policy-based_2022,udatha_reinforcement_2023,gangopadhyayHierarchicalProgramTriggeredReinforcement2022,frijiDQNBasedAutonomousCarFollowing2020,jaafraContextAwareAutonomousDriving2019}\\
\addlinespace

Lane-invasion
& Y & The ego vehicle crosses lane markings or invades adjacent lanes in an unintended manner. 
Sometimes used interchangeably with deviation if crossing lane boundaries is disallowed. &  \cite{Deep_Reinforcement_Learning_based_control_algorithms:_Training_and_validation_using_the_ROS_Framework,
Domain_Adaptation_In_Reinforcement_Learning,Autonomous_Driving_via_Knowledge-Enhanced,
frauenknecht_data-efficient_2023,coutoHierarchicalGenerativeAdversarial2024}\\
\addlinespace

Deviation
& Y & The ego vehicle strays from the intended path or lane centre beyond a tolerance 
(e.g., 2--3\,m), or drives off-road entirely. &  \cite{End-to-End_Model-Free_Reinforcement_Learning_for_Urban_Driving,
End-to-End_Urban_Driving_by_Imitating_a_Reinforcement_Learning_Coach,
Domain_Adaptation_In_Reinforcement_Learning,Autonomous_Driving_via_Knowledge-Enhanced,
A_Decision_Control_Method,An_Integrated_Reward_Function,
Urban_Autonomous_Driving_of_Emergency_Vehicles_with_Reinforcement_Learning,
wang_vision-based_2023,jia_safe_2023,wei_continual_2023,
wang_benchmarking_2021,carton_using_2021,khalil_exploiting_2023,
wu_lane_2022,deng_context_2023,savari_online_2021,
xu_end--end_2023,deng_context-aware_2024,wu_reinforcement_2023,
zhao_end--end_2024,bai_fen-dqn_2023,wu_proximal_2023,muhammed_developing_2021,
aghdasian_autonomous_2023,liao_lateral_2023,peng_imitative_2020,jin_vwpefficient_2024,albilani_guided_2023,li_dynamic_2022}\\
\addlinespace

Timeout
& N & The scenario’s simulation clock expires (e.g., 8\,s to cross an intersection, 
300\,s to complete a route) without collision or reaching the goal. &  \cite{Learning_Urban_Driving_Policies_using_Deep_Reinforcement_Learning,
Cognitive_Reinforcement_Learning,Autonomous_Driving_via_Knowledge-Enhanced,
Deep_Reinforcement_Learning_Control_Strategy_at_Roundabout_for_i-CAR,
Urban_Autonomous_Driving_of_Emergency_Vehicles_with_Reinforcement_Learning,
wang_benchmarking_2021,carton_using_2021,khalil_exploiting_2023,
mohammed_unified_2024,chen_motion_2024,xu_decision-making_2022,
xu_end--end_2023,cheng_longitudinal_2021,wu_reinforcement_2023,
yang_decision-making_2022,youssef_deep_2019,gupta_hylear_2023,hussonnois_end--end_2022,martinez_gomez_temporal_2023,li_decision-making_2024,li_dynamic_2022,baheri_vision-based_2020,shi_efficient_2023,ahmed_policy-based_2022,udatha_reinforcement_2023,noauthor_explainable_nodate,hishmehDeerHeadlightsShort2020,221016567DeFIXDetecting}\\
\addlinespace

Max-timesteps 
& N & A hard limit on the total number of simulator steps (e.g., 800, 1000, or 3000 steps). 
Once reached, the episode terminates automatically. &  \cite{Domain_Adaptation_In_Reinforcement_Learning,
Model-Based_Reinforcement_Learning_with_Isolated_Imaginations,
Deep_Reinforcement_Learning_for_Autonomous_Vehicle_Intersection_Navigation,
Integration_of_Motion_Prediction_with_End-to-end_Latent_RL_for_Self-Driving_Vehicles,
A_Reinforcement_Learning_Based_Approach_for_Controlling_Autonomous_Vehicles_in_Complex_Scenarios,
chronis_driving_2021,zhang_toward_2022,frauenknecht_data-efficient_2023,
bai_fen-dqn_2023,muhammed_developing_2021,huang_simoun_2023,
peng_imitative_2020,zhao_end--end_2024} \\
\addlinespace

Opposite-direction 
& Y & The ego vehicle orients and/or drives 180$^\circ$ against the valid heading, 
often leading to traveling into oncoming traffic. & \cite{Reinforcement_Learning-Based_Guidance_of_Autonomous_Vehicles,
peng_imitative_2020} \\
\addlinespace

Others
& N/A & Various custom or less common triggers, such as 
\emph{excess-speed} (driving well above limits), 
\emph{over-pass-destination} (traveling beyond the target), 
\emph{stuck/blocking} (blocked in traffic or making no progress), 
\emph{excessive-low-reward} (accumulating large negative rewards), 
\emph{no-improvement} (no training progress for many episodes), 
\emph{following-failure}, etc. &  \cite{Reinforcement_Learning-Based_Guidance_of_Autonomous_Vehicles,
wang_vision-based_2023,chronis_driving_2021,wei_continual_2023,zhao_end--end_2024} \\
\addlinespace

\bottomrule
\end{tabular}
\end{adjustbox}
\end{table*}

\section{Evaluation Metric}

Evaluation metrics are critical for benchmarking RL agents in autonomous driving, offering quantitative insights into performance across diverse tasks and environments. The reviewed literature reveals a broad spectrum of metrics that reflect priorities such as safety, task completion, control stability, behavioral realism, and training efficiency. These metrics are often aligned with specific objectives—whether ensuring collision-free navigation, adherence to traffic rules, or replicating human-like behavior in maneuvers such as lane-following and intersection handling.

Based on our analysis, we categorize the metrics into four primary groups: (1) task completion and task progress (e.g., success rate, route completion, driving distance); (2) safety and comfort (e.g., collision rate, infraction score, lane deviation); (3) Efficiency (e.g., completion time, average speed); and (4) overall-preformance and overall driving quality (e.g., reward, driving score). %Composite or domain-specific metrics—such as the driving score or critical score—also appear in several works to aggregate multi-objective criteria into a unified evaluation. 
Fig.~\ref{fig:evaluation_distribution} provides an overview of the distribution of evaluation metrics across the surveyed literature, while Table~\ref{tab:evaluation_metrics} summarizes the most commonly used metrics along with representative studies that adopt them.

\begin{figure}[htbp]
    \centering
    \includegraphics[width=0.5\textwidth]{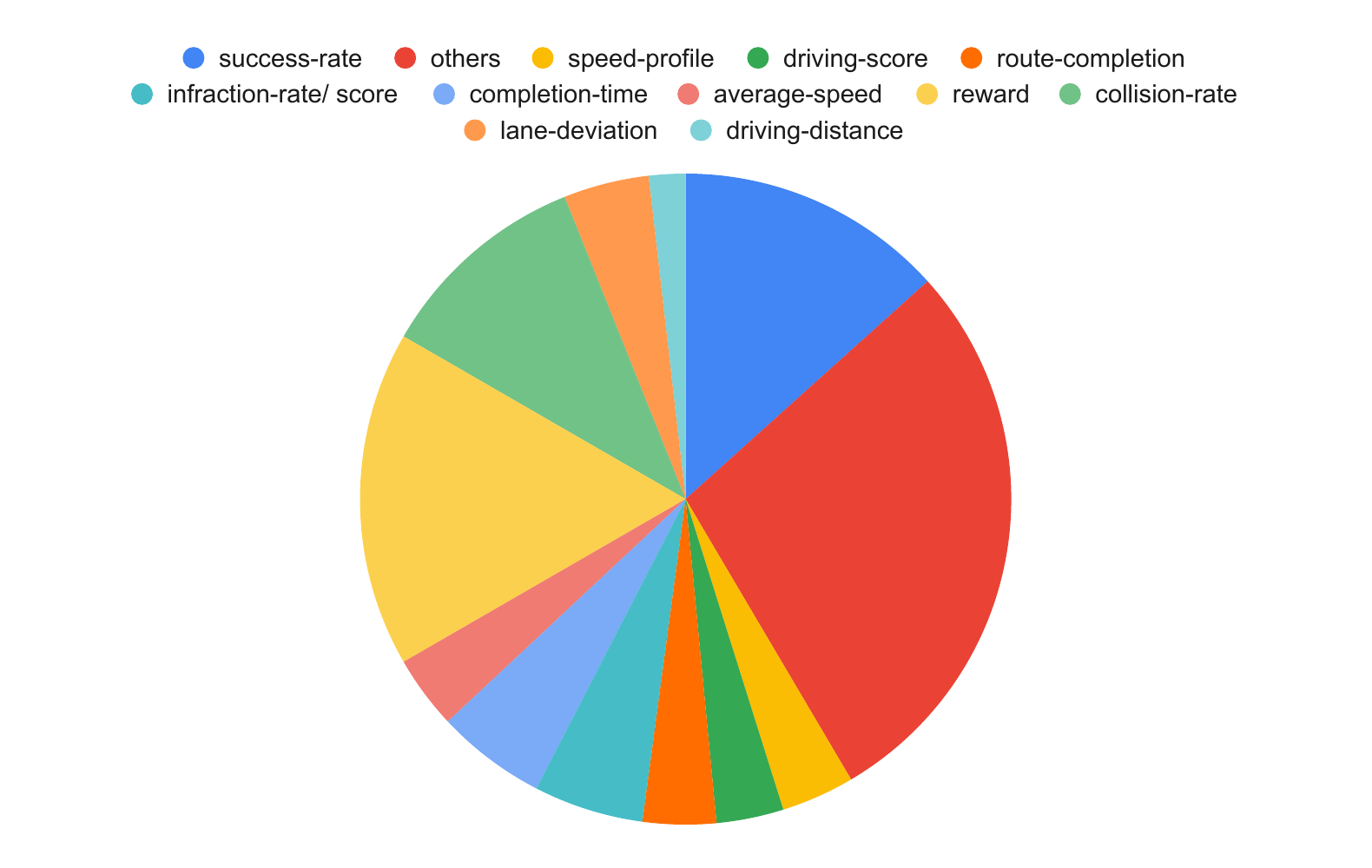} % Adjust path and size
    \caption{Distribution of evaluation metrics used in RL for autonomous driving tasks.}
    \label{fig:evaluation_distribution}
\end{figure}

\begin{table*}[htbp]
\centering
\small
\caption{Common Evaluation Metrics Used in RL for Autonomous Driving}
\label{tab:evaluation_metrics}

\begin{adjustbox}{max width=\textwidth}
\begin{tabular}{p{3cm} p{3cm} p{7cm} p{5.5cm}} % Adjust column width as needed
\toprule
\textbf{Metric} & \textbf{Aspect} & \textbf{Description} &\textbf{References} \\
\midrule

Success Rate & Task-completion & Measures the percentage of episodes or routes completed without failure.&  \cite{Versatile_and_Efficien,End-to-End_Model-Free_Reinforcement_Learning_for_Urban_Driving,Vision-Based_Autonomous_Car_Racing_Using_Deep_Imitative_Reinforcement_Learning,End-to-End_Urban_Driving_by_Imitating_a_Reinforcement_Learning_Coach,Learning_Urban_Driving_Policies_using_Deep_Reinforcement_Learning,Learning_to_drive_from_a_world_on_rails,Autonomous_Driving_via_Knowledge-Enhanced,Deductive_Reinforcement_Learning,Safe_Navigation_Based_on_Deep_Q-Network_Algorithm_Using,A_Bayesian_Approach_to_Reinforcement_Learning_of_Vision-Based_Vehicular_Control,A_Hierarchical_Autonomous_Driving_Framework,Hierarchical_Reinforcement_Learning_with_Successor_Representation_for_Intelligent_Vehicle_Collision_Avoidance_of_Dynamic_Pedestrian,Boosting_Offline_Reinforcement_Learning,Urban_Autonomous_Driving_of_Emergency_Vehicles_with_Reinforcement_Learning,Deep-Reinforcement-Learning-Based_Driving_Policy_at_Intersections,gutierrez-morenoDecisionMakingAutonomous2024,hanLearningDriveUsing2022,heAdaptiveDecisionMaking2022,coutoHierarchicalGenerativeAdversarial2024,zhangSelflearningLaneKeeping2021,naveedTrajectoryPlanningAutonomous2020,frijiDQNBasedAutonomousCarFollowing2020,wang_vision-based_2023,
wei_continual_2023,
wang_benchmarking_2021,
carton_using_2021,
khalil_exploiting_2023,
wu_lane_2022,
chen_motion_2024,
deng_context_2023,
deng_context-aware_2024,
wu_reinforcement_2023,
bai_fen-dqn_2023,
youssef_deep_2019,
liao_lateral_2023,
peng_imitative_2020,
elallid_dqn-based_2022,xu_decision-making_2022,deshpande_navigation_2021,albilani_guided_2023,li_decision-making_2024,baheri_vision-based_2020,ahmed_policy-based_2022,udatha_reinforcement_2023}

\\
\addlinespace

Collision Rate & Safety & Tracks the frequency of collisions per episode or over the evaluation period. &   \cite{jia_safe_2023,
xu_decision-making_2022,
khalil_exploiting_2023,
wu_lane_2022,
mohammed_unified_2024,
chen_motion_2024,
zhang_toward_2022,
goel_adaptive_2021,
manikandan_ad_2023,
cheng_longitudinal_2021,
deng_context-aware_2024,
yang_decision-making_2022,
bai_fen-dqn_2023,
muhammed_developing_2021,
aghdasian_autonomous_2023,
huang_simoun_2023,
elallid_dqn-based_2022,gupta_hylear_2023,martinez_gomez_temporal_2023,jin_vwpefficient_2024,li_decision-making_2024,shi_efficient_2023,udatha_reinforcement_2023,Cognitive_Reinforcement_Learning,Reinforced_Curriculum_Learning_For_Autonomous_Driving,Deep_Reinforcement_Learning_for_Autonomous_Vehicle_Intersection_Navigation,Safe_Navigation_Based_on_Deep_Q-Network_Algorithm_Using,A_Reinforcement_Learning_Based_Approach_for_Controlling_Autonomous_Vehicles_in_Complex_Scenarios,Towards_Autonomous_Driving,A_Deep_Q-Network_Reinforcement_Learning-Based,Urban_Autonomous_Driving_of_Emergency_Vehicles_with_Reinforcement_Learning,hanLearningDriveUsing2022,coutoHierarchicalGenerativeAdversarial2024,naveedTrajectoryPlanningAutonomous2020,frijiDQNBasedAutonomousCarFollowing2020}

\\
\addlinespace

Route Completion & Task-progress & Indicates the percentage of the planned route successfully followed by the agent. &  \cite{Think2Drive,GRI:General_Reinforced_Imitation,End-to-End_Urban_Driving_by_Imitating_a_Reinforcement_Learning_Coach,Learning_to_drive_from_a_world_on_rails,Decision_Making_for_Autonomous_Driving_Via_Multimodal_Transforme,Boosting_Offline_Reinforcement_Learning,coutoHierarchicalGenerativeAdversarial2024,221016567DeFIXDetecting,mohammed_unified_2024,
deng_context_2023,
chronis_driving_2021,jin_vwpefficient_2024}

%\cite{Think2Drive}, \cite{GRI:General_Reinforced_Imitation}, \cite{End-to-End_Urban_Driving_by_Imitating_a_Reinforcement_Learning_Coach}, \cite{Learning_to_drive_from_a_world_on_rails}, \cite{Decision_Making_for_Autonomous_Driving_Via_Multimodal_Transforme}, \cite{Boosting_Offline_Reinforcement_Learning}, \cite{mohammed_unified_2024, deng_context_2023, chronis_driving_2021},\cite{coutoHierarchicalGenerativeAdversarial2024},\cite{221016567DeFIXDetecting},\cite{jin_vwpefficient_2024}. 

\\
\addlinespace

Infraction Rate / Score & Safety  & Captures violations such as off-road driving, collisions, or running red lights. &  \cite{Think2Drive,GRI:General_Reinforced_Imitation,End-to-End_Model-Free_Reinforcement_Learning_for_Urban_Driving,End-to-End_Urban_Driving_by_Imitating_a_Reinforcement_Learning_Coach,Learning_Urban_Driving_Policies_using_Deep_Reinforcement_Learning,Efficient_Learning_of_Urban_Driving,Learning_to_drive_from_a_world_on_rails,Decision_Making_for_Autonomous_Driving_Via_Multimodal_Transforme,Deductive_Reinforcement_Learning,Boosting_Offline_Reinforcement_Learning,hanLearningDriveUsing2022,coutoHierarchicalGenerativeAdversarial2024,mohammed_unified_2024,
deng_context_2023,
deng_context-aware_2024,
bai_fen-dqn_2023,jin_vwpefficient_2024,221016567DeFIXDetecting}

%\cite{Think2Drive}, \cite{GRI:General_Reinforced_Imitation}, \cite{End-to-End_Urban_Driving_by_Imitating_a_Reinforcement_Learning_Coach}, \cite{Learning_Urban_Driving_Policies_using_Deep_Reinforcement_Learning}, \cite{Efficient_Learning_of_Urban_Driving}, \cite{Learning_to_drive_from_a_world_on_rails}, \cite{Decision_Making_for_Autonomous_Driving_Via_Multimodal_Transforme}, \cite{Deductive_Reinforcement_Learning}, \cite{Boosting_Offline_Reinforcement_Learning}, \cite{mohammed_unified_2024, deng_context_2023, deng_context-aware_2024, bai_fen-dqn_2023},\cite{hanLearningDriveUsing2022}, \cite{coutoHierarchicalGenerativeAdversarial2024},\cite{jin_vwpefficient_2024}.
\\
\addlinespace

Completion Time & Efficiency & Measures how long it takes the agent to finish a task or reach a goal. &

\cite{Deep_reinforcement_learning_based_control_for_Autonomous_Vehicles_in_CARLA,Deep_Reinforcement_Learning_based_control_algorithms:_Training_and_validation_using_the_ROS_Framework,Vision-Based_Autonomous_Car_Racing_Using_Deep_Imitative_Reinforcement_Learning,Domain_Adaptation_In_Reinforcement_Learning,Deep_Reinforcement_Learning_for_Autonomous_Vehicle_Intersection_Navigation,Safe_Navigation_Based_on_Deep_Q-Network_Algorithm_Using,A_Reinforcement_Learning_Based_Approach_for_Controlling_Autonomous_Vehicles_in_Complex_Scenarios,Hierarchical_Reinforcement_Learning_with_Successor_Representation_for_Intelligent_Vehicle_Collision_Avoidance_of_Dynamic_Pedestrian,wu_lane_2022,
zhang_toward_2022,
cheng_longitudinal_2021,
zhao_end--end_2024,
liao_lateral_2023,
huang_simoun_2023,gupta_hylear_2023,deshpande_navigation_2021,martinez_gomez_temporal_2023,shi_efficient_2023}

%\cite{Deep_reinforcement_learning_based_control_for_Autonomous_Vehicles_in_CARLA}, \cite{Deep_Reinforcement_Learning_based_control_algorithms:_Training_and_validation_using_the_ROS_Framework}, \cite{Vision-Based_Autonomous_Car_Racing_Using_Deep_Imitative_Reinforcement_Learning}, \cite{Deep_Reinforcement_Learning_for_Autonomous_Vehicle_Intersection_Navigation}, \cite{Safe_Navigation_Based_on_Deep_Q-Network_Algorithm_Using}, \cite{A_Reinforcement_Learning_Based_Approach_for_Controlling_Autonomous_Vehicles_in_Complex_Scenarios}, \cite{Hierarchical_Reinforcement_Learning_with_Successor_Representation_for_Intelligent_Vehicle_Collision_Avoidance_of_Dynamic_Pedestrian}, \cite{wu_lane_2022, zhang_toward_2022, cheng_longitudinal_2021, zhao_end--end_2024, liao_lateral_2023, huang_simoun_2023},\cite{gupta_hylear_2023},\cite{deshpande_navigation_2021},\cite{martinez_gomez_temporal_2023},\cite{shi_efficient_2023}.

\\
\addlinespace

Average Speed & Efficiency & Reflects the mean speed of the ego vehicle. Sometimes analyzed with variance to evaluate consistency. & \cite{wang_benchmarking_2021,
khalil_exploiting_2023,
chen_motion_2024,
yang_decision-making_2022,deshpande_navigation_2021,albilani_guided_2023,Vision-Based_Autonomous_Car_Racing_Using_Deep_Imitative_Reinforcement_Learning,Efficient_Learning_of_Urban_Driving,Reinforced_Curriculum_Learning_For_Autonomous_Driving,Deep_Reinforcement_Learning_Control_Strategy_at_Roundabout_for_i-CAR,Addressing_Lane_Keeping_and_Intersections,igoeMultiAlphaSoftActorCritic2023}

%\cite{Vision-Based_Autonomous_Car_Racing_Using_Deep_Imitative_Reinforcement_Learning}, \cite{Reinforced_Curriculum_Learning_For_Autonomous_Driving}, \cite{Deep_Reinforcement_Learning_Control_Strategy_at_Roundabout_for_i-CAR}, \cite{Addressing_Lane_Keeping_and_Intersections}, \cite{wang_benchmarking_2021, khalil_exploiting_2023, chen_motion_2024, yang_decision-making_2022},\cite{igoeMultiAlphaSoftActorCritic2023},\cite{deshpande_navigation_2021},\cite{albilani_guided_2023}. 
\\
\addlinespace

%\textbf{Average Episode Length} & Number of steps or time an episode lasts before task completion or failure. It indicates how well the policy sustains progress. Seen in \cite{Versatile_and_Efficien}, \cite{Deep_Reinforcement_Learning_Control_Strategy_at_Roundabout_for_i-CAR}, \cite{wang_benchmarking_2021,mohammed_unified_2024}. \\

Reward / Return & Overall performance &Average reward obtained per episode during evaluation. It reflects policy performance with respect to the task’s reward function.  & 
\cite{Domain_Adaptation_In_Reinforcement_Learning,Model-Based_Reinforcement_Learning_with_Isolated_Imaginations,Cognitive_Reinforcement_Learning,Reinforced_Curriculum_Learning_For_Autonomous_Driving,Autonomous_Driving_via_Knowledge-Enhanced,A_Decision_Control_Method,Deep_Reinforcement_Learning_for_Autonomous_Vehicle_Intersection_Navigation,A_Hierarchical_Autonomous_Driving_Framework,Enhanced_Off-Policy_Reinforcement_Learning_With_Focused_Experience_Replay,Integration_of_Motion_Prediction_with_End-to-end_Latent_RL_for_Self-Driving_Vehicles,A_Reinforcement_Learning_Based_Approach_for_Controlling_Autonomous_Vehicles_in_Complex_Scenarios,Addressing_Lane_Keeping_and_Intersections,Using_the_CARLA_Simulator_to_Train_A_Deep_Q_Self-Driving,Integrating_Deep_Reinforcement_Learning_with_Model-based_Path_Planners_for_Automated_Driving,Reinforcement_Learning-Based_Guidance_of_Autonomous_Vehicles,Boosting_Offline_Reinforcement_Learning,Towards_Autonomous_Driving,A_Deep_Q-Network_Reinforcement_Learning-Based,gutierrez-morenoDecisionMakingAutonomous2024,hanLearningDriveUsing2022,igoeMultiAlphaSoftActorCritic2023,heAdaptiveDecisionMaking2022,tsaiAutonomousVehicleFollowingTechnique2023,naveedTrajectoryPlanningAutonomous2020,gangopadhyayHierarchicalProgramTriggeredReinforcement2022,frijiDQNBasedAutonomousCarFollowing2020,jaafraContextAwareAutonomousDriving2019,hishmehDeerHeadlightsShort2020,xu_decision-making_2022,
khalil_exploiting_2023,
chen_motion_2024,
deng_context_2023,
zhang_toward_2022,
goel_adaptive_2021,
manikandan_ad_2023,
savari_online_2021,
xu_end--end_2023,
deng_context-aware_2024,
wu_reinforcement_2023,
zhao_end--end_2024,
frauenknecht_data-efficient_2023,
bai_fen-dqn_2023,
wu_proximal_2023,
muhammed_developing_2021,
youssef_deep_2019,
liao_lateral_2023,
huang_simoun_2023,
elallid_dqn-based_2022,hussonnois_end--end_2022,albilani_guided_2023,jin_vwpefficient_2024,li_dynamic_2022,ahmed_policy-based_2022,noauthor_explainable_nodate}

%\cite{Domain_Adaptation_In_Reinforcement_Learning} , \cite{Model-Based_Reinforcement_Learning_with_Isolated_Imaginations}, \cite{Cognitive_Reinforcement_Learning}, \cite{Reinforced_Curriculum_Learning_For_Autonomous_Driving}, \cite{Autonomous_Driving_via_Knowledge-Enhanced}, \cite{A_Decision_Control_Method} , \cite{Deep_Reinforcement_Learning_for_Autonomous_Vehicle_Intersection_Navigation}, \cite{A_Hierarchical_Autonomous_Driving_Framework}, \cite{Enhanced_Off-Policy_Reinforcement_Learning_With_Focused_Experience_Replay} , \cite{Integration_of_Motion_Prediction_with_End-to-end_Latent_RL_for_Self-Driving_Vehicles}, \cite{A_Reinforcement_Learning_Based_Approach_for_Controlling_Autonomous_Vehicles_in_Complex_Scenarios}, \cite{Addressing_Lane_Keeping_and_Intersections}, \cite{Using_the_CARLA_Simulator_to_Train_A_Deep_Q_Self-Driving}, \cite{Integrating_Deep_Reinforcement_Learning_with_Model-based_Path_Planners_for_Automated_Driving}, \cite{Reinforcement_Learning-Based_Guidance_of_Autonomous_Vehicles}, \cite{Boosting_Offline_Reinforcement_Learning} , \cite{Towards_Autonomous_Driving}, \cite{A_Deep_Q-Network_Reinforcement_Learning-Based}, \cite{xu_decision-making_2022, khalil_exploiting_2023, chen_motion_2024, deng_context_2023, zhang_toward_2022, goel_adaptive_2021, manikandan_ad_2023, savari_online_2021, xu_end--end_2023, deng_context-aware_2024, wu_reinforcement_2023, zhao_end--end_2024, frauenknecht_data-efficient_2023, bai_fen-dqn_2023, wu_proximal_2023, muhammed_developing_2021, youssef_deep_2019, liao_lateral_2023, huang_simoun_2023, elallid_dqn-based_2022}. 
\\
\addlinespace

Driving Score & Overall driving quality & A composite metric combining route completion, infraction penalties, and sometimes efficiency. & 
\cite{Think2Drive,GRI:General_Reinforced_Imitation,End-to-End_Urban_Driving_by_Imitating_a_Reinforcement_Learning_Coach,Learning_to_drive_from_a_world_on_rails,Decision_Making_for_Autonomous_Driving_Via_Multimodal_Transforme,Boosting_Offline_Reinforcement_Learning,coutoHierarchicalGenerativeAdversarial2024,221016567DeFIXDetecting,deng_context_2023,
huang_simoun_2023,jin_vwpefficient_2024}
%\cite{Think2Drive}, \cite{GRI:General_Reinforced_Imitation}, \cite{End-to-End_Urban_Driving_by_Imitating_a_Reinforcement_Learning_Coach}, \cite{Learning_to_drive_from_a_world_on_rails}, \cite{Decision_Making_for_Autonomous_Driving_Via_Multimodal_Transforme}, \cite{Boosting_Offline_Reinforcement_Learning}, \cite{deng_context_2023, huang_simoun_2023}. 
\\
\addlinespace

Speed Profile & Efficiency/Comfort  & A speed profile is a time-based or distance-based representation of how a vehicle's speed changes along its route. It gives a detailed picture of the vehicle's acceleration and deceleration during a driving episode.&  \cite{Versatile_and_Efficien,Using_the_CARLA_Simulator_to_Train_A_Deep_Q_Self-Driving,Learning_Automated_Driving_in_Complex_Intersection_Scenarios_Based,An_Integrated_Reward_Function,Hierarchical_Reinforcement_Learning_with_Successor_Representation_for_Intelligent_Vehicle_Collision_Avoidance_of_Dynamic_Pedestrian,igoeMultiAlphaSoftActorCritic2023,cheng_longitudinal_2021,
yang_decision-making_2022,
bai_fen-dqn_2023,
wu_proximal_2023,
muhammed_developing_2021,jin_vwpefficient_2024}  \\
Lane Deviation & Safety/Comfort & Lane deviation measures how far a vehicle drifts or stays away from the center of its lane while driving. It is a critical metric for evaluating how well an AV (or human driver) maintains lane discipline. &  \cite{Safe_Navigation_Based_on_Deep_Q-Network_Algorithm_Using,Addressing_Lane_Keeping_and_Intersections,An_Integrated_Reward_Function,jia_safe_2023,
khalil_exploiting_2023,
wu_lane_2022,
mohammed_unified_2024,
goel_adaptive_2021,
zhao_end--end_2024,
frauenknecht_data-efficient_2023,
wu_proximal_2023,
aghdasian_autonomous_2023,albilani_guided_2023,jin_vwpefficient_2024}
 \\
 \addlinespace
 
Driving Distance & Task-progress/Efficiency & Driving distance refers to the total length of the path that a vehicle actually travels during a driving episode or simulation.  & \cite{Addressing_Lane_Keeping_and_Intersections,Using_the_CARLA_Simulator_to_Train_A_Deep_Q_Self-Driving,A_Deep_Q-Network_Reinforcement_Learning-Based,deshpande_navigation_2021,albilani_guided_2023,jin_vwpefficient_2024}  \\
\addlinespace

Others & N/A & There are some papers that used other metrics including but not limited to \emph{mean-route-error}: Average deviation from planned route (trajectory error), \emph{inference-time}:Time taken by model to make a decision, \emph{heading-deviation}: Difference between vehicle heading and road direction, \emph{no-collision}: Boolean or count of episodes without collision, \emph{average-episode-length}: Average number of steps per episode, \emph{convergence}: Metric to track if training is stabilizing,\emph{lane-invasion-rate}: Frequency of lane marking violations, \emph{training-time}: Total training duration, episode length  & \cite{li_dynamic_2022,Deep_reinforcement_learning_based_control_for_Autonomous_Vehicles_in_CARLA,Deep_Reinforcement_Learning_based_control_algorithms:_Training_and_validation_using_the_ROS_Framework},\cite{gupta_hylear_2023,noauthor_explainable_nodate,Autonomous_Driving_via_Knowledge-Enhanced},  \cite{albilani_guided_2023,A_Decision_Control_Method,An_Integrated_Reward_Function}, \cite{deshpande_navigation_2021,ahmed_policy-based_2022,martinez_gomez_temporal_2023,A_Bayesian_Approach_to_Reinforcement_Learning_of_Vision-Based_Vehicular_Control}, \cite{Versatile_and_Efficien,Deep_Reinforcement_Learning_Control_Strategy_at_Roundabout_for_i-CAR,wang_benchmarking_2021,mohammed_unified_2024}, \cite{jin_vwpefficient_2024,A_Methodology_Based_on_Deep_Reinforcement_Learning,Towards_Autonomous_Driving,gutierrez-morenoDecisionMakingAutonomous2024},  \cite{A_Deep_Q-Network_Reinforcement_Learning-Based,Urban_Autonomous_Driving_of_Emergency_Vehicles_with_Reinforcement_Learning,naveedTrajectoryPlanningAutonomous2020},  \cite{gutierrez-morenoDecisionMakingAutonomous2024,gupta_hylear_2023,hussonnois_end--end_2022},  \cite{zhao_end--end_2024,bai_fen-dqn_2023,muhammed_developing_2021}.
\\

\bottomrule
\end{tabular}
\end{adjustbox}
\end{table*}

While foundational metrics like success rate, collision rate, and reward dominate the literature, many studies introduce specialized metrics tailored to specific needs. For example, \cite{Versatile_and_Efficien} assess lateral and longitudinal smoothness via the standard deviation of steering and speed, respectively. Other works incorporate human-likeness scores \cite{cheng_longitudinal_2021}, command execution accuracy \cite{A_Methodology_Based_on_Deep_Reinforcement_Learning}, or critical driving comfort measures \cite{Learning_Automated_Driving_in_Complex_Intersection_Scenarios_Based}. Additionally, training-focused evaluations such as action-value convergence, episode length trends, and convergence diagnostics are increasingly used to track learning stability and robustness \cite{xu_end--end_2023}.

In total, we identified 95 unique evaluation metrics across the surveyed literature. Although not all are widely adopted, many provide valuable insights for specific deployment scenarios, particularly in safety-critical or human-in-the-loop contexts. To maintain clarity, Table \ref{tab:evaluation_metrics} highlights metrics used in at least five studies (except the others categoty that has the least number of 3).

\section{Towns and Scenario}
In CARLA-based RL research, towns refer to different simulated environments (e.g., Town01 to Town10), each with unique layouts, road structures, and environmental features. Scenarios, on the other hand, define specific driving tasks or challenges within these towns, such as lane following, intersection handling, or collision avoidance. While CARLA includes some predefined maps and features, most scenarios are manually created by researchers to focus on particular behaviors or challenges. Some studies adopt standardized benchmarks like CoRL 2017, NoCrash, or the CARLA Leaderboard, which provide consistent evaluation settings. Others design custom routes and traffic situations to test specific hypotheses, evaluate generalization to unseen roads, or simulate rare edge cases. The choice of town and scenario plays a key role in shaping task difficulty and determining how well a trained policy performs in new or complex environments. 
\begin{table}[t] % single-column float
\centering
\footnotesize
\setlength{\tabcolsep}{2pt}         % tighter columns
\renewcommand{\arraystretch}{0.92}  % tighter rows
\caption{Scenario categories vs. CARLA towns. A checkmark (\cmark) indicates the scenario has been studied in that town. “Custom” = non‑official or heavily modified maps.}
%\resizebox{\columnwidth}{!}{%
\begin{tabular}{lccccccccccc}
\toprule
\textbf{Scenario Type} &
\rotatebox{90}{Town01} & \rotatebox{90}{Town02} & \rotatebox{90}{Town03} &
\rotatebox{90}{Town04} & \rotatebox{90}{Town05} & \rotatebox{90}{Town06} &
\rotatebox{90}{Town07} & \rotatebox{90}{Town08} & \rotatebox{90}{Town09} &
\rotatebox{90}{Town10} & \rotatebox{90}{Custom} \\
\midrule
Urban driving / route completion     & \cmark & \cmark & \cmark & \cmark & \cmark & \cmark & \cmark &  &  & \cmark &  \\
Intersections / T-junctions          & \cmark &        & \cmark &        &        &        &        &  &  &        & \cmark \\
Roundabouts                          &        &        & \cmark &        &        &        &        &  &  &        &        \\
Highway / lane-change                &        &        &        & \cmark & \cmark & \cmark &        &  &  &        & \cmark \\
Parking / partial control            &        &        &        &        & \cmark & \cmark &        &  &  &        &        \\
Racing / high-speed tracks           &        &        &        &        &        &        &        &  &  &        & \cmark \\
Domain adaptation / multi-weather    & \cmark & \cmark & \cmark & \cmark &     \cmark   & \cmark &\cmark        &  &  &  &        \\
Custom envs / special setups         &        &        &        &        &        &        &        &  &  &        & \cmark \\
\bottomrule
\end{tabular}%
%}
\label{tab:town_scenario_matrix}
\end{table}

Table~\ref{tab:town_scenario_matrix} provides a consolidated overview of the scenario types examined in the literature and their corresponding CARLA towns. While CARLA offers ten official towns, the majority of studies focus on a limited subset, particularly Town01 and Town02, which feature relatively simple urban layouts suited for route completion and intersection handling. More complex designs such as roundabouts (Town03) and multi-lane highways (Town04–Town06) are less frequently explored, though they offer richer testing grounds for lane-change and merging maneuvers. Parking and partial-control experiments are concentrated in Town05–Town06 due to the inclusion of parking lots, whereas specialized tasks such as high-speed racing are typically conducted in custom-designed environments outside the default towns. This distribution highlights both the historical availability of certain maps and a tendency in the community to prioritize simpler, more accessible settings over more challenging layouts.

\subsection{Urban Driving in Official CARLA Towns}
A large subset of papers focuses on general urban driving and route-completion tasks using the default CARLA towns (Town01 through Town10) under varying levels of traffic density, weather, and route complexity. Several works concentrate on Town01 and Town02 for basic or intermediate urban tasks, such as lane keeping and navigation with occasional turns. For instance, \cite{Deep_reinforcement_learning_based_control_for_Autonomous_Vehicles_in_CARLA,Deep_Reinforcement_Learning_based_control_algorithms:_Training_and_validation_using_the_ROS_Framework} both use Town01 with predefined routes (sometimes with dynamic waypoints but no obstacles). Similarly, \cite{A_Bayesian_Approach_to_Reinforcement_Learning_of_Vision-Based_Vehicular_Control,A_Hierarchical_Autonomous_Driving_Framework,Addressing_Lane_Keeping_and_Intersections} employ Town01 and Town02 to study how RL agents adapt from simpler to more complex road structures, including traffic lights and pedestrian traffic. Other works (\cite{Learning_Urban_Driving_Policies_using_Deep_Reinforcement_Learning,Deductive_Reinforcement_Learning,A_Bayesian_Approach_to_Reinforcement_Learning_of_Vision-Based_Vehicular_Control,Boosting_Offline_Reinforcement_Learning,youssef_deep_2019}) similarly center on Town01/Town02 benchmark (e.g., CoRL2017 or NoCrash benchmarks).

Town03 is a frequent choice for more complex urban settings, with denser traffic or trickier intersections. Papers such as \cite{End-to-End_Urban_Driving_by_Imitating_a_Reinforcement_Learning_Coach,Autonomous_Driving_via_Knowledge-Enhanced,Decision_Making_for_Autonomous_Driving_Via_Multimodal_Transforme,Integration_of_Motion_Prediction_with_End-to-end_Latent_RL_for_Self-Driving_Vehicles} train or evaluate on Town03 with heavy traffic (up to 100 vehicles or more), often testing advanced methods like motion-prediction-based RL. Similarly, \cite{Towards_Autonomous_Driving,A_Deep_Q-Network_Reinforcement_Learning-Based,A_Reinforcement_Learning_Based_Approach_for_Controlling_Autonomous_Vehicles_in_Complex_Scenarios} also adopt Town03 for navigating crossroads, roundabouts, and congested roads. Some authors (\cite{Think2Drive,GRI:General_Reinforced_Imitation,Versatile_and_Efficien}) use multiple towns or LeaderBoard routes encompassing Town03 to rigorously validate their policies.

Meanwhile, Town04 appears in works such as \cite{A_Decision_Control_Method,A_Methodology_Based_on_Deep_Reinforcement_Learning,huang_simoun_2023,jia_safe_2023,khalil_exploiting_2023}, often featuring highways, urban blocks, or both. Several papers (\cite{wei_continual_2023,deng_context_2023,deng_context-aware_2024,carton_using_2021}) also highlight Town04 in meta-learning or continual-learning contexts, sampling diverse traffic and weather conditions. Town05 is popular for route-following or intermediate urban tasks (\cite{End-to-End_Model-Free_Reinforcement_Learning_for_Urban_Driving,Integrating_Deep_Reinforcement_Learning_with_Model-based_Path_Planners_for_Automated_Driving,Reinforcement_Learning-Based_Guidance_of_Autonomous_Vehicles,wu_reinforcement_2023}), and Town06 and Town07 often appear in more specialized scenarios (e.g., \cite{End-to-End_Urban_Driving_by_Imitating_a_Reinforcement_Learning_Coach,Learning_to_drive_from_a_world_on_rails,Reinforced_Curriculum_Learning_For_Autonomous_Driving,Domain_Adaptation_In_Reinforcement_Learning,zhao_end--end_2024,cheng_longitudinal_2021}). Town10 appears as well, such as in \cite{Using_the_CARLA_Simulator_to_Train_A_Deep_Q_Self-Driving}, which tests dynamic weather conditions and city-style roads.

Within these general urban-driving papers, some focus on dense traffic or advanced collision-avoidance behaviors, like \cite{Efficient_Learning_of_Urban_Driving} (Town02 with 30 vehicles and 50 pedestrians), \cite{Urban_Autonomous_Driving_of_Emergency_Vehicles_with_Reinforcement_Learning} (emergency vehicles in heavy traffic), and \cite{Deep-Reinforcement-Learning-Based_Driving_Policy_at_Intersections} (scaled-up intersections). Others emphasize particular training regimens, such as curriculum learning (\cite{Reinforced_Curriculum_Learning_For_Autonomous_Driving}), offline RL (\cite{Boosting_Offline_Reinforcement_Learning}), or an imitation+RL hybrid (\cite{End-to-End_Urban_Driving_by_Imitating_a_Reinforcement_Learning_Coach}, \cite{peng_imitative_2020}). In all, these works collectively address the challenge of teaching an autonomous agent to navigate city roads with traffic lights, pedestrians, and dynamic vehicle populations.

\subsection{Intersection-Focused and T-Junction Scenarios}
A separate set of studies zooms in on intersection handling, particularly T-junctions or unsignalized intersections. For instance, \cite{Deep_Reinforcement_Learning_for_Autonomous_Vehicle_Intersection_Navigation} evaluates navigation in a T-intersection with progressively increasing pedestrian/vehicle density (up to 450 vehicles/pedestrians). Town03 also appears frequently for unprotected or unsignalized intersections (\cite{xu_decision-making_2022}), testing the agent’s ability to yield and merge properly. In \cite{Deep-Reinforcement-Learning-Based_Driving_Policy_at_Intersections}, training focuses on a single left-turn intersection, with testing on four additional intersections of increasing complexity, while \cite{Learning_Automated_Driving_in_Complex_Intersection_Scenarios_Based} designs “complex left-turn across path” collision-prone maneuvers. Similarly, \cite{bai_fen-dqn_2023} addresses a simpler single intersection in Town01, emphasizing red-light compliance and safe car-following behaviors. Other works such as \cite{A_Reinforcement_Learning_Based_Approach_for_Controlling_Autonomous_Vehicles_in_Complex_Scenarios} also highlight intersection-dense environments, scaling traffic to hundreds of vehicles and pedestrians. These intersection-focused studies often delve into safety-critical decision-making—when to brake, how to sequence turns, and how to handle partial occlusions (e.g. \cite{chen_motion_2024} with pedestrians in the blind-spot.)

\subsection{Roundabouts and Specialized Urban Maneuvers}
Several papers concentrate on roundabouts, an environment that combines intersection-like decision-making with circular traffic flow. In Town03, \cite{Deep_Reinforcement_Learning_Control_Strategy_at_Roundabout_for_i-CAR} contrasts a four-way roundabout (with intersections) to a simpler U-turn loop, while \cite{End-to-End_Urban_Driving_by_Imitating_a_Reinforcement_Learning_Coach} and \cite{Learning_to_drive_from_a_world_on_rails} also include roundabouts in their urban routes (alongside highways and city blocks). Roundabouts force the RL agent to negotiate merges, exits, and near-constant lateral control challenges.

Additionally, some works target narrower urban driving maneuvers beyond intersections or roundabouts. For example, \cite{Vision-Based_Autonomous_Car_Racing_Using_Deep_Imitative_Reinforcement_Learning} addresses a specialized high-speed track within an urban-ish environment, albeit more akin to a racing scenario. \cite{goel_adaptive_2021} and \cite{manikandan_ad_2023} highlight speed-change maneuvers in an urban setting, examining how RL policies handle varying velocity constraints (e.g., from 8 m/s to 16 m/s). Meanwhile, studies like \cite{chronis_driving_2021} define multiple routes in Town03, some with sharp 45–180° turns, to stress-test an end-to-end policy’s turning ability in dense urban grids.

\subsection{Highway and Lane-Change Scenarios}
Another broad category focuses on highway driving or lane-change maneuvers in multi-lane roads. Certain papers combine official towns with highway segments—for example, \cite{End-to-End_Model-Free_Reinforcement_Learning_for_Urban_Driving} trains partly in Town05 and tests on a highway environment to see if the RL agent generalizes from urban to highway. In \cite{Model-Based_Reinforcement_Learning_with_Isolated_Imaginations} sets up a highway scenario in Town04 with 30 vehicles in day/night conditions, while \cite{Cognitive_Reinforcement_Learning} uses a dual-lane highway with fluent vs. congested traffic. \cite{yang_decision-making_2022} presents a five-lane road in Town06, replicating ramp entries and variable traffic speeds.

Lane-changing is explicitly studied in \cite{wang_benchmarking_2021} (a custom three-lane highway), \cite{wu_lane_2022} (a 2-lane straight road with discretionary lane changes), and \cite{liao_lateral_2023} (two-lane roads with static/dynamic obstacles). These works typically measure collisions, comfortable lateral maneuvers, and safe merges, often restricting the environment to minimal intersections or signals. Such highway-focused research underscores how RL must handle higher speeds and more frequent lane merges than in typical urban blocks.

\subsection{Parking and Partial-Task Scenarios}
A smaller cluster of studies focuses on parking maneuvers or partial-control tasks (e.g., only longitudinal control). For instance, \cite{wu_reinforcement_2023} tackles vertical and parallel parking in Town05, showcasing how RL can learn precise maneuvering into parking bays. 

Considering the partial task scenarios \cite{cheng_longitudinal_2021} in Town06 considers only throttle and brake with a single lead vehicle, while steering remains static—this isolates speed-control strategies for collision avoidance. Such partial or specialized tasks often serve to simplify the action space so the RL agent can master one dimension of driving (speed or lateral alignment) before moving to more complex scenarios.

\subsection{Racing or High-Speed Specialized Tracks}
A distinct set of references moves beyond ordinary urban or highway driving into racing or closed-course scenarios. \cite{Vision-Based_Autonomous_Car_Racing_Using_Deep_Imitative_Reinforcement_Learning} implements a 1 km custom racetrack within CARLA, containing tight turns and obstacles in “easy” or “hard” variations. Some works (\cite{frauenknecht_data-efficient_2023,muhammed_developing_2021}) also mention custom closed-circuit setups, though they may not always be labeled “racing.” These tasks emphasize high-speed cornering, minimal margin for lane deviation, and sometimes transfer to real-world scaled vehicles. By contrast, typical urban RL tasks rarely push the car to racing speeds or require strictly apexing corners.

\subsection{Domain Adaptation and Multi-Weather Training}
Many papers investigate weather variations and domain adaptation—teaching agents to generalize from sunny noon to rain, fog, dusk, or even new towns. Works like \cite{Reinforced_Curriculum_Learning_For_Autonomous_Driving} explicitly train in “soft” weather presets (e.g clear, cloudy noon, etc) and test in “hard” conditions (e.g heavy rain, sunset, etc). \cite{GRI:General_Reinforced_Imitation,peng_imitative_2020,Boosting_Offline_Reinforcement_Learning,khalil_exploiting_2023} also vary meteorological settings (e.g., HardRainNoon, WetNoon, ClearSunset) to measure how robustly a learned policy transfers. Meanwhile, \cite{Domain_Adaptation_In_Reinforcement_Learning} trains the model in Town07 on late evening and tests in clear noon or hard rain, illustrating domain shift in both lighting and precipitation.

Separately, meta-learning or continual learning approaches (\cite{deng_context_2023,deng_context-aware_2024,carton_using_2021,wei_continual_2023}) treat each new weather, traffic pattern, or town layout as a “task.” The RL agent samples from multiple tasks at training time, improving its capacity to adapt quickly to unseen conditions (e.g., new towns or more severe weather). Some studies \cite{deng_context_2023, deng_context-aware_2024} incorporate Town01–Town04 as meta-training environments, then measure performance on Town05–Town06.

\subsection{Custom Environments and Specialized Configurations}
Finally, a notable subset of studies forgo official CARLA towns entirely or modify them heavily. They create custom worlds for tighter experimental control, specialized tasks, or real-to-sim synergy. For example, \cite{wang_benchmarking_2021} builds a three-lane highway map, \cite{liao_lateral_2023} a two-lane 420 m road, and \cite{chen_motion_2024} a scenario with an occluding parked van to test pedestrian-blind-spot crossing. \cite{zhang_toward_2022} deliberately uses a minimal “Town” to isolate reward function variations, while \cite{savari_online_2021} sets up a simpler lane-keeping environment for online RL. Some papers (\cite{zhu_rita_2023}, \cite{wu_proximal_2023}) mention using CARLA generally but do not specify which map or scenario, often focusing instead on algorithmic innovations (e.g., new RL frameworks or 3D perception modules).

Racing or track-based custom worlds appear in \cite{Vision-Based_Autonomous_Car_Racing_Using_Deep_Imitative_Reinforcement_Learning} and \cite{frauenknecht_data-efficient_2023}, the latter modeling a real test track for industrial validation. Others, like \cite{aghdasian_autonomous_2023} and \cite{muhammed_developing_2021}, mention creating diverse random routes or specialized setups, sometimes spanning multiple towns or custom spawns, to avoid overfitting. These custom environments underscore the flexibility of CARLA for both academic research and practical testing scenarios beyond the standard Town0X maps.

\section{Limitation}
Despite recent progress, RL-based autonomous driving in CARLA continues to face several persistent challenges that hinder scalability, safety, and real-world deployment. These limitations span fundamental issues such as sample efficiency, generalization, safety guarantees, perception constraints, scenario realism, and long-horizon planning, among others. Table~\ref{tab:limitations} provides a structured summary of the major limitation categories, while the following subsections discuss each in detail, supported by representative studies from the literature.
\begin{table}[!t]
\centering
%\small
\footnotesize   
\caption{Summary of Limitations in RL-based Autonomous Driving with CARLA}
\setlength\tabcolsep{1pt}
%\begin{tabularx}{\textwidth}{@{}p{3.5cm} X p{3.5cm}@{}}
\begin{tabularx}{\textwidth}{@{}p{2.5cm} X p{2cm}@{}}
\toprule
\textbf{Limitation Category} & \textbf{Key Issues and Challenges} & \textbf{Representative References} \\
\midrule

Sample Efficiency and Data Cost & 
Low sample efficiency; reliance on large datasets and thousands of episodes. Expert demonstrations assumed optimal despite noise (e.g., $\sim$10\% poor actions in autopilot logs). Sparse rewards and reliance on finite data hinder learning. & 
\cite{GRI:General_Reinforced_Imitation, huang_simoun_2023, xu_end--end_2023, deng_context_2023} \\[3pt]

Generalization / Sim-to-Real Gap & 
Strong performance in-simulation but poor transfer due to visual, dynamic, and traffic realism gaps. Policies trained in a single town/weather regime fail under new conditions. Limited sim-to-real evaluations. Safety-aware RL still exhibits systematic errors (e.g., wide turns). Scenario-specific training hampers transferability. & 
\cite{End-to-End_Urban_Driving_by_Imitating_a_Reinforcement_Learning_Coach, Deep-Reinforcement-Learning-Based_Driving_Policy_at_Intersections, coutoHierarchicalGenerativeAdversarial2024, zhao_real-time_2021, youssef_deep_2019, cheng_longitudinal_2021, savari_online_2021, carton_using_2021, wang_benchmarking_2021, albilani_guided_2023, naveedTrajectoryPlanningAutonomous2020} \\[3pt]

Safety Guarantees and Risk-awareness & 
Lack of explicit safe-RL mechanisms; unstable across traffic situations without fine-tuned hyperparameters. Safety often delegated to external shields. Simulations omit actuation errors and rare corner cases. Constrained RL improves safety but reduces comfort. Safety–utility trade-off remains underexplored. & 
\cite{Decision_Making_for_Autonomous_Driving_Via_Multimodal_Transforme, Urban_Autonomous_Driving_of_Emergency_Vehicles_with_Reinforcement_Learning, gangopadhyayHierarchicalProgramTriggeredReinforcement2022, yang_decision-making_2022, deng_context-aware_2024, wei_continual_2023} \\[3pt]

Sparse / Poorly Shaped Rewards & 
Rewards often sparse, inconsistent, or overly coarse. Aggregating multiple objectives into single scalars hides trade-offs. Few studies analyze macroscopic traffic efficiency. Oversimplified rewards can trap agents in local optima. Lack of systematic reward design analysis. & 
\cite{GRI:General_Reinforced_Imitation, Efficient_Learning_of_Urban_Driving, A_Hierarchical_Autonomous_Driving_Framework, jaafraContextAwareAutonomousDriving2019, zhang_toward_2022, wu_lane_2022} \\[3pt]

Perception Limits and Sensor Fusion & 
Over-reliance on CARLA’s waypoint oracle or RGB-only inputs. Limited use of LiDAR, RADAR, GPS, or inertial data. Simulation setups often differ from real sensor suites (e.g., lack of rear cameras, ignored depth/LiDAR channels). Poor robustness to adverse weather or unseen towns. & 
\cite{Deep_reinforcement_learning_based_control_for_Autonomous_Vehicles_in_CARLA, Vision-Based_Autonomous_Car_Racing_Using_Deep_Imitative_Reinforcement_Learning, Autonomous_Driving_via_Knowledge-Enhanced, Deep_Reinforcement_Learning_for_Autonomous_Vehicle_Intersection_Navigation, A_Methodology_Based_on_Deep_Reinforcement_Learning, Deep_Reinforcement_Learning_Control_Strategy_at_Roundabout_for_i-CAR, Towards_Autonomous_Driving, gutierrez-morenoDecisionMakingAutonomous2024, jin_vwpefficient_2024, frijiDQNBasedAutonomousCarFollowing2020, manikandan_ad_2023, wang_vision-based_2023, mohammed_unified_2024} \\[3pt]

Scenario Realism and Multi-Agent Interaction & 
Simplified scenarios: fixed throttle, static obstacles, discretized action spaces, narrow geographic coverage. Missing pedestrians, bicycles, or complex multi-agent interactions. Evaluations often under fixed weather and limited maps. Robustness to unseen layouts rarely tested. & 
\cite{A_Methodology_Based_on_Deep_Reinforcement_Learning, Addressing_Lane_Keeping_and_Intersections, Integrating_Deep_Reinforcement_Learning_with_Model-based_Path_Planners_for_Automated_Driving, Learning_Automated_Driving_in_Complex_Intersection_Scenarios_Based, A_Deep_Q-Network_Reinforcement_Learning-Based, 221016567DeFIXDetecting, chronis_driving_2021, goel_adaptive_2021, chen_motion_2024, xu_decision-making_2022, jia_safe_2023, udatha_reinforcement_2023, Integration_of_Motion_Prediction_with_End-to-end_Latent_RL_for_Self-Driving_Vehicles, Towards_Autonomous_Driving} \\[3pt]

Hierarchical / Long-Horizon Planning & 
Current agents are short-horizon and reactive, lacking explicit route planning or long-term reasoning. Few integrate trajectory prediction or multi-path planning. Limited foresight prevents goal-directed, mission-level control. & 
\cite{Efficient_Learning_of_Urban_Driving, A_Methodology_Based_on_Deep_Reinforcement_Learning, Deep-Reinforcement-Learning-Based_Driving_Policy_at_Intersections, hishmehDeerHeadlightsShort2020, gupta_hylear_2023} \\[3pt]

Hyper-Parameter Sensitivity and Training Stability & 
Performance highly sensitive to hyper-parameters, batch sizes, and replay strategies. On-policy methods (e.g., PPO) data-hungry; off-policy methods fragile without tuning. Limited GPU budgets and short training times worsen instability. CNN backbones not optimized for urban scenes. & 
\cite{Learning_Urban_Driving_Policies_using_Deep_Reinforcement_Learning, Deductive_Reinforcement_Learning, Enhanced_Off-Policy_Reinforcement_Learning_With_Focused_Experience_Replay, khalil_exploiting_2023, Deep_Reinforcement_Learning_Control_Strategy_at_Roundabout_for_i-CAR, jaafraContextAwareAutonomousDriving2019, aghdasian_autonomous_2023, End-to-End_Urban_Driving_by_Imitating_a_Reinforcement_Learning_Coach} \\[3pt]

Insufficient Real-World Validation & 
Overwhelming reliance on simulation; most methods defer real-world testing. Model-free RL impractical due to data demands. Prototype systems lack real sensor data. Controllers trained in simulation fail in real-world layouts. Real-world scalability and robustness remain unverified. & 
\cite{Deep_Reinforcement_Learning_based_control_algorithms:_Training_and_validation_using_the_ROS_Framework, End-to-End_Model-Free_Reinforcement_Learning_for_Urban_Driving, Deductive_Reinforcement_Learning, Autonomous_Driving_via_Knowledge-Enhanced, gutierrez-morenoDecisionMakingAutonomous2024, Learning_Urban_Driving_Policies_using_Deep_Reinforcement_Learning, Integrating_Deep_Reinforcement_Learning_with_Model-based_Path_Planners_for_Automated_Driving, Urban_Autonomous_Driving_of_Emergency_Vehicles_with_Reinforcement_Learning, Using_the_CARLA_Simulator_to_Train_A_Deep_Q_Self-Driving, frauenknecht_data-efficient_2023, zhao_end--end_2024, wu_reinforcement_2023, liao_lateral_2023, noauthor_explainable_nodate} \\[3pt]

Traffic-Rule and Social-Norm Compliance & 
Few frameworks enforce explicit traffic rules or social norms. Agents may generate illegal manoeuvres or ignore communicative cues (e.g., signals, brake lights). Road user models remain homogeneous, neglecting cultural/social diversity. & 
\cite{A_Decision_Control_Method, deshpande_navigation_2021, jin_vwpefficient_2024,Integration_of_Motion_Prediction_with_End-to-end_Latent_RL_for_Self-Driving_Vehicles, li_decision-making_2024} \\[3pt]

Performance Gap & 
Despite progress, RL policies underperform expert humans and SOTA baselines. Poor handling of dense traffic and non-standard intersections. Weak lateral control on curves. Policies lack rationality, interpretability, and trustworthiness. & 
\cite{Vision-Based_Autonomous_Car_Racing_Using_Deep_Imitative_Reinforcement_Learning, Reinforced_Curriculum_Learning_For_Autonomous_Driving, Deep-Reinforcement-Learning-Based_Driving_Policy_at_Intersections, li_dynamic_2022, Decision_Making_for_Autonomous_Driving_Via_Multimodal_Transforme} \\[3pt]

Limited Behavior Diversity & 
Agents exhibit narrow manoeuvre sets (lane-keeping, intersections, traffic lights). Rarely handle lane changes, roundabouts, or cooperative merges. Heavy reliance on scripted autopilot and homogeneous actors. Limited scenario templates and lack of heterogeneous traffic agents. & 
\cite{Learning_Urban_Driving_Policies_using_Deep_Reinforcement_Learning, gutierrez-morenoDecisionMakingAutonomous2024, Cognitive_Reinforcement_Learning, Reinforced_Curriculum_Learning_For_Autonomous_Driving, A_Decision_Control_Method, Deep_Reinforcement_Learning_for_Autonomous_Vehicle_Intersection_Navigation, A_Deep_Q-Network_Reinforcement_Learning-Based, frijiDQNBasedAutonomousCarFollowing2020, zhao_real-time_2021, martinez_gomez_temporal_2023, tsaiAutonomousVehicleFollowingTechnique2023, Integration_of_Motion_Prediction_with_End-to-end_Latent_RL_for_Self-Driving_Vehicles, Using_the_CARLA_Simulator_to_Train_A_Deep_Q_Self-Driving, elallid_dqn-based_2022, peng_imitative_2020, wu_proximal_2023, ahmed_policy-based_2022, Addressing_Lane_Keeping_and_Intersections} \\

\bottomrule
\end{tabularx}
\label{tab:limitations}
\end{table}

\subsection{Sample Efficiency and Data Cost}

Many recent autonomous-driving frameworks exhibit limited sample efficiency and impose high data costs. A common assumption that every expert demonstration is perfectly optimal leads to assigning a uniform, high reward; in datasets such as the CARLA autopilot logs, approximately 10\% of the recorded actions are actually poor, injecting reward noise that hinders policy learning \cite{GRI:General_Reinforced_Imitation}. Models with dual-path architectures that improve observation interpretability, such as \textsc{Simoun}, still learn slowly because they must rely on finite datasets and sparse reward signals \cite{huang_simoun_2023}. Other end-to-end approaches likewise struggle with low sample efficiency, demanding thousands of interaction episodes before achieving acceptable performance \cite{xu_end--end_2023}. Overall, inadequate sample efficiency and weak generalization constrain RL agents in complex urban traffic, where rare corner cases are difficult to encounter within limited training tasks; meta-RL augmented with safety constraints or adversarial learning has been proposed as one route to more efficient and safer driving, but its benefits remain to be fully validated \cite{deng_context_2023}.

\subsection{Generalization / Transfer and Sim-to-Real Gap}

Despite impressive in-simulation results, contemporary autonomous-driving agents continue to face substantial challenges in generalisation, transfer, and sim-to-real deployment. \textsc{Roach} \cite{End-to-End_Urban_Driving_by_Imitating_a_Reinforcement_Learning_Coach}, for example, performs strongly in CARLA yet struggles to transfer because of visual, dynamical, and interaction mismatches, and because simulators still model surrounding traffic with limited realism . A pronounced distribution gap between the environments encountered during training and those met during testing further degrades policy reliability \cite{Deep-Reinforcement-Learning-Based_Driving_Policy_at_Intersections}. Policies trained in a single town or under a single weather regime often fail when exposed to new cities or conditions, and many studies omit any true sim-to-real evaluation \cite{coutoHierarchicalGenerativeAdversarial2024}. End-to-end conditional imitation learning, which maps images directly to control commands, finds it difficult to extract the correlations needed for robust decisions and cannot guarantee real-time performance in dense urban traffic \cite{zhao_real-time_2021}. Even safety-aware deep RL variants still display systematic errors—such as consistently wide turns—and, given the safety stakes, remain unsuitable for real-world deployment until success rates approach 100\% \cite{youssef_deep_2019}. Network-based longitudinal controllers are expected to benefit from additional human demonstrations, suggesting that richer data could improve real-world adaptability \cite{cheng_longitudinal_2021}. Current online safety-learning methods are unable to compute next-state estimates for dangerous actions in CARLA and must reuse the current state; future work proposes synthesising those transitions with GANs to assess generalisability \cite{savari_online_2021}. Techniques that rely on segmentation to improve generalisation separate perception from control and may underuse shared representations; forthcoming work will explore alternative architectures and stronger data augmentation \cite{carton_using_2021}. Existing benchmark suites lack low-speed traffic and therefore fail to expose agents to an important subset of urban scenarios; planned extensions will broaden test cases and include model-based RL baselines \cite{wang_benchmarking_2021}. Finally, guided RL benefits from explicit rules, yet too many handcrafted constraints risk overfitting and impairing generalisation; evaluations on the more demanding CARLA Leaderboard are intended to clarify this trade-off \cite{albilani_guided_2023}. Scenario-specific policy training further hampers transfer. Manoeuvres learned for a given junction or speed profile rarely integrate into a unified behavioural graph, so skills do not carry over to unseen traffic compositions or road morphologies \cite{naveedTrajectoryPlanningAutonomous2020}

\subsection{Safety guarantees and Risk-awareness}

Many recent reinforcement-learning approaches for autonomous driving provide only limited assurances of safe operation. Most do not incorporate explicit safe-RL mechanisms, so there is no formal guarantee that the learned policy will avoid hazardous manoeuvres once deployed \cite{Decision_Making_for_Autonomous_Driving_Via_Multimodal_Transforme}. Empirical studies confirm that agents can remain unstable across a range of traffic situations unless their reward structure or hyper-parameters are finely tuned \cite{Urban_Autonomous_Driving_of_Emergency_Vehicles_with_Reinforcement_Learning}. Some frameworks delegate responsibility for collision avoidance to an external symbolic “safety shield,” effectively decoupling safety from learning \cite{gangopadhyayHierarchicalProgramTriggeredReinforcement2022}. Simulation evaluations tend to be overly idealised: they neglect actuation errors and therefore overestimate real-world safety margins, while coarse decision- and control-update intervals further mask timing-critical failures \cite{yang_decision-making_2022}. Because RL agents encounter only a fraction of long-tail corner cases during finite training, unsafe behaviour can still emerge under rare conditions; proposed remedies include embedding safety constraints directly into the objective or adding adversarial curricula to expose risky scenarios earlier \cite{deng_context-aware_2024}. Constrained variants such as EM–SAC demonstrate that stricter limits on deceleration reduce risk but at the cost of ride comfort, illustrating a broader trade-off between safety and utility that remains insufficiently quantified \cite{wei_continual_2023}. Overall, providing verifiable safety guarantees and robust risk awareness is an open challenge that future work must address before RL-based policies can be trusted in safety-critical autonomous-driving applications.

% \cite{muhammed_developing_2021} There is still more research needed to qualify for safety and security. Our future work includes extending this framework to simulate adversarial attacks and analyze the response of different defense mechanisms as part of the AI agent model. \textcolor{blue}{this part seems to be disconnected from the last part no?}

\subsection{Sparse / Poorly Shaped Rewards}

A pervasive limitation of current autonomous-driving research is the continued reliance on sparse or poorly shaped reward signals. In imitation-reinforcement hybrids, identical low-level manoeuvres receive a high reward when executed by the demonstrator but a low reward when generated during exploration, creating an internal inconsistency that can inflate the estimated value of demonstration actions and destabilise learning \cite{GRI:General_Reinforced_Imitation}. More broadly, existing studies seldom analyse how reward design influences macroscopic traffic efficiency \cite{Efficient_Learning_of_Urban_Driving}. Frameworks that embed hierarchical decision layers often retain rudimentary reward definitions whose coarse granularity restricts both flexibility and sample efficiency \cite{A_Hierarchical_Autonomous_Driving_Framework}. Even when multiple objectives are acknowledged, they are typically aggregated into a single scalar, obscuring trade-offs that—in real deployments—must be handled separately; a truly multi-dimensional reward formulation remains to be implemented \cite{jaafraContextAwareAutonomousDriving2019}. Experimental evaluations are likewise confined to a small set of handcrafted reward variants, limiting generality \cite{zhang_toward_2022}. Finally, oversimplified rewards can trap agents in local optima \cite{wu_lane_2022}. Collectively, these shortcomings highlight the need for better-shaped, task-decomposed, and safety-aware reward functions, as well as for systematic studies of how reward design affects both microscopic behaviour and network-level traffic outcomes.

\subsection{Perception limits and sensor fusion}

Most contemporary RL pipelines for autonomous driving continue to operate with impoverished perceptual inputs and minimal sensor fusion, undermining their robustness and real-world fidelity. Numerous studies build directly on CARLA’s internal waypoint oracle, thereby bypassing the need to learn spatial reasoning and limiting adaptability once high-level guidance is absent; future deployments must replace this crutch with camera- and LiDAR-driven perception that matches a production vehicle’s sensor stack \cite{Deep_reinforcement_learning_based_control_for_Autonomous_Vehicles_in_CARLA}. Vision modules themselves are frequently restricted to raw RGB frames, which are sample-inefficient and brittle; semantic renderings would provide more task-relevant abstractions and yield safer policies \cite{Vision-Based_Autonomous_Car_Racing_Using_Deep_Imitative_Reinforcement_Learning}. Several frameworks omit complementary modalities such as LiDAR, RADAR, GPS, or inertial data, despite their proven value for localisation and obstacle detection in adverse conditions \cite{Autonomous_Driving_via_Knowledge-Enhanced,Deep_Reinforcement_Learning_for_Autonomous_Vehicle_Intersection_Navigation,A_Methodology_Based_on_Deep_Reinforcement_Learning,Deep_Reinforcement_Learning_Control_Strategy_at_Roundabout_for_i-CAR,Towards_Autonomous_Driving,gutierrez-morenoDecisionMakingAutonomous2024,jin_vwpefficient_2024}. Even when depth is available, other essential channels—for instance LiDAR in RGB-D setups—are ignored \cite{frijiDQNBasedAutonomousCarFollowing2020}, and real-vehicle prototypes sometimes lack rear-facing cameras or exhibit mechanical steering dead-zones that constrain manoeuvrability \cite{manikandan_ad_2023}. Limited perception using semantic segmentation exposes models to poorly handled weather shifts, and degraded success rates in unseen towns and rain scenarios \cite{wang_vision-based_2023}. The over-idealised sensor suites used in simulation depart markedly from on-road hardware, as acknowledged by recent work that calls for richer multi-modal inputs, promoting more modular, encoder-based vision backbones \cite{mohammed_unified_2024}. Until such multi-sensor fusion and noise-aware perception become standard, the generality and safety of RL-based driving policies will remain fundamentally constrained.

\subsection{Scenario Realism and Multi-Agent Interaction}

Most RL evaluations for autonomous driving still rely on highly idealised scenarios that diverge from real-world traffic. Some studies fix longitudinal control to a constant throttle, preventing the agent from modulating acceleration in response to context \cite{A_Methodology_Based_on_Deep_Reinforcement_Learning}. Collision handling is likewise simplified: agents are tuned to skirt static props such as poles or fences but receive no training on moving vehicles or pedestrians, leaving dynamic-obstacle avoidance essentially untested \cite{Addressing_Lane_Keeping_and_Intersections}. At the control layer, many algorithms discretise steering and acceleration commands; coarse, fixed action sets reduce manoeuvre smoothness and precision which is not the case with real world driving which relies on continuous action space \cite{Integrating_Deep_Reinforcement_Learning_with_Model-based_Path_Planners_for_Automated_Driving,Learning_Automated_Driving_in_Complex_Intersection_Scenarios_Based,A_Deep_Q-Network_Reinforcement_Learning-Based}. Geographic coverage is narrow as well: the DeFIX policy, for example, is validated only on CARLA’s Town05, so its performance in unseen maps remains unknown \cite{221016567DeFIXDetecting}.
Entire pipelines are built and tested in virtual worlds devoid of live pedestrians or cross-traffic, with future work merely proposing to add additional cameras or more demanding scenes \cite{chronis_driving_2021}. Controllers evaluated on empty roads achieve high success rates that quickly erode once dynamic obstacles or poor visibility are introduced, motivating plans to pair the controller with an external motion planner for ramp merges, overtakes, or low-visibility driving \cite{goel_adaptive_2021}. Other studies validate their method on only pedestrian-interaction scenarios; it should further be evaluated on more complex situations  \cite{chen_motion_2024}. Most decision-making modules ignore bicycles, scooters, or pedestrians altogether—an omission acknowledged as a major avenue for future research on multi-agent intersections \cite{xu_decision-making_2022}. Likewise, frameworks tuned for low-speed dual-lane scenarios have yet to prove transferable to broader road networks or on-road trials \cite{jia_safe_2023}. Work on highway merging is typically limited to a simpler lane merging scenarios; extending these models to multiple interacting agents and standardizing benchmark datasets is recognised as essential for fair comparison and uncertainty modelling \cite{udatha_reinforcement_2023}. In some studies experimental protocols themselves reveal further weaknesses: mostly evaluations assume fixed weather conditions, leaving algorithmic resilience to rain, fog or low-illumination settings unverified \cite{Integration_of_Motion_Prediction_with_End-to-end_Latent_RL_for_Self-Driving_Vehicles}. Similarly, although strong results are often reported on a restricted subset of CARLA scenarios, robustness to unseen layouts is rarely demonstrated \cite{Towards_Autonomous_Driving}. 

Together, these observations reveal a pronounced gap between the tidy, single-agent, discretised scenarios prevalent in current research and the dense, multi-actor, continuous-control conditions faced on public roads; closing this gap will require richer multi-agent benchmarks, continuous-action policies, and systematic evaluations that span diverse maps, traffic mixes, and environmental conditions. 

\subsection{Hierarchical / Long-Horizon Planning}

Current RL agents for autonomous driving are still limited to short-horizon, reactive control and offer only rudimentary support for hierarchical or long-range planning. Some studies acknowledge that incorporating trajectory-prediction modules could substantially improve foresight but leave such components to future work \cite{Efficient_Learning_of_Urban_Driving}. Many controllers can avoid collisions and lane invasions yet cannot steer toward an explicit destination, because they operate without a route planner or goal-directed policy layer \cite{A_Methodology_Based_on_Deep_Reinforcement_Learning}. Even when the state representation encodes local road geometry, it omits navigation cues for instance intended future trajectories, preventing the policy from reasoning beyond the immediate scene \cite{Deep-Reinforcement-Learning-Based_Driving_Policy_at_Intersections}. Some systems can perform only simple, short-term manoeuvre selection and cannot sequence actions over extended horizons \cite{hishmehDeerHeadlightsShort2020}. Some frameworks like HyLEAR \cite{gupta_hylear_2023} favours comfortable but often longer routes, and ongoing work is adding multi-path planning, prediction of neighbouring vehicles, and noise-aware perception precisely to extend its look-ahead capability. Overall, the absence of integrated, long-horizon planning modules remains a significant barrier to deploying RL-based driving stacks in real-world, goal-oriented missions.

\subsection{Hyper-Parameter Sensitivity and Training Stability}
A persistent obstacle to reliable RL based autonomous-driving is the pronounced sensitivity of both data efficiency and final performance to algorithmic hyper-parameters, network capacity, and training budgets. On-policy schemes such as PPO consume millions of interactions while discarding past experience, making them data-hungry and fragile in rare or edge-case situations; shifting to off-policy algorithms has been proposed as a remedy \cite{Learning_Urban_Driving_Policies_using_Deep_Reinforcement_Learning}. The absence of meta-RL modules limits cross-task adaptability \cite{Deductive_Reinforcement_Learning}. Even within off-policy replay buffers, performance can hinge on finely tuned parameters: Focused Experience Replay requires careful selection of the sampling-spread coefficient $\beta_1$; poor choices induce premature convergence or loss of behavioural diversity \cite{Enhanced_Off-Policy_Reinforcement_Learning_With_Focused_Experience_Replay}. Maximum-entropy variants show similar brittleness—larger batch sizes markedly reduce post-turn collisions and lane departures \cite{khalil_exploiting_2023}.
Constraints on hardware and training time further undermine stability: limited GPUs preclude use of richer perception branches, while truncated training runs leave DQN agents well short of convergence \cite{Deep_Reinforcement_Learning_Control_Strategy_at_Roundabout_for_i-CAR, khalil_exploiting_2023}. Existing CNN backbones are rarely optimised for urban scenes; Neural Architecture Search (NAS) is flagged as future work to mitigate this mismatch \cite{jaafraContextAwareAutonomousDriving2019}. Researchers also report vanishing gradients and slow convergence during feature extraction \cite{aghdasian_autonomous_2023}, and even expert distillation models plateau below leaderboard ceilings—suggesting that larger or more expressive networks are still needed \cite{End-to-End_Urban_Driving_by_Imitating_a_Reinforcement_Learning_Coach}. Collectively, these findings highlight the fragility of current pipelines to hyper-parameter sensitivity and resource limitations, underscoring the need for systematic sensitivity studies, automated tuning, and more compute-efficient training algorithms to achieve robust, transferable performance.

\subsection{Insufficient Real-World Validation}

To date, the overwhelming majority of deep-RL driving studies remain confined to simulation, leaving their real-world generalisation essentially unverified. Several groups explicitly defer prototype deployment to future work: ROS-based controllers are good choice for transfer to an NVIDIA-powered test vehicle \cite{Deep_Reinforcement_Learning_based_control_algorithms:_Training_and_validation_using_the_ROS_Framework}, affordance-encoded policies intend to retrain on real camera feeds before road world deployment \cite{End-to-End_Model-Free_Reinforcement_Learning_for_Urban_Driving}, and both deductive RL, knowledge-enhanced, and decision making frameworks acknowledge that only simulated benchmarks have been considered so far \cite{Deductive_Reinforcement_Learning,Autonomous_Driving_via_Knowledge-Enhanced,gutierrez-morenoDecisionMakingAutonomous2024}. Simulation-only evaluation also characterises work on urban driving policies, intersection navigation, emergency-vehicle manoeuvring, and hybrid RL–planning pipelines, all of which report no physical experiments \cite{Learning_Urban_Driving_Policies_using_Deep_Reinforcement_Learning,Integrating_Deep_Reinforcement_Learning_with_Model-based_Path_Planners_for_Automated_Driving,Urban_Autonomous_Driving_of_Emergency_Vehicles_with_Reinforcement_Learning}. Even studies that build custom campus maps or golf-cart prototypes have yet to collect any real sensor data \cite{Using_the_CARLA_Simulator_to_Train_A_Deep_Q_Self-Driving}.
The impracticality of deploying Model Free policies like SAC and PPO in the real world is apparent as they require large amounts of training data and they might still deviate from the intended trajectory within short amount of time  \cite{frauenknecht_data-efficient_2023, zhao_end--end_2024}. Parking controllers trained for 200 simulated episodes might fail in richer layouts, underscoring the gap between simulated and real parking environment \cite{wu_reinforcement_2023}. Likewise, control studies considering the driving styles still rely on hand-tuned rewards that must be recalibrated with human driving data before field tests \cite{liao_lateral_2023}. Finally, uncertainty-aware decision modules double computational latency and have so far been exercised only in high-fidelity simulators \cite{noauthor_explainable_nodate}.
Across all these efforts, real-world scalability, robustness to sensor noise, and compliance with traffic regulations remain open challenges; comprehensive road trials are needed to validate simulated gains and to expose failure modes that only emerge in uncontrolled environments.

\subsection{Traffic-rule and social-norm compliance}

Current decision-making frameworks for autonomous driving lack rigorous mechanisms for traffic-rule and social-norm adherence. Most RL pipelines optimise for task success without an explicit rule-checking layer, leaving the legality of generated manoeuvres uncertain \cite{A_Decision_Control_Method, deshpande_navigation_2021, jin_vwpefficient_2024}. Some studies likewise neglect communicative cues—such as turn indicators and brake lights—that are vital for inferring the intentions of surrounding vehicles \cite{Integration_of_Motion_Prediction_with_End-to-end_Latent_RL_for_Self-Driving_Vehicles}. Behavioural models of other road users remain overly homogeneous, focusing on kinematic states while ignoring cultural differences, driver habits and situational social preferences \cite{li_decision-making_2024}. Accordingly, future research needs to embed formal traffic regulations directly into optimisation objectives or enforce them through constraint-satisfaction layers, enrich training environments with heterogeneous, rule-following agents to achieve robust social-norm compliance and legally safe behaviour \cite{deshpande_navigation_2021, jin_vwpefficient_2024}.

\subsection{Performance Gap}

Despite recent advances, the decision-making and control policies reported in the current literature remain noticeably inferior to both experienced human drivers and the best-performing machine baselines. Deep Imitative Reinforcement Learning (DIRL), for example, offers markedly better data efficiency and robustness, yet its ultimate driving proficiency still falls short of expert human performance, suggesting that policy expressiveness and learning capacity are insufficient for the complexities of real-world traffic \cite{Vision-Based_Autonomous_Car_Racing_Using_Deep_Imitative_Reinforcement_Learning}. In comparisons with SOTA such as CIRL \cite{Liang_2018_ECCV}, CAL \cite{sauer2018conditionalaffordancelearningdriving} and CIRLS \cite{Codevilla_2019_ICCV}, some proposed methods underscoring persistent deficiencies in learning efficiency, generalisation and sample usage \cite{Reinforced_Curriculum_Learning_For_Autonomous_Driving}. When exposed to denser traffic or non-standard intersections, trained agents exhibit erratic interaction handling, reflecting limited capacity to model the negotiation dynamics of multi-agent driving \cite{Deep-Reinforcement-Learning-Based_Driving_Policy_at_Intersections}. At the control level, lateral tracking accuracy deteriorates on curved segments, indicating an incomplete internalisation of vehicle dynamics that would likely be accentuated on a physical platform \cite{li_dynamic_2022}.
Finally, while several frameworks optimise for efficiency, they seldom incorporate explicit notions of decision rationality, interpretability or passenger trust, leaving the qualitative soundness of the chosen manoeuvres largely unexamined \cite{Decision_Making_for_Autonomous_Driving_Via_Multimodal_Transforme}. Addressing these shortcomings—by benchmarking against diverse weather and traffic scenarios, tightening integration of vehicle-dynamics knowledge, and enforcing interpretable decision criteria—remains essential for closing the gap between current autonomous policies, state-of-the-art baselines and human expertise.

\subsection{Limited Behavior Diversity (Different Manuvers)}

Existing deep-RL and imitation-learning pipelines exhibit a narrow behavioural repertoire that falls well short of the manoeuvre diversity required for dependable urban autonomy. Most studies confine training and evaluation to elementary tasks—lane keeping, traffic-light compliance or intersection handling—thereby sidestepping more interactive operations such as negotiated lane changes, roundabout circulation or multi-agent gap selection in heterogeneous traffic streams \cite{Learning_Urban_Driving_Policies_using_Deep_Reinforcement_Learning, gutierrez-morenoDecisionMakingAutonomous2024}. Heavy reliance on manually crafted production rules further constrains behavioural coverage: encoding human cognition by hand is labour-intensive, difficult to scale and ill-suited to novel situations that deviate from the rule base \cite{Cognitive_Reinforcement_Learning}. Even when consistency is achieved across several CARLA towns, overall performance still lags behind on finer-grained tasks, underscoring poor manoeuvre generalisation \cite{Reinforced_Curriculum_Learning_For_Autonomous_Driving}.
The experimental landscape is likewise limited. Many papers validate only one or two scenario templates—for instances, T-intersections, or five-lane highways —without pedestrians, irregular geometries or conflicting traffic patterns \cite{A_Decision_Control_Method, Deep_Reinforcement_Learning_for_Autonomous_Vehicle_Intersection_Navigation, A_Deep_Q-Network_Reinforcement_Learning-Based,frijiDQNBasedAutonomousCarFollowing2020,zhao_real-time_2021,martinez_gomez_temporal_2023,tsaiAutonomousVehicleFollowingTechnique2023}. Training often depends on CARLA’s rule-abiding autopilot or a small, behaviourally homogeneous actor set, depriving agents of exposure to jaywalking pedestrians, reckless drivers or dense multi-class flows \cite{Integration_of_Motion_Prediction_with_End-to-end_Latent_RL_for_Self-Driving_Vehicles, Using_the_CARLA_Simulator_to_Train_A_Deep_Q_Self-Driving}.
Although incremental improvements are reported when additional episodes or temporal abstractions are introduced, authors themselves acknowledge the need for vastly richer scenario catalogues, higher-fidelity multi-agent interactions and more powerful high-level decision modules before reliable coverage of urban manoeuvre diversity can be claimed \cite{elallid_dqn-based_2022, peng_imitative_2020, wu_proximal_2023, ahmed_policy-based_2022,Addressing_Lane_Keeping_and_Intersections}.
Addressing these shortcomings will require large-scale, heterogeneous simulation suites that blend rule-breaking agents with stochastic pedestrians, formal integration of lateral and longitudinal decision layers, and benchmark protocols that span the full gamut of urban driving behaviours—from complex merges to cooperative gap-sharing—in order to deliver policies that generalise beyond narrowly scripted tasks.

\section{conclusion}
We have presented a comprehensive survey of RL research in the CARLA simulator, covering nearly one hundred peer-reviewed studies spanning algorithms, state and action spaces, reward design, terminal conditions, and evaluation metrics. By organizing these contributions along common axes, we have provided newcomers with a clear map of the CARLA landscape and offered seasoned researchers a critical synthesis of where the field stands today.

Three prominent trends emerge from our review. First, CARLA-based work remains dominated by model-free approaches—over 80\% of studies rely on variants of PPO, SAC, DQN, DDPG, or TD3—while model-based and hybrid methods are comparatively under-explored. Second, although bird’s-eye-view and semantically segmented images afford richer spatial structure, most works still employ kinematic-based or front-camera RGB state representations. Third, while a handful of standardized metrics (e.g., collision rate, lane deviation, success rate) recur across studies, many papers introduce bespoke evaluation measures tailored to specific tasks or scenarios, which can hinder direct comparison.

Despite these advances, several critical challenges remain. Sample efficiency continues to be a bottleneck, as state-of-the-art algorithms demand extensive trial-and-error even in simulation, limiting their real-world applicability. Insufficient real-world validation has left most CARLA-trained policies untested outside the simulator, obscuring their performance under true driving conditions. Generalization across towns, weather conditions, and sensor modalities remains elusive, and the sim-to-real gap persists despite early efforts in domain randomization. Safety guarantees and risk-sensitive decision making are seldom integrated into core learning pipelines, while sparse or poorly shaped reward functions further destabilize training. Perception studies often rely on a single modality—typically RGB imagery—neglecting the complementary strengths of LiDAR, RADAR, and GPS. Finally, traffic-rule compliance and social-norm adherence are rarely embedded in learned policies, and most evaluations use simplified single-agent scenarios that fail to capture the complexity of real urban environments.

Addressing these gaps will require concerted efforts on multiple fronts. First, model-based and hybrid approaches—where learned dynamics models guide planning—hold promise for dramatically improving sample efficiency and accelerating convergence. Second, integrating formal safety constraints, uncertainty estimation, and risk-aware objectives into RL algorithms will be essential for certifiable autonomy. Third, a push toward standardized benchmarks, including multi-town, multi-weather scenario suites with unified evaluation metrics, will enable fair comparisons and reproducible progress. Finally, bridging the sim-to-real divide will necessitate richer domain-adaptation techniques, adversarial curricula, and real-vehicle pilot deployments to uncover and mitigate failure modes that only emerge outside the simulator.

\section*{Declarations}

\subsection*{Funding}
This research received no external funding.

\subsection*{Conflicts of interest/Competing interests}
The authors declare that they have no conflicts of interest or competing interests.

\subsection*{Availability of data and material (data transparency)}
Not applicable.

\subsection*{Code availability (software application or custom code)}
Not applicable.

\subsection*{Authors’ contributions}
Elahe Delavari and Feeza Khan Khanzada contributed equally to this work. Both authors read and extracted information from the reviewed papers. Elahe Delavari prepared some of the tables and carried out the final edits and modifications for submission. Prof. Jaerock Kwon verified the correctness of the information and suggested revisions. All authors read and approved the final manuscript.

\subsection*{Acknowledgements}
Not applicable.

\bibliography{Bibliography.bib}


\begin{thebibliography}{134}
\ifx \bisbn   \undefined \def \bisbn  #1{ISBN #1}\fi
\ifx \binits  \undefined \def \binits#1{#1}\fi
\ifx \bauthor  \undefined \def \bauthor#1{#1}\fi
\ifx \batitle  \undefined \def \batitle#1{#1}\fi
\ifx \bjtitle  \undefined \def \bjtitle#1{#1}\fi
\ifx \bvolume  \undefined \def \bvolume#1{\textbf{#1}}\fi
\ifx \byear  \undefined \def \byear#1{#1}\fi
\ifx \bissue  \undefined \def \bissue#1{#1}\fi
\ifx \bfpage  \undefined \def \bfpage#1{#1}\fi
\ifx \blpage  \undefined \def \blpage #1{#1}\fi
\ifx \burl  \undefined \def \burl#1{\textsf{#1}}\fi
\ifx \doiurl  \undefined \def \doiurl#1{\url{https://doi.org/#1}}\fi
\ifx \betal  \undefined \def \betal{\textit{et al.}}\fi
\ifx \binstitute  \undefined \def \binstitute#1{#1}\fi
\ifx \binstitutionaled  \undefined \def \binstitutionaled#1{#1}\fi
\ifx \bctitle  \undefined \def \bctitle#1{#1}\fi
\ifx \beditor  \undefined \def \beditor#1{#1}\fi
\ifx \bpublisher  \undefined \def \bpublisher#1{#1}\fi
\ifx \bbtitle  \undefined \def \bbtitle#1{#1}\fi
\ifx \bedition  \undefined \def \bedition#1{#1}\fi
\ifx \bseriesno  \undefined \def \bseriesno#1{#1}\fi
\ifx \blocation  \undefined \def \blocation#1{#1}\fi
\ifx \bsertitle  \undefined \def \bsertitle#1{#1}\fi
\ifx \bsnm \undefined \def \bsnm#1{#1}\fi
\ifx \bsuffix \undefined \def \bsuffix#1{#1}\fi
\ifx \bparticle \undefined \def \bparticle#1{#1}\fi
\ifx \barticle \undefined \def \barticle#1{#1}\fi
\bibcommenthead
\ifx \bconfdate \undefined \def \bconfdate #1{#1}\fi
\ifx \botherref \undefined \def \botherref #1{#1}\fi
\ifx \url \undefined \def \url#1{\textsf{#1}}\fi
\ifx \bchapter \undefined \def \bchapter#1{#1}\fi
\ifx \bbook \undefined \def \bbook#1{#1}\fi
\ifx \bcomment \undefined \def \bcomment#1{#1}\fi
\ifx \oauthor \undefined \def \oauthor#1{#1}\fi
\ifx \citeauthoryear \undefined \def \citeauthoryear#1{#1}\fi
\ifx \endbibitem  \undefined \def \endbibitem {}\fi
\ifx \bconflocation  \undefined \def \bconflocation#1{#1}\fi
\ifx \arxivurl  \undefined \def \arxivurl#1{\textsf{#1}}\fi
\csname PreBibitemsHook\endcsname

%%% 1
\bibitem[\protect\citeauthoryear{{National Highway Traffic Safety Administration}}{2015}]{nhtsa2015crashes}
\begin{botherref}
\oauthor{\bsnm{{National Highway Traffic Safety Administration}}}:
Critical reasons for crashes investigated in the national motor vehicle crash causation survey.
Technical Report DOT HS 812 115,
U.S. Department of Transportation
(2015).
Accessed: 2025-04-09.
\url{https://crashstats.nhtsa.dot.gov/Api/Public/ViewPublication/812115}
\end{botherref}
\endbibitem

%%% 2
\bibitem[\protect\citeauthoryear{Sallab et~al.}{2017}]{Sallab_2017}
\begin{barticle}
\bauthor{\bsnm{Sallab}, \binits{A.E.}},
\bauthor{\bsnm{Abdou}, \binits{M.}},
\bauthor{\bsnm{Perot}, \binits{E.}},
\bauthor{\bsnm{Yogamani}, \binits{S.}}:
\batitle{Deep reinforcement learning framework for autonomous driving}.
\bjtitle{Electronic Imaging}
\bvolume{29}(\bissue{19}),
\bfpage{70}--\blpage{76}
(\byear{2017})
\doiurl{10.2352/issn.2470-1173.2017.19.avm-023}
\end{barticle}
\endbibitem

%%% 3
\bibitem[\protect\citeauthoryear{Dosovitskiy et~al.}{2017}]{CARLA}
\begin{bchapter}
\bauthor{\bsnm{Dosovitskiy}, \binits{A.}},
\bauthor{\bsnm{Ros}, \binits{G.}},
\bauthor{\bsnm{Codevilla}, \binits{F.}},
\bauthor{\bsnm{Lopez}, \binits{A.}},
\bauthor{\bsnm{Koltun}, \binits{V.}}:
\bctitle{{CARLA}: {An} {Open} {Urban} {Driving} {Simulator}}.
In: \bbtitle{Proceedings of the 1st {Annual} {Conference} on {Robot} {Learning}},
pp. \bfpage{1}--\blpage{16}.
\bpublisher{PMLR}, \blocation{???}
(\byear{2017}).
\bcomment{ISSN: 2640-3498}.
\burl{https://proceedings.mlr.press/v78/dosovitskiy17a.html}
Accessed 2024-07-11
\end{bchapter}
\endbibitem

%%% 4
\bibitem[\protect\citeauthoryear{Wu et~al.}{2022}]{wu_lane_2022}
\begin{bchapter}
\bauthor{\bsnm{Wu}, \binits{Y.}},
\bauthor{\bsnm{Yin}, \binits{Z.}},
\bauthor{\bsnm{Yu}, \binits{J.}},
\bauthor{\bsnm{Zhang}, \binits{M.}}:
\bctitle{Lane {Change} {Decision}-{Making} through {Deep} {Reinforcement} {Learning} with {Driver}'s {Inputs}}.
In: \bbtitle{2022 {IEEE} 7th {International} {Conference} on {Intelligent} {Transportation} {Engineering} ({ICITE})},
pp. \bfpage{314}--\blpage{319}
(\byear{2022}).
\doiurl{10.1109/ICITE56321.2022.10101421} .
\burl{https://ieeexplore.ieee.org/document/10101421/?arnumber=10101421}
Accessed 2025-01-10
\end{bchapter}
\endbibitem

%%% 5
\bibitem[\protect\citeauthoryear{Goel and Chauhan}{2021}]{goel_adaptive_2021}
\begin{bchapter}
\bauthor{\bsnm{Goel}, \binits{A.}},
\bauthor{\bsnm{Chauhan}, \binits{S.}}:
\bctitle{Adaptive {Look}-ahead distance for {Pure} {Pursuit} {Controller} with {Deep} {Reinforcement} {Learning} {Techniques}}.
In: \bbtitle{Advances in {Robotics} - 5th {International} {Conference} of {The} {Robotics} {Society}},
pp. \bfpage{1}--\blpage{5}.
\bpublisher{ACM},
\blocation{Kanpur India}
(\byear{2021}).
\doiurl{10.1145/3478586.3478600} .
\burl{https://dl.acm.org/doi/10.1145/3478586.3478600}
Accessed 2025-01-10
\end{bchapter}
\endbibitem

%%% 6
\bibitem[\protect\citeauthoryear{Youssef and Houda}{2019}]{youssef_deep_2019}
\begin{bchapter}
\bauthor{\bsnm{Youssef}, \binits{F.}},
\bauthor{\bsnm{Houda}, \binits{B.}}:
\bctitle{Deep reinforcement learning with external control: self-driving car application}.
In: \bbtitle{Proceedings of the 4th {International} {Conference} on {Smart} {City} {Applications}},
pp. \bfpage{1}--\blpage{7}.
\bpublisher{ACM},
\blocation{Casablanca Morocco}
(\byear{2019}).
\doiurl{10.1145/3368756.3369038} .
\burl{https://dl.acm.org/doi/10.1145/3368756.3369038}
Accessed 2025-01-10
\end{bchapter}
\endbibitem

%%% 7
\bibitem[\protect\citeauthoryear{Carton et~al.}{2021}]{carton_using_2021}
\begin{bchapter}
\bauthor{\bsnm{Carton}, \binits{F.}},
\bauthor{\bsnm{Filliat}, \binits{D.}},
\bauthor{\bsnm{Rabarisoa}, \binits{J.}},
\bauthor{\bsnm{Pham}, \binits{Q.C.}}:
\bctitle{Using {Semantic} {Information} to {Improve} {Generalization} of {Reinforcement} {Learning} {Policies} for {Autonomous} {Driving}}.
In: \bbtitle{2021 {IEEE} {Winter} {Conference} on {Applications} of {Computer} {Vision} {Workshops} ({WACVW})},
pp. \bfpage{144}--\blpage{151}.
\bpublisher{IEEE},
\blocation{Waikola, HI, USA}
(\byear{2021}).
\doiurl{10.1109/WACVW52041.2021.00020} .
\burl{https://ieeexplore.ieee.org/document/9407810/}
Accessed 2025-01-10
\end{bchapter}
\endbibitem

%%% 8
\bibitem[\protect\citeauthoryear{Mohammed et~al.}{2024}]{mohammed_unified_2024}
\begin{bchapter}
\bauthor{\bsnm{Mohammed}, \binits{S.}},
\bauthor{\bsnm{Argun}, \binits{A.}},
\bauthor{\bsnm{Ascheid}, \binits{G.}}:
\bctitle{A {Unified} {Approach} to {Autonomous} {Driving} in a {High}-{Fidelity} {Simulator} {Using} {Vision}-{Based} {Reinforcement} {Learning}}.
In: \bbtitle{2024 {IEEE}/{SICE} {International} {Symposium} on {System} {Integration} ({SII})},
pp. \bfpage{1093}--\blpage{1098}
(\byear{2024}).
\doiurl{10.1109/SII58957.2024.10417385} .
\bcomment{ISSN: 2474-2325}.
\burl{https://ieeexplore.ieee.org/document/10417385/?arnumber=10417385}
Accessed 2025-01-10
\end{bchapter}
\endbibitem

%%% 9
\bibitem[\protect\citeauthoryear{Zhang and Nakamoto}{2022}]{zhang_toward_2022}
\begin{bchapter}
\bauthor{\bsnm{Zhang}, \binits{M.}},
\bauthor{\bsnm{Nakamoto}, \binits{Y.}}:
\bctitle{Toward ensuring better learning performance in reinforcement learning}.
In: \bbtitle{2022 {Tenth} {International} {Symposium} on {Computing} and {Networking} {Workshops} ({CANDARW})},
pp. \bfpage{134}--\blpage{139}
(\byear{2022}).
\doiurl{10.1109/CANDARW57323.2022.00037} .
\bcomment{ISSN: 2832-1324}.
\burl{https://ieeexplore.ieee.org/document/10062717/?arnumber=10062717}
Accessed 2025-01-10
\end{bchapter}
\endbibitem

%%% 10
\bibitem[\protect\citeauthoryear{Yang et~al.}{2022}]{yang_decision-making_2022}
\begin{bchapter}
\bauthor{\bsnm{Yang}, \binits{Z.}},
\bauthor{\bsnm{Pei}, \binits{X.}},
\bauthor{\bsnm{Xu}, \binits{J.}},
\bauthor{\bsnm{Zhang}, \binits{X.}},
\bauthor{\bsnm{Xi}, \binits{W.}}:
\bctitle{Decision-making in {Autonomous} {Driving} by {Reinforcement} {Learning} {Combined} with {Planning} \& control}.
In: \bbtitle{2022 6th {CAA} {International} {Conference} on {Vehicular} {Control} and {Intelligence} ({CVCI})},
pp. \bfpage{1}--\blpage{6}
(\byear{2022}).
\doiurl{10.1109/CVCI56766.2022.9964691} .
\burl{https://ieeexplore.ieee.org/document/9964691/?arnumber=9964691}
Accessed 2025-01-10
\end{bchapter}
\endbibitem

%%% 11
\bibitem[\protect\citeauthoryear{Wang et~al.}{2023}]{wang_vision-based_2023}
\begin{barticle}
\bauthor{\bsnm{Wang}, \binits{J.}},
\bauthor{\bsnm{Sun}, \binits{H.}},
\bauthor{\bsnm{Zhu}, \binits{C.}}:
\batitle{Vision-{Based} {Autonomous} {Driving}: {A} {Hierarchical} {Reinforcement} {Learning} {Approach}}.
\bjtitle{IEEE Transactions on Vehicular Technology}
\bvolume{72}(\bissue{9}),
\bfpage{11213}--\blpage{11226}
(\byear{2023})
\doiurl{10.1109/TVT.2023.3266940} .
\bcomment{Conference Name: IEEE Transactions on Vehicular Technology}.
Accessed 2025-01-10
\end{barticle}
\endbibitem

%%% 12
\bibitem[\protect\citeauthoryear{Wei et~al.}{2023}]{wei_continual_2023}
\begin{bchapter}
\bauthor{\bsnm{Wei}, \binits{D.}},
\bauthor{\bsnm{Xing}, \binits{J.}},
\bauthor{\bsnm{Yang}, \binits{S.}},
\bauthor{\bsnm{Lu}, \binits{Y.}},
\bauthor{\bsnm{Huang}, \binits{Y.}}:
\bctitle{Continual {Reinforcement} {Learning} for {Autonomous} {Driving} with {Application} on {Velocity} {Control} under {Various} {Environment}}.
In: \bbtitle{2023 7th {CAA} {International} {Conference} on {Vehicular} {Control} and {Intelligence} ({CVCI})},
pp. \bfpage{1}--\blpage{8}
(\byear{2023}).
\doiurl{10.1109/CVCI59596.2023.10397147} .
\burl{https://ieeexplore.ieee.org/document/10397147/?arnumber=10397147}
Accessed 2025-01-10
\end{bchapter}
\endbibitem

%%% 13
\bibitem[\protect\citeauthoryear{Zhu et~al.}{2023}]{zhu_rita_2023}
\begin{bchapter}
\bauthor{\bsnm{Zhu}, \binits{Z.}},
\bauthor{\bsnm{Zhang}, \binits{S.}},
\bauthor{\bsnm{Zhuang}, \binits{Y.}},
\bauthor{\bsnm{Liu}, \binits{Y.}},
\bauthor{\bsnm{Liu}, \binits{M.}},
\bauthor{\bsnm{Gong}, \binits{Z.}},
\bauthor{\bsnm{Kai}, \binits{S.}},
\bauthor{\bsnm{Gu}, \binits{Q.}},
\bauthor{\bsnm{Wang}, \binits{B.}},
\bauthor{\bsnm{Cheng}, \binits{S.}},
\bauthor{\bsnm{Wang}, \binits{X.}},
\bauthor{\bsnm{Hao}, \binits{J.}},
\bauthor{\bsnm{Yu}, \binits{Y.}}:
\bctitle{{RITA}: {Boost} {Driving} {Simulators} with {Realistic} {Interactive} {Traffic} {Flow}}.
In: \bbtitle{Proceedings of the {Fifth} {International} {Conference} on {Distributed} {Artificial} {Intelligence}}.
\bsertitle{{DAI} '23},
pp. \bfpage{1}--\blpage{10}.
\bpublisher{Association for Computing Machinery},
\blocation{New York, NY, USA}
(\byear{2023}).
\doiurl{10.1145/3627676.3627681} .
\burl{https://doi.org/10.1145/3627676.3627681}
Accessed 2025-01-10
\end{bchapter}
\endbibitem

%%% 14
\bibitem[\protect\citeauthoryear{Li et~al.}{2024}]{Think2Drive}
\begin{botherref}
\oauthor{\bsnm{Li}, \binits{Q.}},
\oauthor{\bsnm{Jia}, \binits{X.}},
\oauthor{\bsnm{Wang}, \binits{S.}},
\oauthor{\bsnm{Yan}, \binits{J.}}:
{{Think2Drive}}: {{Efficient Reinforcement Learning}} by {{Thinking}} in {{Latent World Model}} for {{Quasi-Realistic Autonomous Driving}} (in {{CARLA-v2}}).
arXiv.
Comment: Accepted by ECCV 2024
(2024).
\doiurl{10.48550/arXiv.2402.16720}
\end{botherref}
\endbibitem

%%% 15
\bibitem[\protect\citeauthoryear{Pan et~al.}{2023}]{Model-Based_Reinforcement_Learning_with_Isolated_Imaginations}
\begin{botherref}
\oauthor{\bsnm{Pan}, \binits{M.}},
\oauthor{\bsnm{Zhu}, \binits{X.}},
\oauthor{\bsnm{Zheng}, \binits{Y.}},
\oauthor{\bsnm{Wang}, \binits{Y.}},
\oauthor{\bsnm{Yang}, \binits{X.}}:
Model-{{Based Reinforcement Learning}} with {{Isolated Imaginations}}.
arXiv.
Comment: arXiv admin note: text overlap with arXiv:2205.13817
(2023).
\doiurl{10.48550/arXiv.2303.14889}
\end{botherref}
\endbibitem

%%% 16
\bibitem[\protect\citeauthoryear{Cai et~al.}{2021}]{Vision-Based_Autonomous_Car_Racing_Using_Deep_Imitative_Reinforcement_Learning}
\begin{botherref}
\oauthor{\bsnm{Cai}, \binits{P.}},
\oauthor{\bsnm{Wang}, \binits{H.}},
\oauthor{\bsnm{Huang}, \binits{H.}},
\oauthor{\bsnm{Liu}, \binits{Y.}},
\oauthor{\bsnm{Liu}, \binits{M.}}:
Vision-{{Based Autonomous Car Racing Using Deep Imitative Reinforcement Learning}}.
arXiv.
Comment: 8 pages, 8 figures. IEEE Robotics and Automation Letters (RA-L) \& IROS 2021
(2021).
\doiurl{10.48550/arXiv.2107.08325}
\end{botherref}
\endbibitem

%%% 17
\bibitem[\protect\citeauthoryear{Chen et~al.}{2021}]{Learning_to_drive_from_a_world_on_rails}
\begin{botherref}
\oauthor{\bsnm{Chen}, \binits{D.}},
\oauthor{\bsnm{Koltun}, \binits{V.}},
\oauthor{\bsnm{Kr{\"a}henb{\"u}hl}, \binits{P.}}:
Learning to Drive from a World on Rails.
arXiv.
Comment: Paper published in ICCV 2021(Oral); Code and data available at: https://dotchen.github.io/world\_on\_rails/
(2021).
\doiurl{10.48550/arXiv.2105.00636}
\end{botherref}
\endbibitem

%%% 18
\bibitem[\protect\citeauthoryear{Wang et~al.}{2024}]{Autonomous_Driving_via_Knowledge-Enhanced}
\begin{botherref}
\oauthor{\bsnm{Wang}, \binits{C.}},
\oauthor{\bsnm{Zhou}, \binits{S.}},
\oauthor{\bsnm{Wang}, \binits{L.}},
\oauthor{\bsnm{Lu}, \binits{Z.}},
\oauthor{\bsnm{Wu}, \binits{C.}},
\oauthor{\bsnm{Wen}, \binits{X.}},
\oauthor{\bsnm{Shou}, \binits{G.}}:
Autonomous {{Driving}} via {{Knowledge-Enhanced Safe Reinforcement Learning}}.
IEEE Transactions on Intelligent Vehicles,
1--14
(2024)
\doiurl{10.1109/TIV.2024.3412916}
\end{botherref}
\endbibitem

%%% 19
\bibitem[\protect\citeauthoryear{Huang et~al.}{2021}]{Deductive_Reinforcement_Learning}
\begin{barticle}
\bauthor{\bsnm{Huang}, \binits{C.}},
\bauthor{\bsnm{Zhang}, \binits{R.}},
\bauthor{\bsnm{Ouyang}, \binits{M.}},
\bauthor{\bsnm{Wei}, \binits{P.}},
\bauthor{\bsnm{Lin}, \binits{J.}},
\bauthor{\bsnm{Su}, \binits{J.}},
\bauthor{\bsnm{Lin}, \binits{L.}}:
\batitle{Deductive {{Reinforcement Learning}} for {{Visual Autonomous Urban Driving Navigation}}}.
\bjtitle{IEEE Transactions on Neural Networks and Learning Systems}
\bvolume{32}(\bissue{12}),
\bfpage{5379}--\blpage{5391}
(\byear{2021})
\doiurl{10.1109/TNNLS.2021.3109284}
\end{barticle}
\endbibitem

%%% 20
\bibitem[\protect\citeauthoryear{Yurtsever et~al.}{2020}]{Integrating_Deep_Reinforcement_Learning_with_Model-based_Path_Planners_for_Automated_Driving}
\begin{botherref}
\oauthor{\bsnm{Yurtsever}, \binits{E.}},
\oauthor{\bsnm{Capito}, \binits{L.}},
\oauthor{\bsnm{Redmill}, \binits{K.}},
\oauthor{\bsnm{Ozguner}, \binits{U.}}:
Integrating {{Deep Reinforcement Learning}} with {{Model-based Path Planners}} for {{Automated Driving}}.
arXiv.
Comment: 6 pages, 5 figures. Accepted for IEEE Intelligent Vehicles Symposium 2020
(2020).
\doiurl{10.48550/arXiv.2002.00434}
\end{botherref}
\endbibitem

%%% 21
\bibitem[\protect\citeauthoryear{Quigley et~al.}{2009}]{quigley2009ros}
\begin{bchapter}
\bauthor{\bsnm{Quigley}, \binits{M.}},
\bauthor{\bsnm{Conley}, \binits{K.}},
\bauthor{\bsnm{Gerkey}, \binits{B.}},
\bauthor{\bsnm{Faust}, \binits{J.}},
\bauthor{\bsnm{Foote}, \binits{T.}},
\bauthor{\bsnm{Leibs}, \binits{J.}},
\bauthor{\bsnm{Wheeler}, \binits{R.}},
\bauthor{\bsnm{Ng}, \binits{A.Y.}}, \betal:
\bctitle{Ros: an open-source robot operating system}.
In: \bbtitle{ICRA Workshop on Open Source Software},
vol. \bseriesno{3},
p. \bfpage{5}
(\byear{2009}).
\bcomment{Kobe}
\end{bchapter}
\endbibitem

%%% 22
\bibitem[\protect\citeauthoryear{{Autoware Foundation}}{}]{autowarefoundaion_autoware}
\begin{botherref}
\oauthor{\bsnm{{Autoware Foundation}}}:
{Autoware}: Open-source software for autonomous driving.
\url{https://github.com/autowarefoundation/autoware}.
Accessed: June 2025
\end{botherref}
\endbibitem

%%% 23
\bibitem[\protect\citeauthoryear{Kaur et~al.}{2021}]{kaur2021surveysimulatorstestingselfdriving}
\begin{botherref}
\oauthor{\bsnm{Kaur}, \binits{P.}},
\oauthor{\bsnm{Taghavi}, \binits{S.}},
\oauthor{\bsnm{Tian}, \binits{Z.}},
\oauthor{\bsnm{Shi}, \binits{W.}}:
A Survey on Simulators for Testing Self-Driving Cars
(2021).
\url{https://arxiv.org/abs/2101.05337}
\end{botherref}
\endbibitem

%%% 24
\bibitem[\protect\citeauthoryear{Rong et~al.}{2020}]{rong2020lgsvlsimulatorhighfidelity}
\begin{botherref}
\oauthor{\bsnm{Rong}, \binits{G.}},
\oauthor{\bsnm{Shin}, \binits{B.H.}},
\oauthor{\bsnm{Tabatabaee}, \binits{H.}},
\oauthor{\bsnm{Lu}, \binits{Q.}},
\oauthor{\bsnm{Lemke}, \binits{S.}},
\oauthor{\bsnm{Možeiko}, \binits{M.}},
\oauthor{\bsnm{Boise}, \binits{E.}},
\oauthor{\bsnm{Uhm}, \binits{G.}},
\oauthor{\bsnm{Gerow}, \binits{M.}},
\oauthor{\bsnm{Mehta}, \binits{S.}},
\oauthor{\bsnm{Agafonov}, \binits{E.}},
\oauthor{\bsnm{Kim}, \binits{T.H.}},
\oauthor{\bsnm{Sterner}, \binits{E.}},
\oauthor{\bsnm{Ushiroda}, \binits{K.}},
\oauthor{\bsnm{Reyes}, \binits{M.}},
\oauthor{\bsnm{Zelenkovsky}, \binits{D.}},
\oauthor{\bsnm{Kim}, \binits{S.}}:
LGSVL Simulator: A High Fidelity Simulator for Autonomous Driving
(2020).
\url{https://arxiv.org/abs/2005.03778}
\end{botherref}
\endbibitem

%%% 25
\bibitem[\protect\citeauthoryear{Chronis et~al.}{2021}]{chronis_driving_2021}
\begin{bchapter}
\bauthor{\bsnm{Chronis}, \binits{C.}},
\bauthor{\bsnm{Sardianos}, \binits{C.}},
\bauthor{\bsnm{Varlamis}, \binits{I.}},
\bauthor{\bsnm{Michail}, \binits{D.}},
\bauthor{\bsnm{Tserpes}, \binits{K.}}:
\bctitle{A driving profile recommender system for autonomous driving using sensor data and reinforcement learning}.
In: \bbtitle{25th {Pan}-{Hellenic} {Conference} on {Informatics}},
pp. \bfpage{33}--\blpage{38}.
\bpublisher{ACM},
\blocation{Volos Greece}
(\byear{2021}).
\doiurl{10.1145/3503823.3503830} .
\burl{https://dl.acm.org/doi/10.1145/3503823.3503830}
Accessed 2025-01-10
\end{bchapter}
\endbibitem

%%% 26
\bibitem[\protect\citeauthoryear{Udatha et~al.}{}]{udatha_reinforcement_2023}
\begin{botherref}
\oauthor{\bsnm{Udatha}, \binits{S.}},
\oauthor{\bsnm{Lyu}, \binits{Y.}},
\oauthor{\bsnm{Dolan}, \binits{J.}}:
Reinforcement learning with probabilistically safe control barrier functions for ramp merging.
In: 2023 {IEEE} International Conference on Robotics and Automation ({ICRA}),
pp. 5625--5630.
\doiurl{10.1109/ICRA48891.2023.10161418} .
\url{https://ieeexplore.ieee.org/document/10161418}
Accessed 2025-05-06
\end{botherref}
\endbibitem

%%% 27
\bibitem[\protect\citeauthoryear{Lillicrap et~al.}{2019}]{DDPG_paper}
\begin{botherref}
\oauthor{\bsnm{Lillicrap}, \binits{T.P.}},
\oauthor{\bsnm{Hunt}, \binits{J.J.}},
\oauthor{\bsnm{Pritzel}, \binits{A.}},
\oauthor{\bsnm{Heess}, \binits{N.}},
\oauthor{\bsnm{Erez}, \binits{T.}},
\oauthor{\bsnm{Tassa}, \binits{Y.}},
\oauthor{\bsnm{Silver}, \binits{D.}},
\oauthor{\bsnm{Wierstra}, \binits{D.}}:
Continuous control with deep reinforcement learning.
arXiv.
arXiv:1509.02971 [cs]
(2019).
\doiurl{10.48550/arXiv.1509.02971} .
\url{http://arxiv.org/abs/1509.02971}
Accessed 2025-03-27
\end{botherref}
\endbibitem

%%% 28
\bibitem[\protect\citeauthoryear{Silver et~al.}{2014}]{DPG_paper}
\begin{bchapter}
\bauthor{\bsnm{Silver}, \binits{D.}},
\bauthor{\bsnm{Lever}, \binits{G.}},
\bauthor{\bsnm{Heess}, \binits{N.}},
\bauthor{\bsnm{Degris}, \binits{T.}},
\bauthor{\bsnm{Wierstra}, \binits{D.}},
\bauthor{\bsnm{Riedmiller}, \binits{M.}}:
\bctitle{Deterministic {Policy} {Gradient} {Algorithms}}.
In: \bbtitle{Proceedings of the 31st {International} {Conference} on {Machine} {Learning}},
pp. \bfpage{387}--\blpage{395}.
\bpublisher{PMLR}, \blocation{???}
(\byear{2014}).
\bcomment{ISSN: 1938-7228}.
\burl{https://proceedings.mlr.press/v32/silver14.html}
Accessed 2025-03-27
\end{bchapter}
\endbibitem

%%% 29
\bibitem[\protect\citeauthoryear{Mnih et~al.}{2015}]{mnih2015human}
\begin{barticle}
\bauthor{\bsnm{Mnih}, \binits{V.}},
\bauthor{\bsnm{Kavukcuoglu}, \binits{K.}},
\bauthor{\bsnm{Silver}, \binits{D.}},
\bauthor{\bsnm{Rusu}, \binits{A.A.}},
\bauthor{\bsnm{Veness}, \binits{J.}},
\bauthor{\bsnm{Bellemare}, \binits{M.G.}},
\bauthor{\bsnm{Graves}, \binits{A.}},
\bauthor{\bsnm{Riedmiller}, \binits{M.}},
\bauthor{\bsnm{Fidjeland}, \binits{A.K.}},
\bauthor{\bsnm{Ostrovski}, \binits{G.}},
\bauthor{\bsnm{Petersen}, \binits{S.}},
\bauthor{\bsnm{Beattie}, \binits{C.}},
\bauthor{\bsnm{Sadik}, \binits{A.}},
\bauthor{\bsnm{Antonoglou}, \binits{I.}},
\bauthor{\bsnm{King}, \binits{H.}},
\bauthor{\bsnm{Kumaran}, \binits{D.}},
\bauthor{\bsnm{Wierstra}, \binits{D.}},
\bauthor{\bsnm{Legg}, \binits{S.}},
\bauthor{\bsnm{Hassabis}, \binits{D.}}:
\batitle{Human-level control through deep reinforcement learning}.
\bjtitle{Nature}
\bvolume{518}(\bissue{7540}),
\bfpage{529}--\blpage{533}
(\byear{2015})
\doiurl{10.1038/nature14236} .
\bcomment{Publisher: Nature Publishing Group}.
Accessed 2025-03-27
\end{barticle}
\endbibitem

%%% 30
\bibitem[\protect\citeauthoryear{Peng et~al.}{2020}]{peng_imitative_2020}
\begin{bchapter}
\bauthor{\bsnm{Peng}, \binits{M.}},
\bauthor{\bsnm{Gong}, \binits{Z.}},
\bauthor{\bsnm{Sun}, \binits{C.}},
\bauthor{\bsnm{Chen}, \binits{L.}},
\bauthor{\bsnm{Cao}, \binits{D.}}:
\bctitle{Imitative reinforcement learning fusing vision and pure pursuit for self-driving}.
In: \bbtitle{2020 IEEE International Conference on Robotics and Automation (ICRA)},
pp. \bfpage{3298}--\blpage{3304}
(\byear{2020}).
\bcomment{IEEE}
\end{bchapter}
\endbibitem

%%% 31
\bibitem[\protect\citeauthoryear{{P{\'e}rez-Gil} et~al.}{2022}]{Deep_reinforcement_learning_based_control_for_Autonomous_Vehicles_in_CARLA}
\begin{barticle}
\bauthor{\bsnm{{P{\'e}rez-Gil}}, \binits{{\'O}.}},
\bauthor{\bsnm{Barea}, \binits{R.}},
\bauthor{\bsnm{{L{\'o}pez-Guill{\'e}n}}, \binits{E.}},
\bauthor{\bsnm{Bergasa}, \binits{L.M.}},
\bauthor{\bsnm{{G{\'o}mez-Hu{\'e}lamo}}, \binits{C.}},
\bauthor{\bsnm{Guti{\'e}rrez}, \binits{R.}},
\bauthor{\bsnm{{D{\'i}az-D{\'i}az}}, \binits{A.}}:
\batitle{Deep reinforcement learning based control for {{Autonomous Vehicles}} in {{CARLA}}}.
\bjtitle{Multimedia Tools and Applications}
\bvolume{81}(\bissue{3}),
\bfpage{3553}--\blpage{3576}
(\byear{2022})
\doiurl{10.1007/s11042-021-11437-3}
\end{barticle}
\endbibitem

%%% 32
\bibitem[\protect\citeauthoryear{Fu et~al.}{2022}]{Decision_Making_for_Autonomous_Driving_Via_Multimodal_Transforme}
\begin{bchapter}
\bauthor{\bsnm{Fu}, \binits{W.}},
\bauthor{\bsnm{Li}, \binits{Y.}},
\bauthor{\bsnm{Ye}, \binits{Z.}},
\bauthor{\bsnm{Liu}, \binits{Q.}}:
\bctitle{Decision {{Making}} for {{Autonomous Driving Via Multimodal Transformer}} and {{Deep Reinforcement Learning}}}.
In: \bbtitle{2022 {{IEEE International Conference}} on {{Real-time Computing}} and {{Robotics}} ({{RCAR}})},
pp. \bfpage{481}--\blpage{486}
(\byear{2022}).
\doiurl{10.1109/RCAR54675.2022.9872180}
\end{bchapter}
\endbibitem

%%% 33
\bibitem[\protect\citeauthoryear{Chen et~al.}{2021}]{Towards_Autonomous_Driving}
\begin{bchapter}
\bauthor{\bsnm{Chen}, \binits{M.}},
\bauthor{\bsnm{Li}, \binits{Y.}},
\bauthor{\bsnm{Liu}, \binits{Q.}},
\bauthor{\bsnm{Lv}, \binits{S.}},
\bauthor{\bsnm{Xu}, \binits{Y.}},
\bauthor{\bsnm{Liu}, \binits{Y.}}:
\bctitle{Towards {{Autonomous Driving Decision}} by {{Combining Self-attention}} and {{Deep Reinforcement Learning}}}.
In: \bbtitle{2021 {{IEEE International Conference}} on {{Real-time Computing}} and {{Robotics}} ({{RCAR}})},
pp. \bfpage{1110}--\blpage{1115}
(\byear{2021}).
\doiurl{10.1109/RCAR52367.2021.9517610}
\end{bchapter}
\endbibitem

%%% 34
\bibitem[\protect\citeauthoryear{Zhang et~al.}{2021}]{zhangSelflearningLaneKeeping2021}
\begin{bchapter}
\bauthor{\bsnm{Zhang}, \binits{B.}},
\bauthor{\bsnm{Xu}, \binits{C.}},
\bauthor{\bsnm{Su}, \binits{Y.}},
\bauthor{\bsnm{Xu}, \binits{J.}}:
\bctitle{A self-learning lane keeping algorithm}.
In: \bbtitle{ICMLCA 2021; 2nd International Conference on Machine Learning and Computer Application},
pp. \bfpage{1}--\blpage{7}
(\byear{2021})
\end{bchapter}
\endbibitem

%%% 35
\bibitem[\protect\citeauthoryear{Tsai et~al.}{2023}]{tsaiAutonomousVehicleFollowingTechnique2023}
\begin{bchapter}
\bauthor{\bsnm{Tsai}, \binits{J.}},
\bauthor{\bsnm{Chang}, \binits{Y.-T.}},
\bauthor{\bsnm{Chuang}, \binits{P.-H.}},
\bauthor{\bsnm{You}, \binits{Z.}}:
\bctitle{An {{Autonomous Vehicle-Following Technique}} for {{Self-Driving Cars Based}} on the {{Semantic Segmentation Technique}}}.
In: \bbtitle{2023 {{IEEE International Symposium}} on {{Robotic}} and {{Sensors Environments}} ({{ROSE}})},
pp. \bfpage{1}--\blpage{7}
(\byear{2023}).
\doiurl{10.1109/ROSE60297.2023.10410810}
\end{bchapter}
\endbibitem

%%% 36
\bibitem[\protect\citeauthoryear{Doe et~al.}{}]{doe_dsorl_2023}
\begin{botherref}
\oauthor{\bsnm{Doe}, \binits{D.M.}},
\oauthor{\bsnm{Chen}, \binits{D.}},
\oauthor{\bsnm{Han}, \binits{K.}},
\oauthor{\bsnm{Wang}, \binits{H.}},
\oauthor{\bsnm{Xie}, \binits{J.}},
\oauthor{\bsnm{Han}, \binits{Z.}}:
{DSORL}: Data source optimization with reinforcement learning scheme for vehicular named data networks
\textbf{24}(10),
11225--11237
\doiurl{10.1109/TITS.2023.3292033} .
Accessed 2025-05-06
\end{botherref}
\endbibitem

%%% 37
\bibitem[\protect\citeauthoryear{Li et~al.}{}]{li_dynamic_2022}
\begin{botherref}
\oauthor{\bsnm{Li}, \binits{L.}},
\oauthor{\bsnm{Jiang}, \binits{W.}},
\oauthor{\bsnm{Shi}, \binits{M.}},
\oauthor{\bsnm{Wu}, \binits{T.}}:
Dynamic target following control for autonomous vehicles with deep reinforcement learning.
In: 2022 International Conference on Advanced Robotics and Mechatronics ({ICARM}),
pp. 386--391.
\doiurl{10.1109/ICARM54641.2022.9959167} .
\url{https://ieeexplore.ieee.org/document/9959167}
Accessed 2025-05-06
\end{botherref}
\endbibitem

%%% 38
\bibitem[\protect\citeauthoryear{Ahmed et~al.}{}]{ahmed_policy-based_2022}
\begin{botherref}
\oauthor{\bsnm{Ahmed}, \binits{M.}},
\oauthor{\bsnm{Abobakr}, \binits{A.}},
\oauthor{\bsnm{Lim}, \binits{C.P.}},
\oauthor{\bsnm{Nahavandi}, \binits{S.}}:
Policy-based reinforcement learning for training autonomous driving agents in urban areas with affordance learning
\textbf{23}(8),
12562--12571
\doiurl{10.1109/TITS.2021.3115235} .
Accessed 2025-05-06
\end{botherref}
\endbibitem

%%% 39
\bibitem[\protect\citeauthoryear{Cheng et~al.}{2021}]{cheng_longitudinal_2021}
\begin{bchapter}
\bauthor{\bsnm{Cheng}, \binits{S.}},
\bauthor{\bsnm{Wang}, \binits{N.}},
\bauthor{\bsnm{Chen}, \binits{F.}},
\bauthor{\bsnm{Pipe}, \binits{T.}}:
\bctitle{Longitudinal {Driving} {Skills} {Transfer} from {Driver} to {Smart} {Vehicle}}.
In: \bbtitle{2021 26th {International} {Conference} on {Automation} and {Computing} ({ICAC})},
pp. \bfpage{1}--\blpage{6}
(\byear{2021}).
\doiurl{10.23919/ICAC50006.2021.9594177} .
\burl{https://ieeexplore.ieee.org/document/9594177/?arnumber=9594177}
Accessed 2025-01-10
\end{bchapter}
\endbibitem

%%% 40
\bibitem[\protect\citeauthoryear{Schulman et~al.}{2017}]{schulman2017proximal}
\begin{botherref}
\oauthor{\bsnm{Schulman}, \binits{J.}},
\oauthor{\bsnm{Wolski}, \binits{F.}},
\oauthor{\bsnm{Dhariwal}, \binits{P.}},
\oauthor{\bsnm{Radford}, \binits{A.}},
\oauthor{\bsnm{Klimov}, \binits{O.}}:
Proximal {Policy} {Optimization} {Algorithms}.
arXiv.
arXiv:1707.06347 [cs]
(2017).
\doiurl{10.48550/arXiv.1707.06347} .
\url{http://arxiv.org/abs/1707.06347}
Accessed 2025-03-27
\end{botherref}
\endbibitem

%%% 41
\bibitem[\protect\citeauthoryear{Deng et~al.}{2023}]{deng_context_2023}
\begin{bchapter}
\bauthor{\bsnm{Deng}, \binits{Q.}},
\bauthor{\bsnm{Zhao}, \binits{Y.}},
\bauthor{\bsnm{Li}, \binits{R.}},
\bauthor{\bsnm{Hu}, \binits{Q.}},
\bauthor{\bsnm{Liu}, \binits{T.}},
\bauthor{\bsnm{Li}, \binits{R.}}:
\bctitle{Context - {Enhanced} {Meta}-{Reinforcement} {Learning} with {Data}-{Reused} {Adaptation} for {Urban} {Autonomous} {Driving}}.
In: \bbtitle{2023 {International} {Joint} {Conference} on {Neural} {Networks} ({IJCNN})},
pp. \bfpage{1}--\blpage{8}
(\byear{2023}).
\doiurl{10.1109/IJCNN54540.2023.10191187} .
\bcomment{ISSN: 2161-4407}.
\burl{https://ieeexplore.ieee.org/document/10191187/?arnumber=10191187}
Accessed 2025-01-10
\end{bchapter}
\endbibitem

%%% 42
\bibitem[\protect\citeauthoryear{Deng et~al.}{2024}]{deng_context-aware_2024}
\begin{barticle}
\bauthor{\bsnm{Deng}, \binits{Q.}},
\bauthor{\bsnm{Li}, \binits{R.}},
\bauthor{\bsnm{Hu}, \binits{Q.}},
\bauthor{\bsnm{Zhao}, \binits{Y.}},
\bauthor{\bsnm{Li}, \binits{R.}}:
\batitle{Context-{Aware} {Meta}-{RL} {With} {Two}-{Stage} {Constrained} {Adaptation} for {Urban} {Driving}}.
\bjtitle{IEEE Transactions on Vehicular Technology}
\bvolume{73}(\bissue{2}),
\bfpage{1567}--\blpage{1581}
(\byear{2024})
\doiurl{10.1109/TVT.2023.3312495} .
\bcomment{Conference Name: IEEE Transactions on Vehicular Technology}.
Accessed 2025-01-10
\end{barticle}
\endbibitem

%%% 43
\bibitem[\protect\citeauthoryear{Zhao et~al.}{2024}]{zhao_end--end_2024}
\begin{bchapter}
\bauthor{\bsnm{Zhao}, \binits{J.}},
\bauthor{\bsnm{Zhao}, \binits{Y.}},
\bauthor{\bsnm{Li}, \binits{W.}},
\bauthor{\bsnm{Zeng}, \binits{C.}}:
\bctitle{End-to-{End} {Autonomous} {Driving} {Algorithm} {Based} on {PPO} and {Its} {Implementation}}.
In: \bbtitle{2024 {IEEE} 13th {Data} {Driven} {Control} and {Learning} {Systems} {Conference} ({DDCLS})},
pp. \bfpage{1852}--\blpage{1858}
(\byear{2024}).
\doiurl{10.1109/DDCLS61622.2024.10606596} .
\bcomment{ISSN: 2767-9861}.
\burl{https://ieeexplore.ieee.org/document/10606596/?arnumber=10606596}
Accessed 2025-01-10
\end{bchapter}
\endbibitem

%%% 44
\bibitem[\protect\citeauthoryear{Wu and Yuan}{2023}]{wu_proximal_2023}
\begin{bchapter}
\bauthor{\bsnm{Wu}, \binits{Y.}},
\bauthor{\bsnm{Yuan}, \binits{X.}}:
\bctitle{Proximal {Policy} {Optimization}-based {Reinforcement} {Learning} for {End}-to-end {Autonomous} {Driving}}.
In: \bbtitle{2023 38th {Youth} {Academic} {Annual} {Conference} of {Chinese} {Association} of {Automation} ({YAC})},
pp. \bfpage{844}--\blpage{849}
(\byear{2023}).
\doiurl{10.1109/YAC59482.2023.10401381} .
\bcomment{ISSN: 2837-8601}.
\burl{https://ieeexplore.ieee.org/document/10401381/?arnumber=10401381}
Accessed 2025-01-10
\end{bchapter}
\endbibitem

%%% 45
\bibitem[\protect\citeauthoryear{Zhang et~al.}{2021}]{End-to-End_Urban_Driving_by_Imitating_a_Reinforcement_Learning_Coach}
\begin{botherref}
\oauthor{\bsnm{Zhang}, \binits{Z.}},
\oauthor{\bsnm{Liniger}, \binits{A.}},
\oauthor{\bsnm{Dai}, \binits{D.}},
\oauthor{\bsnm{Yu}, \binits{F.}},
\oauthor{\bsnm{Gool}, \binits{L.V.}}:
End-to-{{End Urban Driving}} by {{Imitating}} a {{Reinforcement Learning Coach}}.
arXiv.
Comment: Published at ICCV 2021
(2021).
\doiurl{10.48550/arXiv.2108.08265}
\end{botherref}
\endbibitem

%%% 46
\bibitem[\protect\citeauthoryear{Agarwal et~al.}{2021}]{Learning_Urban_Driving_Policies_using_Deep_Reinforcement_Learning}
\begin{bchapter}
\bauthor{\bsnm{Agarwal}, \binits{T.}},
\bauthor{\bsnm{Arora}, \binits{H.}},
\bauthor{\bsnm{Schneider}, \binits{J.}}:
\bctitle{Learning {{Urban Driving Policies}} using {{Deep Reinforcement Learning}}}.
In: \bbtitle{2021 {{IEEE International Intelligent Transportation Systems Conference}} ({{ITSC}})},
pp. \bfpage{607}--\blpage{614}
(\byear{2021}).
\doiurl{10.1109/ITSC48978.2021.9564412}
\end{bchapter}
\endbibitem

%%% 47
\bibitem[\protect\citeauthoryear{Trumpp et~al.}{2023}]{Efficient_Learning_of_Urban_Driving}
\begin{botherref}
\oauthor{\bsnm{Trumpp}, \binits{R.}},
\oauthor{\bsnm{B{\"u}chner}, \binits{M.}},
\oauthor{\bsnm{Valada}, \binits{A.}},
\oauthor{\bsnm{Caccamo}, \binits{M.}}:
Efficient {{Learning}} of {{Urban Driving Policies Using Bird}}'s-{{Eye-View State Representations}}.
arXiv.
Comment: IEEE International Conference on Intelligent Transportation Systems 2023
(2023).
\doiurl{10.48550/arXiv.2305.19904}
\end{botherref}
\endbibitem

%%% 48
\bibitem[\protect\citeauthoryear{Xing et~al.}{2021}]{Domain_Adaptation_In_Reinforcement_Learning}
\begin{botherref}
\oauthor{\bsnm{Xing}, \binits{J.}},
\oauthor{\bsnm{Nagata}, \binits{T.}},
\oauthor{\bsnm{Chen}, \binits{K.}},
\oauthor{\bsnm{Zou}, \binits{X.}},
\oauthor{\bsnm{Neftci}, \binits{E.}},
\oauthor{\bsnm{Krichmar}, \binits{J.L.}}:
Domain {{Adaptation In Reinforcement Learning Via Latent Unified State Representation}}.
arXiv.
Comment: Accepted by AAAI 2021; Fixed a typo in equation 3
(2021).
\doiurl{10.48550/arXiv.2102.05714}
\end{botherref}
\endbibitem

%%% 49
\bibitem[\protect\citeauthoryear{Anzalone et~al.}{2021}]{Reinforced_Curriculum_Learning_For_Autonomous_Driving}
\begin{bchapter}
\bauthor{\bsnm{Anzalone}, \binits{L.}},
\bauthor{\bsnm{Barra}, \binits{S.}},
\bauthor{\bsnm{Nappi}, \binits{M.}}:
\bctitle{Reinforced {{Curriculum Learning For Autonomous Driving In Carla}}}.
In: \bbtitle{2021 {{IEEE International Conference}} on {{Image Processing}} ({{ICIP}})},
pp. \bfpage{3318}--\blpage{3322}
(\byear{2021}).
\doiurl{10.1109/ICIP42928.2021.9506673}
\end{bchapter}
\endbibitem

%%% 50
\bibitem[\protect\citeauthoryear{Silva and Grassi}{2021}]{Addressing_Lane_Keeping_and_Intersections}
\begin{bchapter}
\bauthor{\bsnm{Silva}, \binits{V.A.S.}},
\bauthor{\bsnm{Grassi}, \binits{V.}}:
\bctitle{Addressing {{Lane Keeping}} and {{Intersections}} using {{Deep Conditional Reinforcement Learning}}}.
In: \bbtitle{2021 {{Latin American Robotics Symposium}} ({{LARS}}), 2021 {{Brazilian Symposium}} on {{Robotics}} ({{SBR}}), and 2021 {{Workshop}} on {{Robotics}} in {{Education}} ({{WRE}})},
pp. \bfpage{330}--\blpage{335}
(\byear{2021}).
\doiurl{10.1109/LARS/SBR/WRE54079.2021.9605436}
\end{bchapter}
\endbibitem

%%% 51
\bibitem[\protect\citeauthoryear{Albilani and Bouzeghoub}{}]{albilani_guided_2023}
\begin{botherref}
\oauthor{\bsnm{Albilani}, \binits{M.}},
\oauthor{\bsnm{Bouzeghoub}, \binits{A.}}:
Guided hierarchical reinforcement learning for safe urban driving.
In: 2023 {IEEE} 35th International Conference on Tools with Artificial Intelligence ({ICTAI}),
pp. 746--753.
\doiurl{10.1109/ICTAI59109.2023.00115} .
{ISSN}: 2375-0197.
\url{https://ieeexplore.ieee.org/document/10356414}
Accessed 2025-05-06
\end{botherref}
\endbibitem

%%% 52
\bibitem[\protect\citeauthoryear{Mart{\'i}nez~G{\'o}mez et~al.}{}]{martinez_gomez_temporal_2023}
\begin{botherref}
\oauthor{\bsnm{Mart{\'i}nez~G{\'o}mez}, \binits{L.M.}},
\oauthor{\bsnm{Daza}, \binits{I.G.}},
\oauthor{\bsnm{Sotelo~V{\'a}zquez}, \binits{M.{\'A}.}}:
Temporal based deep reinforcement learning for crowded lane merging maneuvers.
In: 2023 {IEEE} 26th International Conference on Intelligent Transportation Systems ({ITSC}),
pp. 2764--2769.
\doiurl{10.1109/ITSC57777.2023.10422486} .
{ISSN}: 2153-0017.
\url{https://ieeexplore.ieee.org/document/10422486}
Accessed 2025-05-06
\end{botherref}
\endbibitem

%%% 53
\bibitem[\protect\citeauthoryear{Jin et~al.}{}]{jin_vwpefficient_2024}
\begin{botherref}
\oauthor{\bsnm{Jin}, \binits{Y.-L.}},
\oauthor{\bsnm{Ji}, \binits{Z.-Y.}},
\oauthor{\bsnm{Zeng}, \binits{D.}},
\oauthor{\bsnm{Zhang}, \binits{X.-P.}}:
{VWP}:an efficient {DRL}-based autonomous driving model
\textbf{26},
2096--2108
\doiurl{10.1109/TMM.2022.3177942} .
Accessed 2025-05-06
\end{botherref}
\endbibitem

%%% 54
\bibitem[\protect\citeauthoryear{Shi et~al.}{}]{shi_efficient_2023}
\begin{botherref}
\oauthor{\bsnm{Shi}, \binits{J.}},
\oauthor{\bsnm{Zhang}, \binits{T.}},
\oauthor{\bsnm{Zhan}, \binits{J.}},
\oauthor{\bsnm{Chen}, \binits{S.}},
\oauthor{\bsnm{Xin}, \binits{J.}},
\oauthor{\bsnm{Zheng}, \binits{N.}}:
Efficient lane-changing behavior planning via reinforcement learning with imitation learning initialization.
In: 2023 {IEEE} Intelligent Vehicles Symposium ({IV}),
pp. 1--8.
\doiurl{10.1109/IV55152.2023.10186577} .
{ISSN}: 2642-7214.
\url{https://ieeexplore.ieee.org/document/10186577}
Accessed 2025-05-06
\end{botherref}
\endbibitem

%%% 55
\bibitem[\protect\citeauthoryear{{Guti{\'e}rrez-Moreno} et~al.}{2024}]{gutierrez-morenoDecisionMakingAutonomous2024}
\begin{bchapter}
\bauthor{\bsnm{{Guti{\'e}rrez-Moreno}}, \binits{R.}},
\bauthor{\bsnm{Barea}, \binits{R.}},
\bauthor{\bsnm{{L{\'o}pez-Guill{\'e}n}}, \binits{E.}},
\bauthor{\bsnm{Arango}, \binits{F.}},
\bauthor{\bsnm{Revenga}, \binits{P.}},
\bauthor{\bsnm{Bergasa}, \binits{L.M.}}:
\bctitle{Decision {{Making}} for {{Autonomous Driving Stack}}: {{Shortening}} the {{Gap}} from {{Simulation}} to {{Real-World Implementations}}}.
In: \bbtitle{2024 {{IEEE Intelligent Vehicles Symposium}} ({{IV}})},
pp. \bfpage{3107}--\blpage{3113}
(\byear{2024}).
\doiurl{10.1109/IV55156.2024.10588560}
\end{bchapter}
\endbibitem

%%% 56
\bibitem[\protect\citeauthoryear{Bai et~al.}{2023}]{bai_fen-dqn_2023}
\begin{bchapter}
\bauthor{\bsnm{Bai}, \binits{Y.}},
\bauthor{\bsnm{Du}, \binits{J.}},
\bauthor{\bsnm{Zhang}, \binits{Y.}},
\bauthor{\bsnm{Huang}, \binits{Y.}}:
\bctitle{{FEN}-{DQN}: {An} {End}-to-{End} {Autonomous} {Driving} {Framework} {Based} on {Reinforcement} {Learning} with {Explicit} {Affordance}}.
In: \bbtitle{2023 7th {CAA} {International} {Conference} on {Vehicular} {Control} and {Intelligence} ({CVCI})},
pp. \bfpage{1}--\blpage{6}
(\byear{2023}).
\doiurl{10.1109/CVCI59596.2023.10397338} .
\burl{https://ieeexplore.ieee.org/document/10397338/?arnumber=10397338}
Accessed 2025-01-10
\end{bchapter}
\endbibitem

%%% 57
\bibitem[\protect\citeauthoryear{Muhammed et~al.}{2021}]{muhammed_developing_2021}
\begin{bchapter}
\bauthor{\bsnm{Muhammed}, \binits{A.}},
\bauthor{\bsnm{Essam}, \binits{H.}},
\bauthor{\bsnm{Alber}, \binits{B.}},
\bauthor{\bsnm{Samuel}, \binits{K.}},
\bauthor{\bsnm{Muhammed}, \binits{H.}},
\bauthor{\bsnm{Wagdy}, \binits{M.}},
\bauthor{\bsnm{Khaled}, \binits{N.}},
\bauthor{\bsnm{Fawzy}, \binits{H.}},
\bauthor{\bsnm{Tarek}, \binits{A.}},
\bauthor{\bsnm{AbdelSalam}, \binits{M.}},
\bauthor{\bsnm{El-Kharashi}, \binits{M.W.}}:
\bctitle{Developing {AI} {Agent} with {Functional} {Mockup} {Units} for {Car} {Autonomous} {Navigation}}.
In: \bbtitle{2021 28th {IEEE} {International} {Conference} on {Electronics}, {Circuits}, and {Systems} ({ICECS})},
pp. \bfpage{1}--\blpage{5}
(\byear{2021}).
\doiurl{10.1109/ICECS53924.2021.9665639} .
\burl{https://ieeexplore.ieee.org/document/9665639/?arnumber=9665639}
Accessed 2025-01-10
\end{bchapter}
\endbibitem

%%% 58
\bibitem[\protect\citeauthoryear{Elallid et~al.}{2022}]{elallid_dqn-based_2022}
\begin{bchapter}
\bauthor{\bsnm{Elallid}, \binits{B.B.}},
\bauthor{\bsnm{Benamar}, \binits{N.}},
\bauthor{\bsnm{Mrani}, \binits{N.}},
\bauthor{\bsnm{Rachidi}, \binits{T.}}:
\bctitle{Dqn-based reinforcement learning for vehicle control of autonomous vehicles interacting with pedestrians}.
In: \bbtitle{2022 International Conference on Innovation and Intelligence for Informatics, Computing, and Technologies (3ICT)},
pp. \bfpage{489}--\blpage{493}
(\byear{2022}).
\bcomment{IEEE}
\end{bchapter}
\endbibitem

%%% 59
\bibitem[\protect\citeauthoryear{Chekroun et~al.}{2022}]{GRI:General_Reinforced_Imitation}
\begin{botherref}
\oauthor{\bsnm{Chekroun}, \binits{R.}},
\oauthor{\bsnm{Toromanoff}, \binits{M.}},
\oauthor{\bsnm{Hornauer}, \binits{S.}},
\oauthor{\bsnm{Moutarde}, \binits{F.}}:
{{GRI}}: {{General Reinforced Imitation}} and Its {{Application}} to {{Vision-Based Autonomous Driving}}.
arXiv
(2022).
\doiurl{10.48550/arXiv.2111.08575}
\end{botherref}
\endbibitem

%%% 60
\bibitem[\protect\citeauthoryear{Toromanoff et~al.}{2020}]{End-to-End_Model-Free_Reinforcement_Learning_for_Urban_Driving}
\begin{botherref}
\oauthor{\bsnm{Toromanoff}, \binits{M.}},
\oauthor{\bsnm{Wirbel}, \binits{E.}},
\oauthor{\bsnm{Moutarde}, \binits{F.}}:
End-to-{{End Model-Free Reinforcement Learning}} for {{Urban Driving}} Using {{Implicit Affordances}}.
arXiv.
Comment: Accepted at main conference of CVPR 2020
(2020).
\doiurl{10.48550/arXiv.1911.10868}
\end{botherref}
\endbibitem

%%% 61
\bibitem[\protect\citeauthoryear{Marouane and Saad}{2024}]{Safe_Navigation_Based_on_Deep_Q-Network_Algorithm_Using}
\begin{bchapter}
\bauthor{\bsnm{Marouane}, \binits{C.}},
\bauthor{\bsnm{Saad}, \binits{B.}}:
\bctitle{Safe {{Navigation Based}} on {{Deep Q-Network Algorithm Using}} an {{Improved Control Architecture}}}.
In: \bbtitle{2024 2nd {{International Conference}} on {{Electrical Engineering}} and {{Automatic Control}} ({{ICEEAC}})},
pp. \bfpage{1}--\blpage{6}
(\byear{2024}).
\doiurl{10.1109/ICEEAC61226.2024.10576248}
\end{bchapter}
\endbibitem

%%% 62
\bibitem[\protect\citeauthoryear{{Muhtadin} et~al.}{2023}]{Deep_Reinforcement_Learning_Control_Strategy_at_Roundabout_for_i-CAR}
\begin{bchapter}
\bauthor{\bsnm{{Muhtadin}}},
\bauthor{\bsnm{Meliaz}, \binits{M.R.}},
\bauthor{\bsnm{Dikairono}, \binits{R.}},
\bauthor{\bsnm{Purnama}, \binits{I.K.E.}},
\bauthor{\bsnm{Purnomo}, \binits{M.H.}}:
\bctitle{Deep {{Reinforcement Learning Control Strategy}} at {{Roundabout}} for i-{{CAR Autonomous Car}}}.
In: \bbtitle{2023 {{International Seminar}} on {{Intelligent Technology}} and {{Its Applications}} ({{ISITIA}})},
pp. \bfpage{473}--\blpage{478}
(\byear{2023}).
\doiurl{10.1109/ISITIA59021.2023.10221077}
\end{bchapter}
\endbibitem

%%% 63
\bibitem[\protect\citeauthoryear{Elallid et~al.}{2023}]{A_Reinforcement_Learning_Based_Approach_for_Controlling_Autonomous_Vehicles_in_Complex_Scenarios}
\begin{bchapter}
\bauthor{\bsnm{Elallid}, \binits{B.B.}},
\bauthor{\bsnm{Bagaa}, \binits{M.}},
\bauthor{\bsnm{Benamar}, \binits{N.}},
\bauthor{\bsnm{Mrani}, \binits{N.}}:
\bctitle{A {{Reinforcement Learning Based Approach}} for {{Controlling Autonomous Vehicles}} in {{Complex Scenarios}}}.
In: \bbtitle{2023 {{International Wireless Communications}} and {{Mobile Computing}} ({{IWCMC}})},
pp. \bfpage{1358}--\blpage{1364}
(\byear{2023}).
\doiurl{10.1109/IWCMC58020.2023.10182377}
\end{bchapter}
\endbibitem

%%% 64
\bibitem[\protect\citeauthoryear{May et~al.}{}]{Using_the_CARLA_Simulator_to_Train_A_Deep_Q_Self-Driving}
\begin{botherref}
\oauthor{\bsnm{May}, \binits{J.}},
\oauthor{\bsnm{Poudel}, \binits{S.}},
\oauthor{\bsnm{Hamdan}, \binits{S.}},
\oauthor{\bsnm{Poudel}, \binits{K.}},
\oauthor{\bsnm{Vargas}, \binits{J.}}:
Using the {{CARLA Simulator}} to {{Train A Deep Q Self-Driving Car}} to {{Control}} a {{Real-World Counterpart}} on {{A College Campus}}
\end{botherref}
\endbibitem

%%% 65
\bibitem[\protect\citeauthoryear{Ahmed et~al.}{2021}]{A_Deep_Q-Network_Reinforcement_Learning-Based}
\begin{bchapter}
\bauthor{\bsnm{Ahmed}, \binits{M.}},
\bauthor{\bsnm{Lim}, \binits{C.P.}},
\bauthor{\bsnm{Nahavandi}, \binits{S.}}:
\bctitle{A {{Deep Q-Network Reinforcement Learning-Based Model}} for {{Autonomous Driving}}}.
In: \bbtitle{2021 {{IEEE International Conference}} on {{Systems}}, {{Man}}, and {{Cybernetics}} ({{SMC}})},
pp. \bfpage{739}--\blpage{744}
(\byear{2021}).
\doiurl{10.1109/SMC52423.2021.9658892}
\end{bchapter}
\endbibitem

%%% 66
\bibitem[\protect\citeauthoryear{Clemmons and Jin}{2023}]{Reinforcement_Learning-Based_Guidance_of_Autonomous_Vehicles}
\begin{bchapter}
\bauthor{\bsnm{Clemmons}, \binits{J.}},
\bauthor{\bsnm{Jin}, \binits{Y.-F.}}:
\bctitle{Reinforcement {{Learning-Based Guidance}} of {{Autonomous Vehicles}}}.
In: \bbtitle{2023 24th {{International Symposium}} on {{Quality Electronic Design}} ({{ISQED}})},
pp. \bfpage{1}--\blpage{6}
(\byear{2023}).
\doiurl{10.1109/ISQED57927.2023.10129362}
\end{bchapter}
\endbibitem

%%% 67
\bibitem[\protect\citeauthoryear{Li et~al.}{2022}]{Learning_Automated_Driving_in_Complex_Intersection_Scenarios_Based}
\begin{barticle}
\bauthor{\bsnm{Li}, \binits{G.}},
\bauthor{\bsnm{Lin}, \binits{S.}},
\bauthor{\bsnm{Li}, \binits{S.}},
\bauthor{\bsnm{Qu}, \binits{X.}}:
\batitle{Learning {{Automated Driving}} in {{Complex Intersection Scenarios Based}} on {{Camera Sensors}}: {{A Deep Reinforcement Learning Approach}}}.
\bjtitle{IEEE Sensors Journal}
\bvolume{22}(\bissue{5}),
\bfpage{4687}--\blpage{4696}
(\byear{2022})
\doiurl{10.1109/JSEN.2022.3146307}
\end{barticle}
\endbibitem

%%% 68
\bibitem[\protect\citeauthoryear{Liu et~al.}{2021}]{A_Methodology_Based_on_Deep_Reinforcement_Learning}
\begin{bchapter}
\bauthor{\bsnm{Liu}, \binits{Z.}},
\bauthor{\bsnm{Hu}, \binits{J.}},
\bauthor{\bsnm{Song}, \binits{T.}},
\bauthor{\bsnm{Huang}, \binits{Z.}}:
\bctitle{A {{Methodology Based}} on {{Deep Reinforcement Learning}} to {{Autonomous Driving}} with {{Double Q-Learning}}}.
In: \bbtitle{2021 7th {{International Conference}} on {{Computer}} and {{Communications}} ({{ICCC}})},
pp. \bfpage{1266}--\blpage{1271}
(\byear{2021}).
\doiurl{10.1109/ICCC54389.2021.9674600}
\end{bchapter}
\endbibitem

%%% 69
\bibitem[\protect\citeauthoryear{Friji et~al.}{2020}]{frijiDQNBasedAutonomousCarFollowing2020}
\begin{bchapter}
\bauthor{\bsnm{Friji}, \binits{H.}},
\bauthor{\bsnm{Ghazzai}, \binits{H.}},
\bauthor{\bsnm{Besbes}, \binits{H.}},
\bauthor{\bsnm{Massoud}, \binits{Y.}}:
\bctitle{A {{DQN-Based Autonomous Car-Following Framework Using RGB-D Frames}}}.
In: \bbtitle{2020 {{IEEE Global Conference}} on {{Artificial Intelligence}} and {{Internet}} of {{Things}} ({{GCAIoT}})},
pp. \bfpage{1}--\blpage{6}
(\byear{2020}).
\doiurl{10.1109/GCAIoT51063.2020.9345899}
\end{bchapter}
\endbibitem

%%% 70
\bibitem[\protect\citeauthoryear{Hishmeh and Awad}{2020}]{hishmehDeerHeadlightsShort2020}
\begin{bchapter}
\bauthor{\bsnm{Hishmeh}, \binits{L.}},
\bauthor{\bsnm{Awad}, \binits{F.}}:
\bctitle{Deer in {{The Headlights}}: {{Short Term Planning}} via {{Reinforcement Learning Algorithms}} for {{Autonomous Vehicles}}}.
In: \bbtitle{2020 11th {{International Conference}} on {{Information}} and {{Communication Systems}} ({{ICICS}})},
pp. \bfpage{255}--\blpage{260}
(\byear{2020}).
\doiurl{10.1109/ICICS49469.2020.239561}
\end{bchapter}
\endbibitem

%%% 71
\bibitem[\protect\citeauthoryear{}{}]{221016567DeFIXDetecting}
\begin{botherref}
[2210.16567] {{DeFIX}}: {{Detecting}} and {{Fixing Failure Scenarios}} with {{Reinforcement Learning}} in {{Imitation Learning Based Autonomous Driving}}.
https://arxiv.org/abs/2210.16567
\end{botherref}
\endbibitem

%%% 72
\bibitem[\protect\citeauthoryear{Deshpande et~al.}{}]{deshpande_navigation_2021}
\begin{botherref}
\oauthor{\bsnm{Deshpande}, \binits{N.}},
\oauthor{\bsnm{Vaufreydaz}, \binits{D.}},
\oauthor{\bsnm{Spalanzani}, \binits{A.}}:
Navigation In Urban Environments Amongst Pedestrians Using Multi-Objective Deep Reinforcement Learning.
{arXiv}.
\doiurl{10.48550/arXiv.2110.05205} .
\url{http://arxiv.org/abs/2110.05205}
Accessed 2025-05-06
\end{botherref}
\endbibitem

%%% 73
\bibitem[\protect\citeauthoryear{}{}]{noauthor_explainable_nodate}
\begin{botherref}
Explainable Artificial Intelligence ({XAI}) Approach for Reinforcement Learning Systems {\textbar} Proceedings of the 39th {ACM}/{SIGAPP} Symposium on Applied Computing.
\url{https://dl.acm.org/doi/10.1145/3605098.3635992}
Accessed 2025-05-06
\end{botherref}
\endbibitem

%%% 74
\bibitem[\protect\citeauthoryear{Haarnoja et~al.}{}]{haarnoja_soft_2018}
\begin{botherref}
\oauthor{\bsnm{Haarnoja}, \binits{T.}},
\oauthor{\bsnm{Zhou}, \binits{A.}},
\oauthor{\bsnm{Abbeel}, \binits{P.}},
\oauthor{\bsnm{Levine}, \binits{S.}}:
Soft Actor-Critic: Off-Policy Maximum Entropy Deep Reinforcement Learning with a Stochastic Actor.
{arXiv}.
\doiurl{10.48550/arXiv.1801.01290} .
\url{http://arxiv.org/abs/1801.01290}
Accessed 2025-07-07
\end{botherref}
\endbibitem

%%% 75
\bibitem[\protect\citeauthoryear{Pomerleau}{1991}]{pomerleau1991efficient}
\begin{barticle}
\bauthor{\bsnm{Pomerleau}, \binits{D.A.}}:
\batitle{Efficient training of artificial neural networks for autonomous navigation}.
\bjtitle{Neural computation}
\bvolume{3}(\bissue{1}),
\bfpage{88}--\blpage{97}
(\byear{1991})
\end{barticle}
\endbibitem

%%% 76
\bibitem[\protect\citeauthoryear{Ho and Ermon}{2016}]{ho2016generative}
\begin{botherref}
\oauthor{\bsnm{Ho}, \binits{J.}},
\oauthor{\bsnm{Ermon}, \binits{S.}}:
Generative adversarial imitation learning.
Advances in neural information processing systems
\textbf{29}
(2016)
\end{botherref}
\endbibitem

%%% 77
\bibitem[\protect\citeauthoryear{Li et~al.}{2017}]{li2017infogail}
\begin{botherref}
\oauthor{\bsnm{Li}, \binits{Y.}},
\oauthor{\bsnm{Song}, \binits{J.}},
\oauthor{\bsnm{Ermon}, \binits{S.}}:
Infogail: Interpretable imitation learning from visual demonstrations.
Advances in neural information processing systems
\textbf{30}
(2017)
\end{botherref}
\endbibitem

%%% 78
\bibitem[\protect\citeauthoryear{Song et~al.}{2018}]{song2018multi}
\begin{botherref}
\oauthor{\bsnm{Song}, \binits{J.}},
\oauthor{\bsnm{Ren}, \binits{H.}},
\oauthor{\bsnm{Sadigh}, \binits{D.}},
\oauthor{\bsnm{Ermon}, \binits{S.}}:
Multi-agent generative adversarial imitation learning.
Advances in neural information processing systems
\textbf{31}
(2018)
\end{botherref}
\endbibitem

%%% 79
\bibitem[\protect\citeauthoryear{Savari and Choe}{2021}]{savari_online_2021}
\begin{bchapter}
\bauthor{\bsnm{Savari}, \binits{M.}},
\bauthor{\bsnm{Choe}, \binits{Y.}}:
\bctitle{Online {Virtual} {Training} in {Soft} {Actor}-{Critic} for {Autonomous} {Driving}}.
In: \bbtitle{2021 {International} {Joint} {Conference} on {Neural} {Networks} ({IJCNN})},
pp. \bfpage{1}--\blpage{8}
(\byear{2021}).
\doiurl{10.1109/IJCNN52387.2021.9533791} .
\bcomment{ISSN: 2161-4407}.
\burl{https://ieeexplore.ieee.org/document/9533791/?arnumber=9533791}
Accessed 2025-01-10
\end{bchapter}
\endbibitem

%%% 80
\bibitem[\protect\citeauthoryear{Wu et~al.}{2023}]{wu_reinforcement_2023}
\begin{bchapter}
\bauthor{\bsnm{Wu}, \binits{Y.}},
\bauthor{\bsnm{Wang}, \binits{L.}},
\bauthor{\bsnm{Lu}, \binits{X.}},
\bauthor{\bsnm{Wu}, \binits{Y.}},
\bauthor{\bsnm{Zhang}, \binits{H.}}:
\bctitle{Reinforcement {Learning}-based {Autonomous} {Parking} with {Expert} {Demonstrations}}.
In: \bbtitle{2023 7th {CAA} {International} {Conference} on {Vehicular} {Control} and {Intelligence} ({CVCI})},
pp. \bfpage{1}--\blpage{6}
(\byear{2023}).
\doiurl{10.1109/CVCI59596.2023.10397136} .
\burl{https://ieeexplore.ieee.org/document/10397136/?arnumber=10397136}
Accessed 2025-01-10
\end{bchapter}
\endbibitem

%%% 81
\bibitem[\protect\citeauthoryear{Aghdasian et~al.}{2023}]{aghdasian_autonomous_2023}
\begin{bchapter}
\bauthor{\bsnm{Aghdasian}, \binits{A.J.}},
\bauthor{\bsnm{Ardakani}, \binits{A.H.}},
\bauthor{\bsnm{Aqabakee}, \binits{K.}},
\bauthor{\bsnm{Abdollahi}, \binits{F.}}:
\bctitle{Autonomous {Driving} using {Residual} {Sensor} {Fusion} and {Deep} {Reinforcement} {Learning}}.
In: \bbtitle{2023 11th {RSI} {International} {Conference} on {Robotics} and {Mechatronics} ({ICRoM})},
pp. \bfpage{265}--\blpage{270}
(\byear{2023}).
\doiurl{10.1109/ICRoM60803.2023.10412516} .
\bcomment{ISSN: 2572-6889}.
\burl{https://ieeexplore.ieee.org/document/10412516/?arnumber=10412516}
Accessed 2025-01-10
\end{bchapter}
\endbibitem

%%% 82
\bibitem[\protect\citeauthoryear{Huang et~al.}{2023}]{huang_simoun_2023}
\begin{bchapter}
\bauthor{\bsnm{Huang}, \binits{Y.}},
\bauthor{\bsnm{Peng}, \binits{P.}},
\bauthor{\bsnm{Zhao}, \binits{Y.}},
\bauthor{\bsnm{Zhai}, \binits{Y.}},
\bauthor{\bsnm{Xu}, \binits{H.}},
\bauthor{\bsnm{Tian}, \binits{Y.}}:
\bctitle{Simoun: {Synergizing} {Interactive} {Motion}-appearance {Understanding} for {Vision}-based {Reinforcement} {Learning}}.
In: \bbtitle{2023 {IEEE}/{CVF} {International} {Conference} on {Computer} {Vision} ({ICCV})},
pp. \bfpage{176}--\blpage{185}
(\byear{2023}).
\doiurl{10.1109/ICCV51070.2023.00023} .
\bcomment{ISSN: 2380-7504}.
\burl{https://ieeexplore.ieee.org/document/10377012/?arnumber=10377012}
Accessed 2025-01-10
\end{bchapter}
\endbibitem

%%% 83
\bibitem[\protect\citeauthoryear{Kong et~al.}{2021}]{Enhanced_Off-Policy_Reinforcement_Learning_With_Focused_Experience_Replay}
\begin{barticle}
\bauthor{\bsnm{Kong}, \binits{S.-H.}},
\bauthor{\bsnm{Nahrendra}, \binits{I.M.A.}},
\bauthor{\bsnm{Paek}, \binits{D.-H.}}:
\batitle{Enhanced {{Off-Policy Reinforcement Learning With Focused Experience Replay}}}.
\bjtitle{IEEE Access}
\bvolume{9},
\bfpage{93152}--\blpage{93164}
(\byear{2021})
\doiurl{10.1109/ACCESS.2021.3085142}
\end{barticle}
\endbibitem

%%% 84
\bibitem[\protect\citeauthoryear{Jo et~al.}{2022}]{An_Integrated_Reward_Function}
\begin{bchapter}
\bauthor{\bsnm{Jo}, \binits{S.-B.}},
\bauthor{\bsnm{Kim}, \binits{P.-S.}},
\bauthor{\bsnm{Jeong}, \binits{H.-Y.}}:
\bctitle{An {{Integrated Reward Function}} of {{End-to-End Deep Reinforcement Learning}} for the {{Longitudinal}} and {{Lateral Control}} of {{Autonomous Vehicles}}}.
In: \bbtitle{2022 {{IEEE}} 95th {{Vehicular Technology Conference}}: ({{VTC2022-Spring}})},
pp. \bfpage{1}--\blpage{5}
(\byear{2022}).
\doiurl{10.1109/VTC2022-Spring54318.2022.9860646}
\end{bchapter}
\endbibitem

%%% 85
\bibitem[\protect\citeauthoryear{Tong et~al.}{2023}]{Urban_Autonomous_Driving_of_Emergency_Vehicles_with_Reinforcement_Learning}
\begin{bchapter}
\bauthor{\bsnm{Tong}, \binits{Z.}},
\bauthor{\bsnm{Ayala}, \binits{A.}},
\bauthor{\bsnm{Sandoval}, \binits{E.B.}},
\bauthor{\bsnm{Cruz}, \binits{F.}}:
\bctitle{Urban {{Autonomous Driving}} of {{Emergency Vehicles}} with {{Reinforcement Learning}}}.
In: \bbtitle{2023 {{IEEE Latin American Conference}} on {{Computational Intelligence}} ({{LA-CCI}})},
pp. \bfpage{1}--\blpage{6}
(\byear{2023}).
\doiurl{10.1109/LA-CCI58595.2023.10409469}
\end{bchapter}
\endbibitem

%%% 86
\bibitem[\protect\citeauthoryear{Han and Yilmaz}{2022}]{hanLearningDriveUsing2022}
\begin{botherref}
\oauthor{\bsnm{Han}, \binits{Y.}},
\oauthor{\bsnm{Yilmaz}, \binits{A.}}:
Learning to {{Drive Using Sparse Imitation Reinforcement Learning}}.
arXiv
(2022).
\doiurl{10.48550/arXiv.2205.12128}
\end{botherref}
\endbibitem

%%% 87
\bibitem[\protect\citeauthoryear{Igoe et~al.}{2023}]{igoeMultiAlphaSoftActorCritic2023}
\begin{bchapter}
\bauthor{\bsnm{Igoe}, \binits{C.}},
\bauthor{\bsnm{Pande}, \binits{S.}},
\bauthor{\bsnm{Venkatraman}, \binits{S.}},
\bauthor{\bsnm{Schneider}, \binits{J.}}:
\bctitle{Multi-{{Alpha Soft Actor-Critic}}: {{Overcoming Stochastic Biases}} in {{Maximum Entropy Reinforcement Learning}}}.
In: \bbtitle{2023 {{IEEE International Conference}} on {{Robotics}} and {{Automation}} ({{ICRA}})},
pp. \bfpage{7162}--\blpage{7168}
(\byear{2023}).
\doiurl{10.1109/ICRA48891.2023.10161395}
\end{bchapter}
\endbibitem

%%% 88
\bibitem[\protect\citeauthoryear{Gupta and Klusch}{}]{gupta_hylear_2023}
\begin{botherref}
\oauthor{\bsnm{Gupta}, \binits{D.}},
\oauthor{\bsnm{Klusch}, \binits{M.}}:
{HyLEAR}: Hybrid deep reinforcement learning and planning for safe and comfortable automated driving.
In: 2023 {IEEE} Intelligent Vehicles Symposium ({IV}),
pp. 1--8.
\doiurl{10.1109/IV55152.2023.10186781} .
{ISSN}: 2642-7214.
\url{https://ieeexplore.ieee.org/document/10186781}
Accessed 2025-05-06
\end{botherref}
\endbibitem

%%% 89
\bibitem[\protect\citeauthoryear{Fujimoto et~al.}{}]{fujimoto_addressing_2018}
\begin{botherref}
\oauthor{\bsnm{Fujimoto}, \binits{S.}},
\oauthor{\bsnm{Hoof}, \binits{H.}},
\oauthor{\bsnm{Meger}, \binits{D.}}:
Addressing function approximation error in actor-critic methods.
In: Proceedings of the 35th International Conference on Machine Learning,
pp. 1587--1596.
{PMLR}.
{ISSN}: 2640-3498.
\url{https://proceedings.mlr.press/v80/fujimoto18a.html}
Accessed 2025-07-07
\end{botherref}
\endbibitem

%%% 90
\bibitem[\protect\citeauthoryear{Jia et~al.}{2023}]{jia_safe_2023}
\begin{bchapter}
\bauthor{\bsnm{Jia}, \binits{R.}},
\bauthor{\bsnm{Lu}, \binits{S.}},
\bauthor{\bsnm{Xie}, \binits{W.}},
\bauthor{\bsnm{Dai}, \binits{L.}},
\bauthor{\bsnm{Wu}, \binits{Q.}},
\bauthor{\bsnm{Zhang}, \binits{J.}}:
\bctitle{Safe {Reinforcement} {Learning} for {Autonomous} {Lane}-{Changing}: {Considering} the {Collision} {Time} and {Lane} {Boundary} {Constraints}}.
In: \bbtitle{2023 {China} {Automation} {Congress} ({CAC})},
pp. \bfpage{3484}--\blpage{3489}
(\byear{2023}).
\doiurl{10.1109/CAC59555.2023.10450529} .
\bcomment{ISSN: 2688-0938}.
\burl{https://ieeexplore.ieee.org/document/10450529/?arnumber=10450529}
Accessed 2025-01-10
\end{bchapter}
\endbibitem

%%% 91
\bibitem[\protect\citeauthoryear{Xu et~al.}{2023}]{xu_end--end_2023}
\begin{bchapter}
\bauthor{\bsnm{Xu}, \binits{T.}},
\bauthor{\bsnm{Wang}, \binits{Q.}},
\bauthor{\bsnm{Yang}, \binits{Y.}},
\bauthor{\bsnm{Meng}, \binits{J.}},
\bauthor{\bsnm{Wang}, \binits{Y.}}:
\bctitle{End-to-{End} {Autonomous} {Driving} {Decision}-{Making} {Solution} {Based} on {Pri}-{TD3}}.
In: \bbtitle{2023 5th {International} {Conference} on {Robotics}, {Intelligent} {Control} and {Artificial} {Intelligence} ({RICAI})},
pp. \bfpage{525}--\blpage{529}
(\byear{2023}).
\doiurl{10.1109/RICAI60863.2023.10489418} .
\burl{https://ieeexplore.ieee.org/document/10489418/?arnumber=10489418}
Accessed 2025-01-10
\end{bchapter}
\endbibitem

%%% 92
\bibitem[\protect\citeauthoryear{Liao et~al.}{2023}]{liao_lateral_2023}
\begin{bchapter}
\bauthor{\bsnm{Liao}, \binits{Y.}},
\bauthor{\bsnm{Yu}, \binits{G.}},
\bauthor{\bsnm{Chen}, \binits{P.}},
\bauthor{\bsnm{Zhou}, \binits{B.}},
\bauthor{\bsnm{Li}, \binits{H.}}:
\bctitle{Lateral {Motion} {Control} for {Obstacle} {Avoidance} in {Autonomous} {Driving} {Based} on {Deep} {Reinforcement} {Learning}}.
In: \bbtitle{2023 {IEEE} 26th {International} {Conference} on {Intelligent} {Transportation} {Systems} ({ITSC})},
pp. \bfpage{5229}--\blpage{5234}
(\byear{2023}).
\doiurl{10.1109/ITSC57777.2023.10422057} .
\bcomment{ISSN: 2153-0017}.
\burl{https://ieeexplore.ieee.org/document/10422057/?arnumber=10422057}
Accessed 2025-01-10
\end{bchapter}
\endbibitem

%%% 93
\bibitem[\protect\citeauthoryear{Elallid et~al.}{2023}]{Deep_Reinforcement_Learning_for_Autonomous_Vehicle_Intersection_Navigation}
\begin{bchapter}
\bauthor{\bsnm{Elallid}, \binits{B.B.}},
\bauthor{\bsnm{El~Alaoui}, \binits{H.}},
\bauthor{\bsnm{Benamar}, \binits{N.}}:
\bctitle{Deep reinforcement learning for autonomous vehicle intersection navigation}.
In: \bbtitle{2023 International Conference on Innovation and Intelligence for Informatics, Computing, and Technologies (3ICT)},
pp. \bfpage{308}--\blpage{313}
(\byear{2023}).
\bcomment{IEEE}
\end{bchapter}
\endbibitem

%%% 94
\bibitem[\protect\citeauthoryear{Liu et~al.}{2024}]{Deep-Reinforcement-Learning-Based_Driving_Policy_at_Intersections}
\begin{barticle}
\bauthor{\bsnm{Liu}, \binits{Y.}},
\bauthor{\bsnm{Zhang}, \binits{Q.}},
\bauthor{\bsnm{Gao}, \binits{Y.}},
\bauthor{\bsnm{Zhao}, \binits{D.}}:
\batitle{Deep-{{Reinforcement-Learning-Based Driving Policy}} at {{Intersections Utilizing Lane Graph Networks}}}.
\bjtitle{IEEE Transactions on Cognitive and Developmental Systems}
\bvolume{16}(\bissue{5}),
\bfpage{1759}--\blpage{1774}
(\byear{2024})
\doiurl{10.1109/TCDS.2024.3384269}
\end{barticle}
\endbibitem

%%% 95
\bibitem[\protect\citeauthoryear{Li et~al.}{}]{li_decision-making_2024}
\begin{botherref}
\oauthor{\bsnm{Li}, \binits{S.}},
\oauthor{\bsnm{Peng}, \binits{K.}},
\oauthor{\bsnm{Hui}, \binits{F.}},
\oauthor{\bsnm{Li}, \binits{Z.}},
\oauthor{\bsnm{Wei}, \binits{C.}},
\oauthor{\bsnm{Wang}, \binits{W.}}:
A decision-making approach for complex unsignalized intersection by deep reinforcement learning
\textbf{73}(11),
16134--16147
\doiurl{10.1109/TVT.2024.3408917} .
Accessed 2025-05-06
\end{botherref}
\endbibitem

%%% 96
\bibitem[\protect\citeauthoryear{Khalil and Mouftah}{2023}]{khalil_exploiting_2023}
\begin{barticle}
\bauthor{\bsnm{Khalil}, \binits{Y.H.}},
\bauthor{\bsnm{Mouftah}, \binits{H.T.}}:
\batitle{Exploiting {Multi}-{Modal} {Fusion} for {Urban} {Autonomous} {Driving} {Using} {Latent} {Deep} {Reinforcement} {Learning}}.
\bjtitle{IEEE Transactions on Vehicular Technology}
\bvolume{72}(\bissue{3}),
\bfpage{2921}--\blpage{2935}
(\byear{2023})
\doiurl{10.1109/TVT.2022.3217299} .
\bcomment{Conference Name: IEEE Transactions on Vehicular Technology}.
Accessed 2025-01-10
\end{barticle}
\endbibitem

%%% 97
\bibitem[\protect\citeauthoryear{Qi et~al.}{2023}]{Cognitive_Reinforcement_Learning}
\begin{bchapter}
\bauthor{\bsnm{Qi}, \binits{H.}},
\bauthor{\bsnm{Hou}, \binits{E.}},
\bauthor{\bsnm{Liu}, \binits{G.}},
\bauthor{\bsnm{Ye}, \binits{P.}}:
\bctitle{Cognitive {{Reinforcement Learning For Autonomous Driving}}}.
In: \bbtitle{2023 {{IEEE}} 3rd {{International Conference}} on {{Digital Twins}} and {{Parallel Intelligence}} ({{DTPI}})},
pp. \bfpage{1}--\blpage{5}
(\byear{2023}).
\doiurl{10.1109/DTPI59677.2023.10365456}
\end{bchapter}
\endbibitem

%%% 98
\bibitem[\protect\citeauthoryear{Couto and Antonelo}{2024}]{coutoHierarchicalGenerativeAdversarial2024}
\begin{botherref}
\oauthor{\bsnm{Couto}, \binits{G.C.K.}},
\oauthor{\bsnm{Antonelo}, \binits{E.A.}}:
Hierarchical generative adversarial imitation learning with mid-level input generation for autonomous driving on urban environments.
IEEE Transactions on Intelligent Vehicles
(2024)
\end{botherref}
\endbibitem

%%% 99
\bibitem[\protect\citeauthoryear{Jaafra et~al.}{2019}]{jaafraContextAwareAutonomousDriving2019}
\begin{bchapter}
\bauthor{\bsnm{Jaafra}, \binits{Y.}},
\bauthor{\bsnm{Deruyver}, \binits{A.}},
\bauthor{\bsnm{Laurent}, \binits{J.L.}},
\bauthor{\bsnm{Naceur}, \binits{M.S.}}:
\bctitle{Context-{{Aware Autonomous Driving Using Meta-Reinforcement Learning}}}.
In: \bbtitle{2019 18th {{IEEE International Conference On Machine Learning And Applications}} ({{ICMLA}})},
pp. \bfpage{450}--\blpage{455}
(\byear{2019}).
\doiurl{10.1109/ICMLA.2019.00084}
\end{bchapter}
\endbibitem

%%% 100
\bibitem[\protect\citeauthoryear{Baheri et~al.}{}]{baheri_vision-based_2020}
\begin{botherref}
\oauthor{\bsnm{Baheri}, \binits{A.}},
\oauthor{\bsnm{Kolmanovsky}, \binits{I.}},
\oauthor{\bsnm{Girard}, \binits{A.}},
\oauthor{\bsnm{Tseng}, \binits{H.E.}},
\oauthor{\bsnm{Filev}, \binits{D.}}:
Vision-based autonomous driving: A model learning approach.
In: 2020 American Control Conference ({ACC}),
pp. 2520--2525.
\doiurl{10.23919/ACC45564.2020.9147510} .
{ISSN}: 2378-5861.
\url{https://ieeexplore.ieee.org/document/9147510}
Accessed 2025-05-06
\end{botherref}
\endbibitem

%%% 101
\bibitem[\protect\citeauthoryear{Wang et~al.}{2022}]{Versatile_and_Efficien}
\begin{botherref}
\oauthor{\bsnm{Wang}, \binits{G.}},
\oauthor{\bsnm{Niu}, \binits{H.}},
\oauthor{\bsnm{Zhu}, \binits{D.}},
\oauthor{\bsnm{Hu}, \binits{J.}},
\oauthor{\bsnm{Zhan}, \binits{X.}},
\oauthor{\bsnm{Zhou}, \binits{G.}}:
A {{Versatile}} and {{Efficient Reinforcement Learning Framework}} for {{Autonomous Driving}}.
arXiv.
Comment: 8 pages, 6 figures
(2022).
\doiurl{10.48550/arXiv.2110.11573}
\end{botherref}
\endbibitem

%%% 102
\bibitem[\protect\citeauthoryear{Wang et~al.}{2021}]{wang_benchmarking_2021}
\begin{bchapter}
\bauthor{\bsnm{Wang}, \binits{J.}},
\bauthor{\bsnm{Zhang}, \binits{Q.}},
\bauthor{\bsnm{Zhao}, \binits{D.}}:
\bctitle{Benchmarking {Lane}-changing {Decision}-making for {Deep} {Reinforcement} {Learning}}.
In: \bbtitle{2021 7th {International} {Conference} on {Robotics} and {Artificial} {Intelligence}},
pp. \bfpage{26}--\blpage{32}.
\bpublisher{ACM},
\blocation{Guangzhou China}
(\byear{2021}).
\doiurl{10.1145/3505688.3505693} .
\burl{https://dl.acm.org/doi/10.1145/3505688.3505693}
Accessed 2025-01-10
\end{bchapter}
\endbibitem

%%% 103
\bibitem[\protect\citeauthoryear{Cai et~al.}{2021}]{A_Decision_Control_Method}
\begin{barticle}
\bauthor{\bsnm{Cai}, \binits{Y.}},
\bauthor{\bsnm{Yang}, \binits{S.}},
\bauthor{\bsnm{Wang}, \binits{H.}},
\bauthor{\bsnm{Teng}, \binits{C.}},
\bauthor{\bsnm{Chen}, \binits{L.}}:
\batitle{A {{Decision Control Method}} for {{Autonomous Driving Based}} on {{Multi-Task Reinforcement Learning}}}.
\bjtitle{IEEE Access}
\bvolume{9},
\bfpage{154553}--\blpage{154562}
(\byear{2021})
\doiurl{10.1109/ACCESS.2021.3126796}
\end{barticle}
\endbibitem

%%% 104
\bibitem[\protect\citeauthoryear{Xu et~al.}{2022}]{xu_decision-making_2022}
\begin{bchapter}
\bauthor{\bsnm{Xu}, \binits{S.-Y.}},
\bauthor{\bsnm{Chen}, \binits{X.-M.}},
\bauthor{\bsnm{Wang}, \binits{Z.-J.}},
\bauthor{\bsnm{Hu}, \binits{Y.-H.}},
\bauthor{\bsnm{Han}, \binits{X.-T.}}:
\bctitle{Decision-{Making} {Models} for {Autonomous} {Vehicles} at {Unsignalized} {Intersections} {Based} on {Deep} {Reinforcement} {Learning}}.
In: \bbtitle{2022 {International} {Conference} on {Advanced} {Robotics} and {Mechatronics} ({ICARM})},
pp. \bfpage{672}--\blpage{677}
(\byear{2022}).
\doiurl{10.1109/ICARM54641.2022.9959664} .
\burl{https://ieeexplore.ieee.org/document/9959664/?arnumber=9959664}
Accessed 2025-01-10
\end{bchapter}
\endbibitem

%%% 105
\bibitem[\protect\citeauthoryear{Manikandan et~al.}{2023}]{manikandan_ad_2023}
\begin{barticle}
\bauthor{\bsnm{Manikandan}, \binits{N.S.}},
\bauthor{\bsnm{Kaliyaperumal}, \binits{G.}},
\bauthor{\bsnm{Wang}, \binits{Y.}}:
\batitle{Ad {Hoc}-{Obstacle} {Avoidance}-{Based} {Navigation} {System} {Using} {Deep} {Reinforcement} {Learning} for {Self}-{Driving} {Vehicles}}.
\bjtitle{IEEE Access}
\bvolume{11},
\bfpage{92285}--\blpage{92297}
(\byear{2023})
\doiurl{10.1109/ACCESS.2023.3297661} .
\bcomment{Conference Name: IEEE Access}.
Accessed 2025-01-10
\end{barticle}
\endbibitem

%%% 106
\bibitem[\protect\citeauthoryear{Frauenknecht et~al.}{2023}]{frauenknecht_data-efficient_2023}
\begin{botherref}
\oauthor{\bsnm{Frauenknecht}, \binits{B.}},
\oauthor{\bsnm{Ehlgen}, \binits{T.}},
\oauthor{\bsnm{Trimpe}, \binits{S.}}:
Data-efficient {Deep} {Reinforcement} {Learning} for {Vehicle} {Trajectory} {Control}.
arXiv.
arXiv:2311.18393 [cs]
(2023).
\doiurl{10.48550/arXiv.2311.18393} .
\url{http://arxiv.org/abs/2311.18393}
Accessed 2025-01-10
\end{botherref}
\endbibitem

%%% 107
\bibitem[\protect\citeauthoryear{Sutton}{}]{sutton_dyna_1991}
\begin{botherref}
\oauthor{\bsnm{Sutton}, \binits{R.S.}}:
Dyna, an integrated architecture for learning, planning, and reacting
\textbf{2}(4),
160--163
\doiurl{10.1145/122344.122377} .
Accessed 2025-07-07
\end{botherref}
\endbibitem

%%% 108
\bibitem[\protect\citeauthoryear{Chua et~al.}{}]{chua_deep_2018}
\begin{botherref}
\oauthor{\bsnm{Chua}, \binits{K.}},
\oauthor{\bsnm{Calandra}, \binits{R.}},
\oauthor{\bsnm{{McAllister}}, \binits{R.}},
\oauthor{\bsnm{Levine}, \binits{S.}}:
Deep Reinforcement Learning in a Handful of Trials using Probabilistic Dynamics Models.
{arXiv}.
\doiurl{10.48550/arXiv.1805.12114} .
\url{http://arxiv.org/abs/1805.12114}
Accessed 2025-07-07
\end{botherref}
\endbibitem

%%% 109
\bibitem[\protect\citeauthoryear{Janner et~al.}{}]{janner_when_2021}
\begin{botherref}
\oauthor{\bsnm{Janner}, \binits{M.}},
\oauthor{\bsnm{Fu}, \binits{J.}},
\oauthor{\bsnm{Zhang}, \binits{M.}},
\oauthor{\bsnm{Levine}, \binits{S.}}:
When to Trust Your Model: Model-Based Policy Optimization.
{arXiv}.
\doiurl{10.48550/arXiv.1906.08253} .
\url{http://arxiv.org/abs/1906.08253}
Accessed 2025-07-07
\end{botherref}
\endbibitem

%%% 110
\bibitem[\protect\citeauthoryear{Hafner et~al.}{}]{hafner_learning_2019}
\begin{botherref}
\oauthor{\bsnm{Hafner}, \binits{D.}},
\oauthor{\bsnm{Lillicrap}, \binits{T.}},
\oauthor{\bsnm{Fischer}, \binits{I.}},
\oauthor{\bsnm{Villegas}, \binits{R.}},
\oauthor{\bsnm{Ha}, \binits{D.}},
\oauthor{\bsnm{Lee}, \binits{H.}},
\oauthor{\bsnm{Davidson}, \binits{J.}}:
Learning latent dynamics for planning from pixels.
In: Proceedings of the 36th International Conference on Machine Learning,
pp. 2555--2565.
{PMLR}.
{ISSN}: 2640-3498.
\url{https://proceedings.mlr.press/v97/hafner19a.html}
Accessed 2025-07-07
\end{botherref}
\endbibitem

%%% 111
\bibitem[\protect\citeauthoryear{Hafner et~al.}{}]{hafner_mastering_2024}
\begin{botherref}
\oauthor{\bsnm{Hafner}, \binits{D.}},
\oauthor{\bsnm{Pasukonis}, \binits{J.}},
\oauthor{\bsnm{Ba}, \binits{J.}},
\oauthor{\bsnm{Lillicrap}, \binits{T.}}:
Mastering Diverse Domains through World Models.
{arXiv}.
\doiurl{10.48550/arXiv.2301.04104} .
\url{http://arxiv.org/abs/2301.04104}
Accessed 2025-07-07
\end{botherref}
\endbibitem

%%% 112
\bibitem[\protect\citeauthoryear{Ghavamzadeh et~al.}{2015}]{ghavamzadeh2015bayesian}
\begin{barticle}
\bauthor{\bsnm{Ghavamzadeh}, \binits{M.}},
\bauthor{\bsnm{Mannor}, \binits{S.}},
\bauthor{\bsnm{Pineau}, \binits{J.}},
\bauthor{\bsnm{Tamar}, \binits{A.}}:
\batitle{Bayesian reinforcement learning: A survey}.
\bjtitle{Foundations and Trends in Machine Learning}
\bvolume{8}(\bissue{5-6}),
\bfpage{359}--\blpage{483}
(\byear{2015})
\end{barticle}
\endbibitem

%%% 113
\bibitem[\protect\citeauthoryear{Gharaee et~al.}{2021}]{A_Bayesian_Approach_to_Reinforcement_Learning_of_Vision-Based_Vehicular_Control}
\begin{botherref}
\oauthor{\bsnm{Gharaee}, \binits{Z.}},
\oauthor{\bsnm{Holmquist}, \binits{K.}},
\oauthor{\bsnm{He}, \binits{L.}},
\oauthor{\bsnm{Felsberg}, \binits{M.}}:
A {{Bayesian Approach}} to {{Reinforcement Learning}} of {{Vision-Based Vehicular Control}}.
arXiv
(2021).
\doiurl{10.48550/arXiv.2104.03807}
\end{botherref}
\endbibitem

%%% 114
\bibitem[\protect\citeauthoryear{Sutton et~al.}{1999}]{sutton1999between}
\begin{bchapter}
\bauthor{\bsnm{Sutton}, \binits{R.S.}},
\bauthor{\bsnm{Precup}, \binits{D.}},
\bauthor{\bsnm{Singh}, \binits{S.}}:
\bctitle{Between mdps and semi-mdps: A framework for temporal abstraction in reinforcement learning}.
In: \bbtitle{Artificial Intelligence},
vol. \bseriesno{112},
pp. \bfpage{181}--\blpage{211}
(\byear{1999})
\end{bchapter}
\endbibitem

%%% 115
\bibitem[\protect\citeauthoryear{Li}{2021}]{A_Hierarchical_Autonomous_Driving_Framework}
\begin{bchapter}
\bauthor{\bsnm{Li}, \binits{Z.}}:
\bctitle{A {{Hierarchical Autonomous Driving Framework Combining Reinforcement Learning}} and {{Imitation Learning}}}.
In: \bbtitle{2021 {{International Conference}} on {{Computer Engineering}} and {{Application}} ({{ICCEA}})},
pp. \bfpage{395}--\blpage{400}
(\byear{2021}).
\doiurl{10.1109/ICCEA53728.2021.00084}
\end{bchapter}
\endbibitem

%%% 116
\bibitem[\protect\citeauthoryear{Wang et~al.}{2024}]{Hierarchical_Reinforcement_Learning_with_Successor_Representation_for_Intelligent_Vehicle_Collision_Avoidance_of_Dynamic_Pedestrian}
\begin{bchapter}
\bauthor{\bsnm{Wang}, \binits{H.}},
\bauthor{\bsnm{Lu}, \binits{C.}},
\bauthor{\bsnm{Gong}, \binits{J.}}:
\bctitle{Hierarchical {{Reinforcement Learning}} with {{Successor Representation}} for {{Intelligent Vehicle Collision Avoidance}} of {{Dynamic Pedestrian}}}.
In: \bbtitle{2024 9th {{International Conference}} on {{Computer}} and {{Communication Systems}} ({{ICCCS}})},
pp. \bfpage{575}--\blpage{581}
(\byear{2024}).
\doiurl{10.1109/ICCCS61882.2024.10603229}
\end{bchapter}
\endbibitem

%%% 117
\bibitem[\protect\citeauthoryear{He et~al.}{2022}]{heAdaptiveDecisionMaking2022}
\begin{botherref}
\oauthor{\bsnm{He}, \binits{X.}},
\oauthor{\bsnm{Yang}, \binits{L.}},
\oauthor{\bsnm{Lu}, \binits{C.}},
\oauthor{\bsnm{Li}, \binits{Z.}},
\oauthor{\bsnm{Gong}, \binits{J.}}:
Adaptive {{Decision Making}} at the {{Intersection}} for {{Autonomous Vehicles Based}} on {{Skill Discovery}}.
arXiv.
Comment: Accepted by IEEE ITSC 2022
(2022).
\doiurl{10.48550/arXiv.2207.11724}
\end{botherref}
\endbibitem

%%% 118
\bibitem[\protect\citeauthoryear{Naveed et~al.}{2020}]{naveedTrajectoryPlanningAutonomous2020}
\begin{botherref}
\oauthor{\bsnm{Naveed}, \binits{K.B.}},
\oauthor{\bsnm{Qiao}, \binits{Z.}},
\oauthor{\bsnm{Dolan}, \binits{J.M.}}:
Trajectory {{Planning}} for {{Autonomous Vehicles Using Hierarchical Reinforcement Learning}}.
arXiv.
Comment: 7 pages, 5 figures
(2020).
\doiurl{10.48550/arXiv.2011.04752}
\end{botherref}
\endbibitem

%%% 119
\bibitem[\protect\citeauthoryear{Gangopadhyay et~al.}{2022}]{gangopadhyayHierarchicalProgramTriggeredReinforcement2022}
\begin{barticle}
\bauthor{\bsnm{Gangopadhyay}, \binits{B.}},
\bauthor{\bsnm{Soora}, \binits{H.}},
\bauthor{\bsnm{Dasgupta}, \binits{P.}}:
\batitle{Hierarchical {{Program-Triggered Reinforcement Learning Agents For Automated Driving}}}.
\bjtitle{IEEE Transactions on Intelligent Transportation Systems}
\bvolume{23}(\bissue{8}),
\bfpage{10902}--\blpage{10911}
(\byear{2022})
\doiurl{10.1109/TITS.2021.3096998}
{\href{https://arxiv.org/abs/2103.13861}{{arXiv:2103.13861}}}
{[cs]}.
\bcomment{Comment: The paper is under consideration in Transactions on Intelligent Transportation Systems}
\end{barticle}
\endbibitem

%%% 120
\bibitem[\protect\citeauthoryear{Ziebart}{2010}]{ziebart2010modeling}
\begin{botherref}
\oauthor{\bsnm{Ziebart}, \binits{B.D.}}:
Modeling purposeful adaptive behavior with the principle of maximum causal entropy.
PhD thesis,
Carnegie Mellon University
(2010)
\end{botherref}
\endbibitem

%%% 121
\bibitem[\protect\citeauthoryear{Haarnoja et~al.}{2017}]{haarnoja2017reinforcement}
\begin{bchapter}
\bauthor{\bsnm{Haarnoja}, \binits{T.}},
\bauthor{\bsnm{Tang}, \binits{H.}},
\bauthor{\bsnm{Abbeel}, \binits{P.}},
\bauthor{\bsnm{Levine}, \binits{S.}}:
\bctitle{Reinforcement learning with deep energy-based policies}.
In: \bbtitle{Proceedings of the 34th International Conference on Machine Learning},
pp. \bfpage{1352}--\blpage{1361}
(\byear{2017})
\end{bchapter}
\endbibitem

%%% 122
\bibitem[\protect\citeauthoryear{Khalil and Mouftah}{2021}]{Integration_of_Motion_Prediction_with_End-to-end_Latent_RL_for_Self-Driving_Vehicles}
\begin{bchapter}
\bauthor{\bsnm{Khalil}, \binits{Y.H.}},
\bauthor{\bsnm{Mouftah}, \binits{H.T.}}:
\bctitle{Integration of {{Motion Prediction}} with {{End-to-end Latent RL}} for {{Self-Driving Vehicles}}}.
In: \bbtitle{2021 {{International Wireless Communications}} and {{Mobile Computing}} ({{IWCMC}})},
pp. \bfpage{1111}--\blpage{1116}
(\byear{2021}).
\doiurl{10.1109/IWCMC51323.2021.9498581}
\end{bchapter}
\endbibitem

%%% 123
\bibitem[\protect\citeauthoryear{Bellemare et~al.}{2017}]{bellemare2017distributional}
\begin{bchapter}
\bauthor{\bsnm{Bellemare}, \binits{M.G.}},
\bauthor{\bsnm{Dabney}, \binits{W.}},
\bauthor{\bsnm{Munos}, \binits{R.}}:
\bctitle{A distributional perspective on reinforcement learning}.
In: \bbtitle{International Conference on Machine Learning},
pp. \bfpage{449}--\blpage{458}
(\byear{2017})
\end{bchapter}
\endbibitem

%%% 124
\bibitem[\protect\citeauthoryear{Yang et~al.}{2019}]{yang2019fully}
\begin{bchapter}
\bauthor{\bsnm{Yang}, \binits{Z.}},
\bauthor{\bsnm{Zhang}, \binits{X.}},
\bauthor{\bsnm{Meng}, \binits{D.}},
\bauthor{\bsnm{Wen}, \binits{Y.}},
\bauthor{\bsnm{Zhang}, \binits{B.}},
\bauthor{\bsnm{Xu}, \binits{Z.}}:
\bctitle{Fully parameterized quantile function for distributional reinforcement learning}.
In: \bbtitle{Advances in Neural Information Processing Systems}
(\byear{2019})
\end{bchapter}
\endbibitem

%%% 125
\bibitem[\protect\citeauthoryear{Chen et~al.}{2024}]{chen_motion_2024}
\begin{bchapter}
\bauthor{\bsnm{Chen}, \binits{X.}},
\bauthor{\bsnm{Yang}, \binits{Y.}},
\bauthor{\bsnm{Xu}, \binits{S.}},
\bauthor{\bsnm{Fu}, \binits{S.}},
\bauthor{\bsnm{Yang}, \binits{D.}}:
\bctitle{Motion {Planning} for {Autonomous} {Vehicles} in {Uncertain} {Environments} {Using} {Hierarchical} {Distributional} {Reinforcement} {Learning}}.
In: \bbtitle{2024 36th {Chinese} {Control} and {Decision} {Conference} ({CCDC})},
pp. \bfpage{1844}--\blpage{1851}
(\byear{2024}).
\doiurl{10.1109/CCDC62350.2024.10587952} .
\bcomment{ISSN: 1948-9447}.
\burl{https://ieeexplore.ieee.org/document/10587952/?arnumber=10587952}
Accessed 2025-01-10
\end{bchapter}
\endbibitem

%%% 126
\bibitem[\protect\citeauthoryear{Sharma et~al.}{2020}]{sharma2020emergent}
\begin{bchapter}
\bauthor{\bsnm{Sharma}, \binits{A.}},
\bauthor{\bsnm{Jain}, \binits{A.}},
\bauthor{\bsnm{Finn}, \binits{C.}},
\bauthor{\bsnm{Levine}, \binits{S.}}:
\bctitle{Emergent latent skills in multi-task reinforcement learning}.
In: \bbtitle{Conference on Robot Learning},
pp. \bfpage{973}--\blpage{991}
(\byear{2020})
\end{bchapter}
\endbibitem

%%% 127
\bibitem[\protect\citeauthoryear{Li et~al.}{2023}]{Boosting_Offline_Reinforcement_Learning}
\begin{botherref}
\oauthor{\bsnm{Li}, \binits{Z.}},
\oauthor{\bsnm{Nie}, \binits{F.}},
\oauthor{\bsnm{Sun}, \binits{Q.}},
\oauthor{\bsnm{Da}, \binits{F.}},
\oauthor{\bsnm{Zhao}, \binits{H.}}:
Boosting {{Offline Reinforcement Learning}} for {{Autonomous Driving}} with {{Hierarchical Latent Skills}}.
arXiv
(2023).
\doiurl{10.48550/arXiv.2309.13614}
\end{botherref}
\endbibitem

%%% 128
\bibitem[\protect\citeauthoryear{Hussonnois and Jun}{}]{hussonnois_end--end_2022}
\begin{botherref}
\oauthor{\bsnm{Hussonnois}, \binits{M.}},
\oauthor{\bsnm{Jun}, \binits{J.-Y.}}:
End-to-end autonomous driving using the ape-x algorithm in carla simulation environment.
In: 2022 Thirteenth International Conference on Ubiquitous and Future Networks ({ICUFN}),
pp. 18--23.
\doiurl{10.1109/ICUFN55119.2022.9829674} .
{ISSN}: 2165-8536.
\url{https://ieeexplore.ieee.org/document/9829674}
Accessed 2025-05-06
\end{botherref}
\endbibitem

%%% 129
\bibitem[\protect\citeauthoryear{Zhao et~al.}{2021}]{zhao_real-time_2021}
\begin{bchapter}
\bauthor{\bsnm{Zhao}, \binits{X.}},
\bauthor{\bsnm{Zhu}, \binits{X.}},
\bauthor{\bsnm{Su}, \binits{Y.}}:
\bctitle{A {Real}-time and {Robust} {Visual}-based {Autonomous} {Driving} {Model}}.
In: \bbtitle{2021 6th {International} {Conference} on {Computational} {Intelligence} and {Applications} ({ICCIA})},
pp. \bfpage{257}--\blpage{261}
(\byear{2021}).
\doiurl{10.1109/ICCIA52886.2021.00057} .
\burl{https://ieeexplore.ieee.org/document/9644099/?arnumber=9644099}
Accessed 2025-01-10
\end{bchapter}
\endbibitem

%%% 130
\bibitem[\protect\citeauthoryear{Wu et~al.}{}]{wu_motionnet_2020}
\begin{botherref}
\oauthor{\bsnm{Wu}, \binits{P.}},
\oauthor{\bsnm{Chen}, \binits{S.}},
\oauthor{\bsnm{Metaxas}, \binits{D.}}:
{MotionNet}: Joint Perception and Motion Prediction for Autonomous Driving Based on Bird's Eye View Maps.
{arXiv}.
\doiurl{10.48550/arXiv.2003.06754} .
\url{http://arxiv.org/abs/2003.06754}
Accessed 2025-07-07
\end{botherref}
\endbibitem

%%% 131
\bibitem[\protect\citeauthoryear{{P{\'e}rez-Gill} et~al.}{2021}]{Deep_Reinforcement_Learning_based_control_algorithms:_Training_and_validation_using_the_ROS_Framework}
\begin{bchapter}
\bauthor{\bsnm{{P{\'e}rez-Gill}}, \binits{{\'O}.}},
\bauthor{\bsnm{Barea}, \binits{R.}},
\bauthor{\bsnm{{L{\'o}pez-Guill{\'e}n}}, \binits{E.}},
\bauthor{\bsnm{Bergasa}, \binits{L.M.}},
\bauthor{\bsnm{{G{\'o}mez-Hu{\'e}lamo}}, \binits{C.}},
\bauthor{\bsnm{Guti{\'e}rrez}, \binits{R.}},
\bauthor{\bsnm{D{\'i}az}, \binits{A.}}:
\bctitle{Deep {{Reinforcement Learning}} based control algorithms: {{Training}} and validation using the {{ROS Framework}} in {{CARLA Simulator}} for {{Self-Driving}} applications}.
In: \bbtitle{2021 {{IEEE Intelligent Vehicles Symposium}} ({{IV}})},
pp. \bfpage{1268}--\blpage{1273}
(\byear{2021}).
\doiurl{10.1109/IV48863.2021.9575616}
\end{bchapter}
\endbibitem

%%% 132
\bibitem[\protect\citeauthoryear{Liang et~al.}{2018}]{Liang_2018_ECCV}
\begin{bchapter}
\bauthor{\bsnm{Liang}, \binits{X.}},
\bauthor{\bsnm{Wang}, \binits{T.}},
\bauthor{\bsnm{Yang}, \binits{L.}},
\bauthor{\bsnm{Xing}, \binits{E.}}:
\bctitle{Cirl: Controllable imitative reinforcement learning for vision-based self-driving}.
In: \bbtitle{Proceedings of the European Conference on Computer Vision (ECCV)}
(\byear{2018})
\end{bchapter}
\endbibitem

%%% 133
\bibitem[\protect\citeauthoryear{Sauer et~al.}{2018}]{sauer2018conditionalaffordancelearningdriving}
\begin{botherref}
\oauthor{\bsnm{Sauer}, \binits{A.}},
\oauthor{\bsnm{Savinov}, \binits{N.}},
\oauthor{\bsnm{Geiger}, \binits{A.}}:
Conditional Affordance Learning for Driving in Urban Environments
(2018).
\url{https://arxiv.org/abs/1806.06498}
\end{botherref}
\endbibitem

%%% 134
\bibitem[\protect\citeauthoryear{Codevilla et~al.}{2019}]{Codevilla_2019_ICCV}
\begin{bchapter}
\bauthor{\bsnm{Codevilla}, \binits{F.}},
\bauthor{\bsnm{Santana}, \binits{E.}},
\bauthor{\bsnm{Lopez}, \binits{A.M.}},
\bauthor{\bsnm{Gaidon}, \binits{A.}}:
\bctitle{Exploring the limitations of behavior cloning for autonomous driving}.
In: \bbtitle{Proceedings of the IEEE/CVF International Conference on Computer Vision (ICCV)}
(\byear{2019})
\end{bchapter}
\endbibitem

\end{thebibliography}

\end{document}